\pdfoutput=1
\documentclass{lovartreport}
\newcolumntype{Y}{>{\raggedright\arraybackslash}X}
\DeclareSymbolFont{textdigits}{T1}{\rmdefault}{m}{n}
\SetSymbolFont{textdigits}{bold}{T1}{\rmdefault}{\bfdefault}{n}
\DeclareMathSymbol{0}{\mathalpha}{textdigits}{`0}
\DeclareMathSymbol{1}{\mathalpha}{textdigits}{`1}
\DeclareMathSymbol{2}{\mathalpha}{textdigits}{`2}
\DeclareMathSymbol{3}{\mathalpha}{textdigits}{`3}
\DeclareMathSymbol{4}{\mathalpha}{textdigits}{`4}
\DeclareMathSymbol{5}{\mathalpha}{textdigits}{`5}
\DeclareMathSymbol{6}{\mathalpha}{textdigits}{`6}
\DeclareMathSymbol{7}{\mathalpha}{textdigits}{`7}
\DeclareMathSymbol{8}{\mathalpha}{textdigits}{`8}
\DeclareMathSymbol{9}{\mathalpha}{textdigits}{`9}
\DeclareMathSymbol{.}{\mathord}{textdigits}{`.}
\newcommand{\sigmark}{\rlap{$^{*}$}}
\renewcommand{\reporttoprule}{\arrayrulecolor{LovartAccent}\toprule\arrayrulecolor{LovartRule}}

\title{What Helps Denoising Can Hurt Memory: Rethinking History in Long Video Generation}
\author{\mbox{Yubin Jing\textsuperscript{1,*}},\quad
  \mbox{Ziyang Zhang\textsuperscript{2,*}},\quad
  \mbox{Meng Chen\textsuperscript{3}},\quad
  \mbox{Yuyao Li\textsuperscript{4,*}},\quad
  \mbox{Haofan Wang\textsuperscript{5}},\quad
  \mbox{Yichen Gong\textsuperscript{5,\dag}}}
\reportbrand{LiblibAI}{Research}
\reporttype{}
\reportnumber{}
\reportshorttitle{Commit Forcing}
\reportauthorlayout{wide}
\reportauthornote{%
  \textsuperscript{1}Georgia Institute of Technology\quad
  \textsuperscript{2}Peking University\quad
  \textsuperscript{3}Zhejiang University\quad
  \textsuperscript{4}Tsinghua University\quad
  \textsuperscript{5}LiblibAI\\
  \textsuperscript{*}Work done during an internship at LiblibAI\quad
  \textsuperscript{\dag}Corresponding author}
\reportlogo{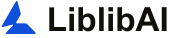}
\reportpartners{%
  \partnerwordmark[trim=21bp 8bp 21bp 8bp,clip]{31mm}{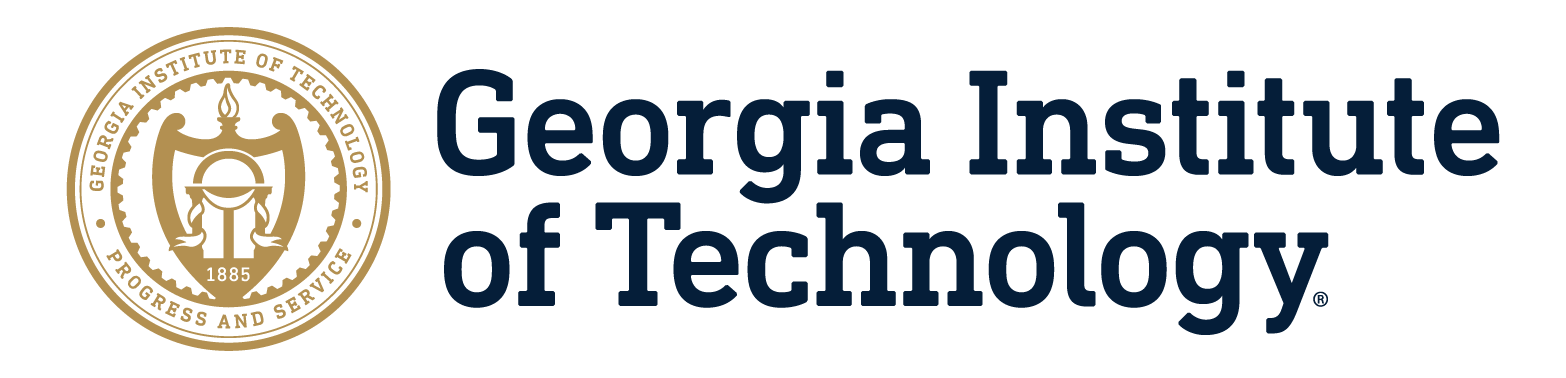}\hfill
  \partnerwordmark{28mm}{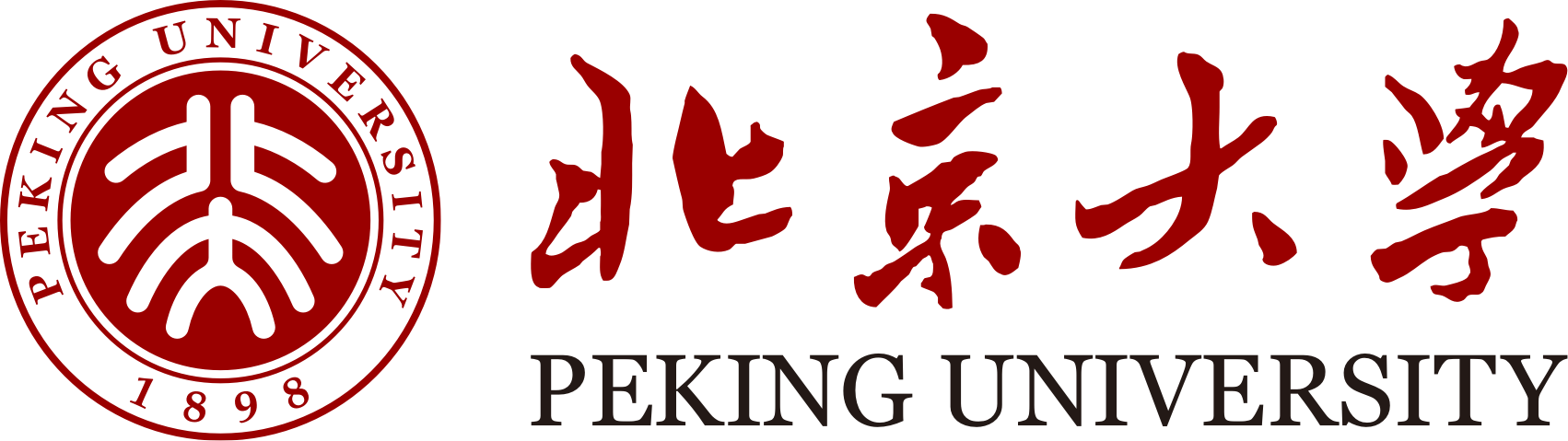}\hfill
  \partnerwordmark{28mm}{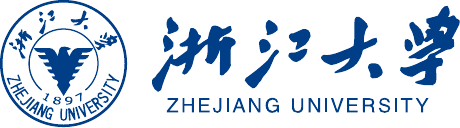}\hfill
  \partnerwordmark[trim=228bp 142bp 275bp 151bp,clip]{25mm}{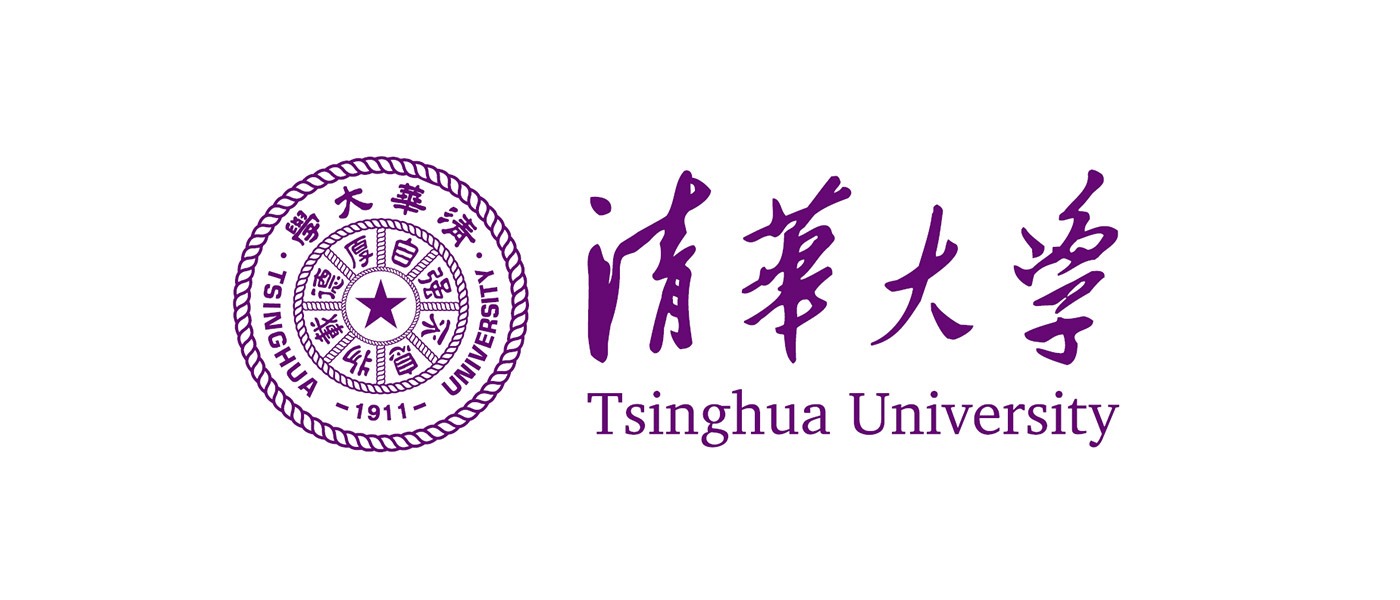}}
\reportmascot{}
\reportabstractstyle{flat}
\date{November 2026}
\makeatletter
\hypersetup{
  pdftitle={\@title},
  pdfauthor={Yubin Jing, Ziyang Zhang, Meng Chen, Yuyao Li, Haofan Wang, Yichen Gong}
}
\makeatother

\begin{document}
\maketitle
\begin{abstract}
Autoregressive video generators produce videos block by block from historical context. They are trained within a bounded horizon, whereas inference can extend far beyond it, causing temporal consistency to gradually decrease. Existing methods focus on eliminating this mismatch, but we instead ask whether the historical context used in training remains suitable for inference. Each block reads historical context in two stages, when denoising generates it and when committing writes it into the KV cache for later blocks. We find that older historical context such as the sink helps denoising but can hurt later blocks when read during committing, even in generators trained to read it there. We also show that as committing reads more preceding blocks, clip-to-clip Subject Consistency rises, plateaus, and then declines. We therefore propose Commit Forcing, a training-free rule that organizes committing into a chain. Each committing pass reads only the preceding block's stored KV, while denoising keeps the base historical context. Commit Forcing improves 60-second VBench Quality and Total on all 6 generators, by up to 1.27 and 1.10 points. Its consistency gains extend to 240-second videos, and it achieves an 82.9\% win rate in overall human preference on 120-second LongLive videos.

\end{abstract}

\begingroup
\setlength{\parskip}{0pt}
\captionsetup{font={small,stretch=1},skip=3pt}
\noindent\begin{minipage}{\linewidth}
\centering
\includegraphics[width=\linewidth]{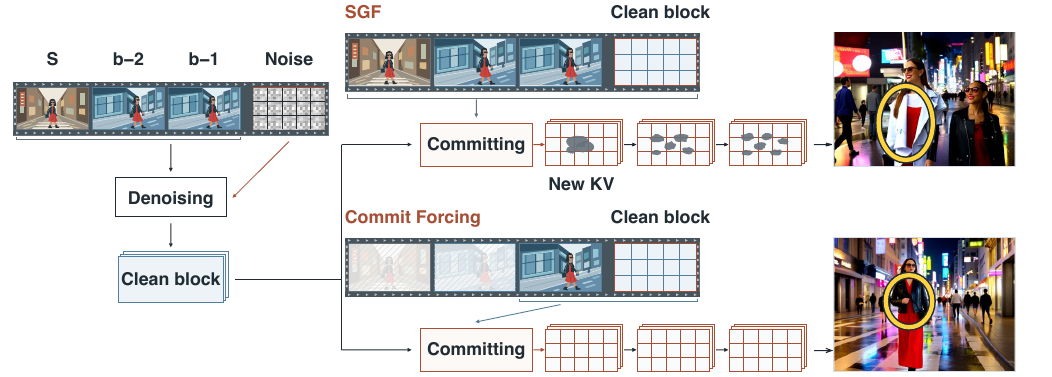}
\captionof{figure}{SGF and Commit Forcing use the same denoising policy but different historical KV during committing. $S$ denotes sink KV, and $b-2$ and $b-1$ index the two preceding blocks. Gray marks depict dependencies from direct reads of older historical context, and the two later KV entries belong to subsequent blocks. Right: SGF frames at 238 seconds, far beyond its 5-second training rollouts, from free generation under each policy, with jacket changes circled.}
\label{fig:history-phase}
\end{minipage}\par
\endgroup

\clearpage
\begingroup
\setlength{\parskip}{0pt}
\reportcontents
\endgroup
\hypertarget{introduction}{%
\section{Introduction}\label{introduction}}

Autoregressive video generators distilled from a bidirectional diffusion teacher~\citep{wan2025wan} produce long videos block by block~\citep{yin2025causvid,huang2025selfforcing}. They reuse historical context through a key-value (KV) cache, which reduces repeated computation over past frames and serves as the long-term memory for later blocks~\citep{gao2025ca2vdm}. Training covers a bounded horizon, such as the 5-second rollouts of Self-Forcing (SF)~\citep{huang2025selfforcing}, whereas inference can extend far beyond it, causing temporal consistency to gradually decrease~\citep{li2026rollingsink}. To reduce this training-inference mismatch, some methods train on longer rollouts~\citep{cui2025selfforcingpp,chen2026contextforcing}. Others select which historical context to keep at inference. SF keeps a recent window, and LongLive and Rolling Sink also retain or replay initial-frame context as a persistent \emph{sink}~\citep{yang2026longlive,li2026rollingsink}. LongLive reads the same sink during training to match inference. These methods all aim to align training and inference. However, attention cost grows quadratically with rollout length, so this alignment becomes increasingly costly, and inference can always extend beyond any training horizon, where temporal consistency still decreases. Rather than eliminating this mismatch, we ask whether the historical context used in training remains suitable for \mbox{inference}.

Each block reads historical context in two stages. A few denoising steps generate the block from cached historical context, and a separate forward pass then writes the clean block's KV into the cache~\citep{huang2025selfforcing}. We refer to these stages as \textbf{denoising} and \textbf{committing}, and prior work also calls the second one cache writing~\citep{gao2025ca2vdm} or KV cache update. Because committing attends to historical context, the same clean block can yield different KV cache entries depending on what committing reads~\citep{shin2026idforcing}. Existing methods choose what committing reads by the role of the sink or by training configurations~\citep{yin2025causvid,huang2025selfforcing,liu2025rollingforcing,shin2026idforcing}. However, they do not systematically study which historical context hurts or helps each stage, or how much of it committing should read.

In this work, we find that historical context useful for denoising is not necessarily useful for committing, whose output conditions later blocks. As the controlled interventions in Table~\ref{tab:longlive-stage-effects} show, older historical context that helps denoising, such as the sink, can hurt later blocks when read during committing, even in generators trained to read it there. At fixed context size and temporal positions, older historical context remains harmful during committing on Self Gradient Forcing (SGF)~\citep{zhuang2026selfgradientforcing} and LongLive. We also show that as committing reads more preceding blocks, consistency rises, plateaus, and then declines on 5 generator configurations. We attribute the rise, which comes with reduced motion, to temporal smoothing during committing, and the decline to harmful older historical \mbox{context}.

Based on these findings, we propose Commit Forcing, a training-free rule that organizes committing into a chain, as illustrated in Figure~\ref{fig:history-phase}. Each committing pass reads only the preceding block's stored KV, which can already carry information from earlier blocks. We apply this rule to 6 pretrained generators with denoising unchanged and without additional DiT forward passes. On 60-second VBench, Commit Forcing improves Quality and Total on all 6 generators, by up to 1.27 and 1.10 points. Clip-to-clip Subject Consistency gains extend to 240-second videos, reaching 5.42 points on SGF. In a blinded 120-second comparison on LongLive, Commit Forcing achieves win rates of 83.5\% for subject and scene consistency and 82.9\% for overall preference, excluding ties.

\noindent Our contributions are as follows:
\nopagebreak[3]
\begin{itemize}[topsep=2pt,partopsep=0pt,parsep=0pt,itemsep=0pt]
\item We show that historical context can have opposing effects in the two stages. While older historical context such as the sink helps denoising, it can hurt later blocks when read during committing, even in generators trained to read it there.
\item We show that as the committing window widens, consistency can rise, plateau, and then decline. We attribute the rise to temporal smoothing through the preceding block, and the decline to harmful older historical context.
\item We propose Commit Forcing, a training-free rule that forms a chain by reading only the preceding block's stored KV during committing and leaves denoising unchanged. It improves Quality and Total on all 6 generators by up to 1.27 and 1.10 points and extends consistency gains to 240-second videos.
\end{itemize}

\hypertarget{related-work}{%
\section{Related Work}\label{related-work}}

\paragraph{Autoregressive video training.}
Teacher forcing trains autoregressive models on clean reference context~\citep{huang2025selfforcing}. Diffusion Forcing adds per-frame noise to this context to bring training closer to autoregressive inference~\citep{chen2024diffusionforcing}. However, reference context, even when noised, differs from the model's own past outputs during autoregressive rollout, creating \emph{exposure bias}~\citep{ranzato2016sequence}. Self-Forcing reduces exposure bias by training on self-generated rollouts but still distills from a bidirectional teacher~\citep{huang2025selfforcing}. Causal Forcing uses an autoregressive teacher for ODE initialization, addressing the architectural mismatch between teacher and student~\citep{zhu2026causalforcing}.

\paragraph{Retaining historical context.}
Caching historical context as clean KV avoids recomputing past-frame representations at each denoising step~\citep{gao2025ca2vdm}. For rollouts beyond the teacher's temporal horizon, Context Forcing (CF) extends the teacher's context to supervise a long-context student~\citep{chen2026contextforcing}, whereas Self-Forcing++ supervises short segments sampled from long rollouts~\citep{cui2025selfforcingpp}. To limit cache size as videos grow longer, LongLive, Rolling Sink, and CF combine recent windows with persistent or refreshed sinks~\citep{yang2026longlive,li2026rollingsink,chen2026contextforcing}. Other methods keep or retrieve selected older historical context beyond the recent window~\citep{layerrecall2026router,xu2026recapforcing,zhao2026relaxforcing,meng2026tethercache}, or improve memory efficiency~\citep{tian2026headforcing,luo2026futureforcing,ji2026forcingkv,li2026arl2}.

\paragraph{Historical context during denoising and committing.}
In CausVid and Self-Forcing, committing reads the same historical context as denoising~\citep{yin2025causvid,huang2025selfforcing}. Some later methods change what each stage reads. Rolling Forcing reads the sink during denoising but not during committing in training~\citep{liu2025rollingforcing}, and ID-Forcing uses self-only committing, which reads no historical KV, to match training configurations~\citep{shin2026idforcing}. Recency Forcing (RF) reweights historical attention during denoising~\citep{cao2026recencyforcing}, and TetherMem routes it by memory age without retraining~\citep{li2026tethermem}. Others train committing directly. SGF backpropagates losses from later blocks through committing~\citep{zhuang2026selfgradientforcing}, SGF+ separates the parameters of the two stages~\citep{su2026sgfplus}, and Uncertainty DMD perturbs clean blocks before committing~\citep{duan2026uncertaintydmd}.

\begingroup
\setlength{\parskip}{2pt plus .5pt minus .5pt}
\setlength{\abovedisplayskip}{4pt plus 1pt minus 1pt}
\setlength{\belowdisplayskip}{4pt plus 1pt minus 1pt}
\setlength{\abovedisplayshortskip}{2pt plus 1pt minus 1pt}
\setlength{\belowdisplayshortskip}{2pt plus 1pt minus 1pt}
\setlength{\textfloatsep}{4pt plus 1pt minus 1pt}
\setlength{\intextsep}{4pt plus 1pt minus 1pt}
\captionsetup{skip=2pt}
\hypertarget{history-mechanisms}{%
\section{Mechanistic Analysis}\label{history-mechanisms}}

To characterize which historical context committing should read, we change what each stage reads in pretrained generators and keep all weights fixed. We compare the effects of the sink in denoising and committing, and vary the source, position, recency, and amount of the context read during committing.

\paragraph{Formulation.}
Let \(x_b\) be the clean latent block at position \(b\), \(z_b\) its noise inputs, \(C_{b-1}\) the KV cache after block \(b-1\), and \(\theta\) the model parameters. We treat the cache as the state carried between blocks. \textbf{Denoising} reads it to generate the current block, and \textbf{committing} constructs a clean KV entry for the block at noise level \(t=0\):
\begin{equation}
\begin{aligned}
x_b&=\mathcal{F}^{\mathrm{d}}_{\theta}(z_b,C_{b-1};D_b),\\
\mathrm{KV}_b&=\mathcal{F}^{\mathrm{c}}_{\theta}\bigl(x_b;\{\mathrm{KV}_j\}_{j\in R_b}\bigr).
\end{aligned}
\label{eq:history-committing}
\end{equation}
\(D_b\) selects the historical context read during denoising, and \(R_b\) indexes the stored entries read during committing. Self-only committing~\citep{shin2026idforcing} uses \(R_b=\varnothing\), and base committing with a sink reads the sink entry \(S\) of block 0 together with recent entries. The base generator then stores \(\mathrm{KV}_b\) in the cache.

\paragraph{Evaluation.}
For paired 48-second videos from LongLive, SGF, and Rolling Sink~\citep{li2026rollingsink}, we compute Consistency, Drift, and Motion from the per-video scores $c_i$, $d_i$, and $r_i$:
\begin{equation}
c_i=\langle u_i^{\mathrm L},u_i^{\mathrm O}\rangle
-\frac{1}{|\mathcal J_i|}\sum\nolimits_{j\in\mathcal J_i}\langle u_i^{\mathrm L},u_j^{\mathrm O}\rangle,
\label{eq:diagnostic-consistency}
\end{equation}
\begin{equation}
\begin{aligned}
d_i &= \bigl|s_i^{\mathrm L}/s_i^{\mathrm E}-1\bigr|,\\
r_i &= m_i^{\mathrm L}/m_i^{\mathrm E}.
\end{aligned}
\label{eq:diagnostic-ratios}
\end{equation}
For video $i$, $u_i^{\mathrm O}$ and $u_i^{\mathrm L}$ are normalized DINOv2 features of its opening and late segments, and $\mathcal J_i$ contains videos generated from the same prompt and condition with other seeds. Consistency thus compares the late segment with the video's own opening against the openings of other videos. Drift compares mean saturation $s$ between the early and late segments, and Motion compares their mean absolute RGB differences $m$ between adjacent frames. We also evaluate 60-second videos on prompts from Rolling Forcing's 200-prompt list~\citep{liu2025rollingforcing} with VBench-Long~\citep{huang2025vbenchpp}, including clip-to-clip Subject Consistency. Appendices~\ref{questions-tested} and~\ref{app:benchmark-scoring} describe extraction, time windows, calibration, and confidence intervals.

\hypertarget{writer-mechanism-probes}{%
\subsection{Opposing Effects}\label{writer-mechanism-probes}}

As shown in Table~\ref{tab:longlive-stage-effects}, reading the sink on LongLive and SGF reduces Drift during denoising but increases it during committing. Excluding the sink from committing lowers Drift from 0.051 to 0.033 on LongLive and from 0.034 to 0.010 on SGF. Excluding it from denoising instead raises Drift above 0.25 on both and lowers Consistency to 0.0116 and 0.0026. We independently vary whether each stage reads the sink. After warm-up, each stage has a 12-frame context on LongLive and a 9-frame context on SGF, including the current 3-frame block. We replace excluded sink KV with KV from three recent trajectory frames and keep sink KV stored.

\begin{table}[!htbp]
\centering
\begin{minipage}{0.65\linewidth}
\centering
\footnotesize
\renewcommand{\arraystretch}{1.0}
\setlength{\tabcolsep}{7pt}
\caption{Effects of reading the sink during denoising and committing on LongLive and SGF. The first row keeps the training-time choice of reading the sink in both stages. $^{*}$Significant difference from reading the sink in both stages.}
\label{tab:longlive-stage-effects}
\begin{tabular*}{\linewidth}{@{\extracolsep{\fill}}cccccc@{}}
\reporttoprule
\multicolumn{2}{c}{\textbf{Reads sink}} & \multicolumn{2}{c}{\textbf{LongLive}} & \multicolumn{2}{c}{\textbf{SGF}} \\
\cmidrule(lr){1-2}\cmidrule(lr){3-4}\cmidrule(lr){5-6}
\textbf{Denoising} & \textbf{Committing} & \textbf{Consistency} $\uparrow$ & \textbf{Drift} $\downarrow$ & \textbf{Consistency} $\uparrow$ & \textbf{Drift} $\downarrow$ \\
\midrule
$\checkmark$ & $\checkmark$ & 0.0970 & 0.051 & 0.1399 & 0.034 \\
$\checkmark$ & $\times$ & 0.0889 & 0.033\sigmark & 0.1592\sigmark & 0.010\sigmark \\
$\times$ & $\checkmark$ & 0.0116\sigmark & 0.257\sigmark & 0.0026\sigmark & 0.253\sigmark \\
$\times$ & $\times$ & 0.0293\sigmark & 0.133\sigmark & 0.0072\sigmark & 0.160\sigmark \\
\bottomrule
\end{tabular*}
\end{minipage}
\end{table}

\endgroup

With denoising unchanged, excluding the sink from committing improves Consistency on SGF, with a nonsignificant change on LongLive. ID-Forcing attributes drift to out-of-distribution KV in the cache and selects committing contexts to match training~\citep{shin2026idforcing}. LongLive and SGF already read the sink during committing in training, and excluding it at inference still reduces Drift. A training--inference mismatch in sink access does not explain this result. With a 60-second training horizon, LongLive already shows these opposing effects in 48-second videos. On SGF, we keep separate caches for denoising and for later committing. Excluding the sink only from the committing pass that writes to the denoising cache improves clip-to-clip Subject Consistency by 3.89 points, close to the 3.98-point gain from excluding it from both passes. The full results are in Appendix~\ref{app:committing-pathways}. We next test whether sink KV still increases Drift at recent positions.

\begingroup
\setlength{\parskip}{2pt plus .5pt minus .5pt}
\setlength{\textfloatsep}{4pt plus 1pt minus 1pt}
\setlength{\intextsep}{4pt plus 1pt minus 1pt}
\captionsetup{skip=2pt}
\subsection{Source Effects}
\label{history-provenance-effects}

On LongLive, we vary the source and assigned positions of three historical frames read during committing, keeping context size and denoising fixed. Table~\ref{tab:committing-source-position} shows that replacing recent KV with sink KV at the same recent positions raises Drift from 0.033 to 0.102. At initial positions, the corresponding means are 0.040 and 0.051.

We also compare KV sources at fixed context sizes on SGF and Rolling Sink. Replacing sink KV with recent KV in a 9-frame context on SGF improves Quality by 1.25 points and clip-to-clip Subject Consistency by 4.54 points, as shown in Table~\ref{tab:sgf-committing-controls}, and replacing replayed KV with recent KV on Rolling Sink raises Consistency from 0.1434 to 0.1540 and the geometric mean of the late-to-early motion ratio by 18.3\%, as reported in Appendix~\ref{app:rolling-sink-source}.

\begin{table}[!htbp]
\centering
\begin{minipage}{0.65\linewidth}
\footnotesize
\renewcommand{\arraystretch}{1.0}
\setlength{\tabcolsep}{8pt}
\caption{Committing on LongLive with KV from three historical frames of different sources and temporal positions. $^{*}$Significant difference from recent KV at recent positions. Intervals appear in Appendix~\ref{app:content-position-controls}.}
\label{tab:committing-source-position}
\begin{tabular*}{\linewidth}{@{\extracolsep{\fill}}cccc@{}}
\reporttoprule
\textbf{Source} & \textbf{Position} & \textbf{Consistency} $\uparrow$ & \textbf{Drift} $\downarrow$ \\
\midrule
Recent & Recent & 0.0887 & 0.033 \\
Recent & Initial & 0.0928 & 0.040 \\
\midrule
Sink & Initial & 0.0971 & 0.051\sigmark \\
Sink & Recent & 0.0723\sigmark & 0.102\sigmark \\
\bottomrule
\end{tabular*}
\end{minipage}
\end{table}

\endgroup

\begingroup
\setlength{\textfloatsep}{5pt}
\subsection{Recency Trade-off}
\label{content-age-during-committing}

We measure recency by the lag $d$, the number of blocks back in the original trajectory. During committing, we use the three current frames and KV from three frames at lag $d\in\{1,2,4,8,16\}$, with all assigned temporal positions fixed. Figure~\ref{fig:committing-content-age} shows that increasing $d$ from 1 to 16 significantly reduces mean clip-to-clip Subject Consistency by 0.69 points on SGF and 2.46 points on LongLive.

\begingroup
\setlength{\parskip}{0pt}
\ifdefined\nolinenumbers\nolinenumbers\fi
\noindent\makebox[\linewidth][c]{%
\begin{minipage}{.65\linewidth}
\centering
\includegraphics[trim=0 4bp 0 0,clip,width=.846153846\linewidth]{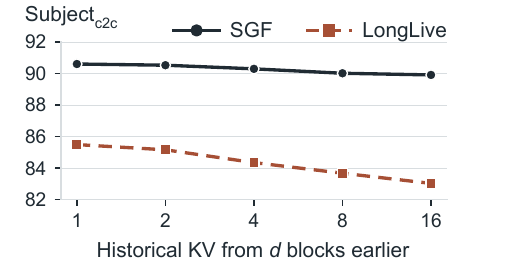}
\captionsetup{skip=0pt}
\captionof{figure}{Mean clip-to-clip Subject Consistency (0--100) when committing reads historical KV from $d$ blocks earlier.}
\label{fig:committing-content-age}
\end{minipage}}\par
\endgroup

\endgroup

Dynamic Degree increases by 5.03 points on SGF and 3.57 points on LongLive, and Quality changes by less than 0.25 points. On SGF, we also rebuild the cache from the same clean latents for each lag and generate the next block at five positions. Opening similarity also decreases significantly by 0.0010 between the shortest and longest lags. The full results are in Appendix~\ref{app:committing-recency-fixed-trajectory}.

\subsection{Committing Window}
\label{committing-tradeoff-limits}

To test how the committing window affects consistency on LongLive, we keep denoising fixed and vary the number of preceding blocks read from zero to three. The left panel of Figure~\ref{fig:committing-window-mechanisms} shows mean clip-to-clip Subject Consistency rising from 86.21 with zero preceding blocks to 87.21 with one, staying at 87.23 with two, and falling to 86.93 with three. Committing with one preceding block exceeds the base by 1.20 points.

With one preceding block, we observe higher consistency and a 3.63-point drop in Dynamic Degree relative to self-only committing. To weaken the preceding block's contribution, we replace the change it causes in the attention output at each position with its mean within the frame. Compared with the usual one-block pass, this control lowers consistency by 0.71 points and raises Dynamic Degree by 2.30 points. The two effects are consistent with temporal smoothing during committing. Table~\ref{tab:committing-averaging-controls} reports the averaging controls.

The second preceding block adds only 0.02 points of consistency. The drop with three preceding blocks is consistent with the effect of reading older KV at a fixed frame count in Section~\ref{content-age-during-committing}.

To measure how long a one-time change in committing persists under each window, we replace the oldest historical input with sink KV for one SGF pass with clean latents fixed. We then resume each setting's usual policy. The right panel of Figure~\ref{fig:committing-window-mechanisms} shows that the value-tensor difference is zero from lag 29 with one preceding block and persists at lag 32 with two.\par

\begingroup
\ifdefined\nolinenumbers\nolinenumbers\fi
\noindent\begin{minipage}{\linewidth}
\centering
\includegraphics[width=\linewidth]{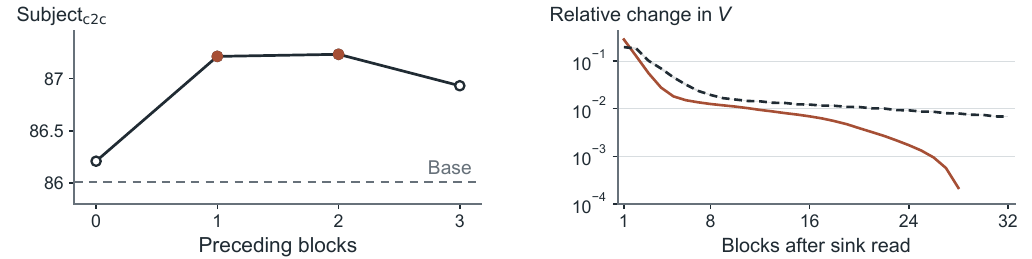}
\captionof{figure}{The left panel plots LongLive clip-to-clip Subject Consistency. Its dashed baseline reads the sink and two preceding blocks. The right panel plots SGF's layer-mean relative $L_2$ change in the value tensor by block after one sink read with fixed clean latents. Solid and dashed curves use one and two preceding blocks, respectively. Bands show paired 95\% bootstrap intervals over 20 prompts.}
\label{fig:committing-window-mechanisms}
\end{minipage}\par
\endgroup


\hypertarget{context-access-at-kv-commit}{%
\section{Commit Forcing}\label{context-access-at-kv-commit}}

We propose Commit Forcing, a training-free rule for long video generation. As shown in Figure~\ref{fig:kv_commit}, we keep the base denoising policy and use the preceding block's stored KV as the sole historical input to committing.

For $b\ge 1$, the committing rule in Eq.~\ref{eq:history-committing} becomes
\begin{equation}
\begin{aligned}
R_b &= \{b-1\},\\
\mathrm{KV}_b &= \mathcal{F}^{\mathrm c}_{\theta}\bigl(x_b;\{\mathrm{KV}_{b-1}\}\bigr).
\end{aligned}
\label{eq:kv-chain}
\end{equation}
The first block has no historical input. The 6 generators in our main evaluation use three latent frames per block, giving a six-frame committing window.

\smallskip
\noindent\begin{minipage}{\linewidth}
\centering
\includegraphics[width=.95\textwidth]{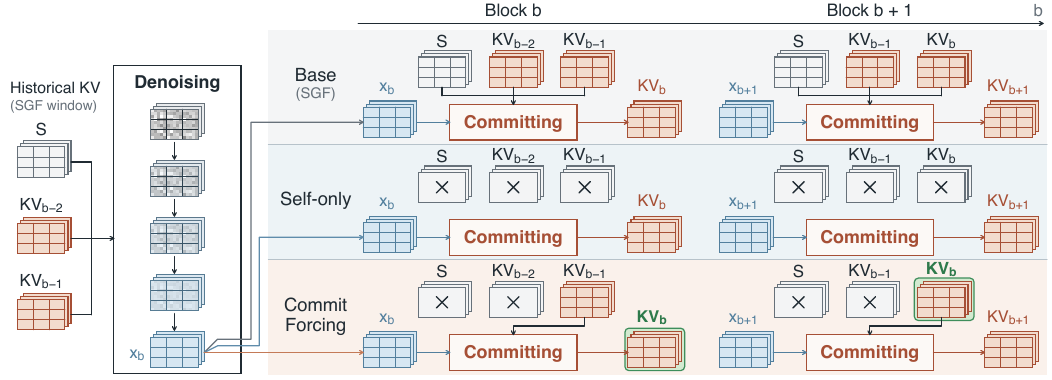}
\captionof{figure}{Denoising keeps the base historical context. Blue marks clean blocks and orange marks KV. Base (SGF) reads sink $S$ and two preceding entries. Self-only reads no historical KV, and Commit Forcing reads only the preceding entry. Crosses mark stored historical KV not read during committing. Green marks $\mathrm{KV}_b$, which Commit Forcing reads when committing block $b+1$.}
\label{fig:kv_commit}
\end{minipage}\par
\smallskip

Earlier blocks can still influence the new KV through the preceding block's stored entry. For example, $\mathrm{KV}_{b-1}$ was constructed by reading $\mathrm{KV}_{b-2}$. Directly reading the two preceding entries would add a second path from $b-2$ to $b$. The indirect path may help explain the small gain from a second direct read on LongLive. A sink read during committing directly connects block~0 to the new KV. Under Commit Forcing, information can travel through successive entries even after the entry that originally carried it leaves the cache.

With clean latents, prompt, and positions fixed, network depth bounds this indirect dependence. As keys and values are projected before attention, layer 0 of $\mathrm{KV}_b$ depends on $x_b$. Layer 1 can also depend on layer 0 of $\mathrm{KV}_{b-1}$, which depends on $x_{b-1}$. Each further step into the past requires another layer of computation. Thus, layer $\ell$ can depend on blocks from $\max(0,b-\ell)$ to $b$. Reading two preceding blocks extends this bound to $\max(0,b-2\ell)$. Appendix~\ref{app:committing-recurrence} gives the derivation. In free generation, the base denoising policy can also carry older context into $x_b$ and hence into the new KV.

\hypertarget{experiments}{%
\section{Experiments}\label{experiments}\label{results}}

\begingroup
\setlength{\extrarowheight}{1.35pt}
\begin{table}[!htbp]
\centering
\footnotesize
\setlength{\tabcolsep}{3pt}
\renewcommand{\arraystretch}{1.2}
\caption{Quality, Semantic, and Total of each base configuration and Commit Forcing on 60-second VBench. Rows are ordered by Total with Commit Forcing. Paired intervals appear in Tables~\ref{tab:standard-main-intervals} and~\ref{tab:standard-semantic-aggregate}, and $^{*}$marks an interval excluding 0. RF-TF denotes the training-free RF reproduction.}
\label{tab:standard-main}
\begin{tabular*}{\textwidth}{@{\extracolsep{\fill}}lccccccccc@{\hspace{4pt}}}
\reporttoprule
& \multicolumn{3}{>{\columncolor{LovartAccent!8}[0pt]}c}{\textbf{Quality}} & \multicolumn{3}{>{\columncolor{LovartAccent!8}[0pt]}c}{\textbf{Semantic}} & \multicolumn{3}{>{\columncolor{LovartAccent!8}[0pt]}c}{\textbf{Total}} \\
\cmidrule(lr){2-4}\cmidrule(lr){5-7}\cmidrule(lr){8-10}
\textbf{Configuration} & \textbf{Base} & \textbf{Commit Forcing} & $\boldsymbol{\Delta}$ & \textbf{Base} & \textbf{Commit Forcing} & $\boldsymbol{\Delta}$ & \textbf{Base} & \textbf{Commit Forcing} & $\boldsymbol{\Delta}$ \\
\midrule
CF & 84.79 & 85.25 & +0.46\sigmark & 72.03 & 72.41 & +0.38 & 82.24 & 82.68 & +0.44\sigmark \\
Rolling Sink & 84.84 & 85.25 & +0.41\sigmark & 75.13 & 75.31 & +0.17 & 82.90 & 83.26 & +0.36\sigmark \\
RF-TF & 84.77 & 85.42 & +0.65\sigmark & 72.10 & 75.00 & +2.90\sigmark & 82.24 & 83.34 & +1.10\sigmark \\
LongLive & 84.70 & 84.99 & +0.29\sigmark & 76.43 & 76.84 & +0.41 & 83.05 & 83.36 & +0.31\sigmark \\
SGF+ & 86.32 & 86.63 & +0.32 & 76.22 & 76.66 & +0.45 & 84.30 & 84.64 & +0.34 \\
SGF & 86.06 & 87.33 & +1.27\sigmark & 74.99 & 74.78 & \textminus{}0.21 & 83.85 & 84.82 & +0.98\sigmark \\
\bottomrule
\end{tabular*}
\end{table}

\endgroup

\hypertarget{experimental-protocol}{%
\subsection{Experimental Setup}\label{experimental-protocol}}

\hypertarget{models-and-paired-generation}{%
\paragraph{Generators.}\label{models-and-paired-generation}}
\hypertarget{scope-of-completed-evaluations}{}\label{scope-of-completed-evaluations}\label{discussion}
We evaluate SGF, SGF+, CF, LongLive, Rolling Sink, and a training-free RF reproduction (RF-TF), as listed in Table~\ref{tab:standard-main}. All 6 are Wan-based and use persistent KV with an explicit committing pass. We apply Commit Forcing to all 6, so committing reads the current block and the preceding block's stored KV, 6 latent frames in total. Within each base-generator comparison, we use the same denoising and sampler settings and pair outputs by prompt. We also compare with ID-Forcing on SF and LongLive. Appendix~\ref{app:model-configurations} details contexts, noise pairing, and baseline sources.

\hypertarget{vbench-long-scoring}{%
\paragraph{Evaluation.}\label{vbench-long-scoring}}
On the standard VBench suite~\citep{huang2024vbench}, which covers 16 dimensions, we generate 946 videos of 60 seconds per condition and report official VBench-Long Quality, Semantic, and Total scores~\citep{huang2025vbenchpp}. On the 200 prompts selected in Rolling Forcing~\citep{liu2025rollingforcing}, we evaluate the same generators and SF with a sink at 60 seconds. We also test selected generators at 120 and 240 seconds. The main evaluations use 16 fps and 832 $\times$ 480 resolution. For the other prompt sets, we compute Quality from its 7 dimensions with VBench's published normalization and weights, and report it separately. We measure long-range consistency with the clip-to-clip component of VBench-Long Subject Consistency. Appendix~\ref{app:evaluation-protocols} gives scoring, statistical, and protocol details.

\hypertarget{standard-suite-results}{%
\subsection{Main Results}\label{standard-suite-results}}

We compare Commit Forcing with each of the 6 base generators on the 60-second standard VBench suite, as shown in Table~\ref{tab:standard-main}. Quality and Total rise on all 6 generators, by 0.57 and 0.59 points on average, and both increases are significant on 5 of them. The largest gains are 1.27 Quality points on SGF and 1.10 Total points on RF-TF, which also gains significantly in Semantic score. SGF already trains committing by backpropagating losses from later blocks through it, and the gains on SGF show that this training does not remove the benefit of changing what committing reads. Appendix~\ref{app:standard-components} gives component scores and the Quality decomposition.

\phantomsection
\hypertarget{current-rollf200-results}{}\label{current-rollf200-results}
On the Rolling Forcing prompt selection, Commit Forcing improves clip-to-clip Subject Consistency significantly on all 6 generators and on SF with a sink, while Quality gains vary by configuration. For each generator, we regress the paired change in consistency on the paired change in Dynamic Degree. On SGF, the raw gain is 4.76 points and the fitted gain at zero change in Dynamic Degree is 4.87 points. Table~\ref{tab:matched-motion} shows significant adjusted gains on all 6 generators.

\begingroup
\setlength{\parskip}{0pt}
\noindent\makebox[\linewidth][c]{%
\begin{minipage}{.65\linewidth}
\centering
\footnotesize
\setlength{\tabcolsep}{8pt}
\renewcommand{\arraystretch}{1.08}
\captionof{table}{Clip-to-clip Subject Consistency gains over base on 200 Rolling Forcing prompts at 60 seconds with generation seed 0. Adjusted gains account for Dynamic Degree (DD).}
\label{tab:matched-motion}
\begin{tabular*}{\linewidth}{@{\extracolsep{\fill}}lccc@{}}
\reporttoprule
\textbf{Generator} & $\boldsymbol{\Delta}$\textbf{DD} & \textbf{Raw} & \textbf{Adjusted [95\% CI]} \\
\midrule
SGF & +11.33 & +4.76 & +4.87\enspace [+3.89,+5.95] \\
SGF+ & +3.22 & +2.76 & +2.78\enspace [+1.89,+3.64] \\
RF-TF & \textminus{}3.92 & +6.36 & +5.99\enspace [+5.09,+6.84] \\
Rolling Sink & +2.02 & +2.45 & +2.52\enspace [+1.83,+3.17] \\
LongLive & +6.42 & +1.20 & +1.38\enspace [+0.95,+1.84] \\
CF & \textminus{}1.05 & +3.23 & +3.20\enspace [+2.55,+3.89] \\
\bottomrule
\end{tabular*}
\end{minipage}}\par

\noindent\makebox[\linewidth][c]{%
\begin{minipage}{.65\linewidth}
\centering
\footnotesize
\renewcommand{\arraystretch}{1.08}
\setlength{\tabcolsep}{4pt}
\captionof{table}{Commit Forcing win rates on LongLive in blinded pairwise comparisons, excluding ties. Bold marks win rates above 50\%.}
\label{tab:human-evaluation}
\begin{tabular*}{\linewidth}{@{\extracolsep{\fill}}lccc@{}}
\reporttoprule
\textbf{Baseline} & \textbf{Duration} & \textbf{Subject and scene} & \textbf{Overall preference} \\
\midrule
Base LongLive & 60 s & 43.2\% & 45.5\% \\
Base LongLive & 120 s & \textbf{83.5\%} & \textbf{82.9\%} \\
Complete ID-Forcing & 60 s & \textbf{90.4\%} & \textbf{88.2\%} \\
\bottomrule
\end{tabular*}
\end{minipage}}\par

\endgroup

\paragraph{Human evaluation.}
Participants view three synchronized excerpts per prompt pair, with method identities hidden and left-right positions randomized. They judge subject and scene consistency with the opening and give an overall preference. At 60 seconds, each comparison uses 18 pairs from the Rolling Forcing prompt selection. In Table~\ref{tab:human-evaluation}, Commit Forcing wins 88.2\% of non-tied overall judgments against complete ID-Forcing and 45.5\% against base LongLive.

At 120 seconds, 5 participants independently judge 30 pairs from ID-Forcing's 128-prompt MovieGen list. Against base LongLive, Commit Forcing wins 83.5\% for subject and scene consistency and 82.9\% for overall preference (Appendix~\ref{app:human-evaluation}).\par

\hypertarget{context-forcing-at-240-seconds-and-memory-ablations}{%
\subsection{Long-Horizon Generation}\label{context-forcing-at-240-seconds-and-memory-ablations}}

We evaluate SGF and CF at 240 seconds on 32 extended MovieGen prompts, and LongLive at 120 and 240 seconds on ID-Forcing's 128-prompt MovieGen list. Table~\ref{tab:long-horizon-main} shows that Commit Forcing improves clip-to-clip Subject Consistency at 240 seconds on all three generators and improves LongLive's Quality at both durations. Figure~\ref{fig:qualitative-grid} shows SGF and CF examples at 240 seconds. Appendix~\ref{app:long-video-components} gives the component scores.

\smallskip
\noindent\begin{minipage}{\linewidth}
\centering
\includegraphics[width=.89\textwidth]{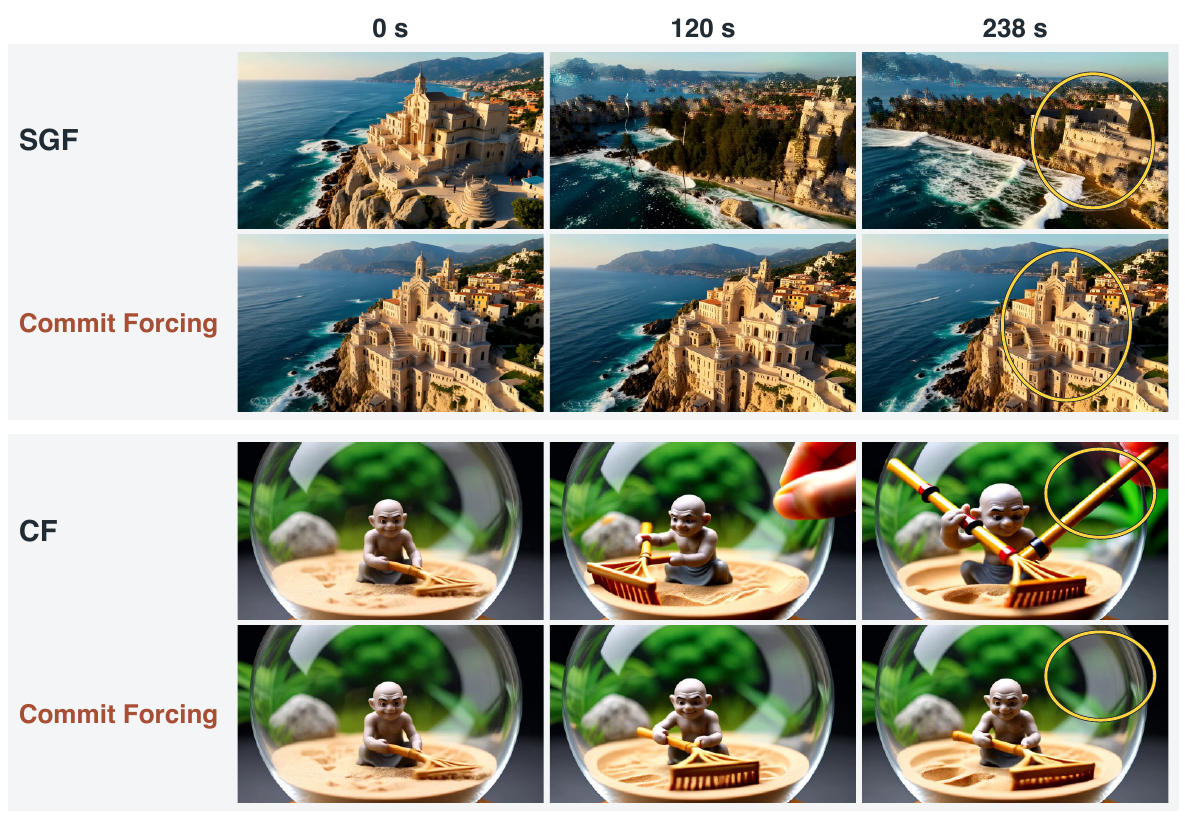}
\captionof{figure}{Base generation (top of each pair) and Commit Forcing use the same prompt and initial noise for each 240-second video. Circles at 238 seconds mark the church that changes in base SGF and the extra tool handle in base CF.}
\label{fig:qualitative-grid}
\end{minipage}\par
\smallskip

\smallskip

\noindent\makebox[\linewidth][c]{%
\begin{minipage}{.65\linewidth}
\centering
\footnotesize
\setlength{\tabcolsep}{8pt}
\renewcommand{\arraystretch}{1.08}
\captionof{table}{Paired gains over base at 120 and 240 seconds. Subject$_{\mathrm{c2c}}$ is clip-to-clip Subject Consistency.}
\label{tab:long-horizon-main}
\begin{tabular*}{\linewidth}{@{\extracolsep{\fill}}lccc@{}}
\reporttoprule
& & \multicolumn{2}{c@{}}{\textbf{Gain [95\% CI]}} \\
\cmidrule(l){3-4}
\textbf{Generator} & \textbf{Time} & \textbf{Quality} & \textbf{Subject}$_{\mathrm{c2c}}$ \\
\midrule
SGF & 240 s & +0.90\enspace [\textminus{}0.002,1.82] & +5.42\enspace [3.02,8.09] \\
CF & 240 s & +0.44\enspace [\textminus{}0.08,0.97] & +2.23\enspace [0.51,3.89] \\
LongLive & 120 s & +0.33\enspace [0.16,0.50] & +1.87\enspace [1.39,2.36] \\
LongLive & 240 s & +0.28\enspace [0.08,0.49] & +2.55\enspace [1.98,3.15] \\
\bottomrule
\end{tabular*}
\end{minipage}}\par

\smallskip

\begingroup
\setlength{\extrarowheight}{1.1pt}
\begin{table}[!htbp]
\centering
\footnotesize
\setlength{\tabcolsep}{0pt}
\renewcommand{\arraystretch}{1.17}
\caption{We compare committing windows on the first 100 Rolling Forcing prompts at 60 seconds with generation seed 0. All scores are $\times100$. Subject$_{\mathrm{c2c}}$ denotes clip-to-clip Subject Consistency. Window length $w$ includes the current 3-frame block, and $w=3$ is self-only committing. Shading marks the 6-frame context used in the main experiments. Base denotes each generator's original committing context.}
\label{tab:committing-window-comparison}
\begin{tabular}{@{}>{\raggedright\arraybackslash}p{.12\linewidth}@{}>{\centering\arraybackslash}p{.09\linewidth}@{}>{\centering\arraybackslash}p{.11\linewidth}@{}>{\centering\arraybackslash}p{.16\linewidth}@{\hspace{.04\linewidth}}>{\raggedright\arraybackslash}p{.12\linewidth}@{}>{\centering\arraybackslash}p{.09\linewidth}@{}>{\centering\arraybackslash}p{.11\linewidth}@{}>{\centering\arraybackslash}p{.16\linewidth}@{}}
\reporttoprule
\textbf{Generator} & \textbf{Window} & \textbf{Subject}$_{\mathrm{c2c}}$ & \mbox{\textbf{Dynamic Degree}} & \textbf{Generator} & \textbf{Window} & \textbf{Subject}$_{\mathrm{c2c}}$ & \mbox{\textbf{Dynamic Degree}} \\
\cmidrule(r){1-4}\cmidrule(l){5-8}
SGF & Base & 86.20 & 58.57 & LongLive & Base & 84.27 & 47.27 \\
 & $w=3$ & 89.96 & 72.47 &  & $w=3$ & 84.60 & 56.53 \\
\reporthighlight
 & $\boldsymbol{w=6}$ & \textbf{90.45} & \textbf{68.40} &  & $\boldsymbol{w=6}$ & \textbf{85.52} & \textbf{54.13} \\
 & $w=9$ & 90.25 & 70.60 &  & $w=9$ & 85.49 & 53.83 \\
 &  &  &  &  & $w=12$ & 85.25 & 53.23 \\
\cmidrule(r){1-4}\cmidrule(l){5-8}
RF-TF & Base & 75.46 & 64.80 & Rolling Sink & Base & 85.07 & 45.93 \\
 & $w=3$ & 78.68 & 67.40 &  & $w=3$ & 86.27 & 54.47 \\
\reporthighlight
 & $\boldsymbol{w=6}$ & \textbf{81.19} & \textbf{63.20} &  & $\boldsymbol{w=6}$ & \textbf{86.86} & \textbf{48.47} \\
 & $w=9$ & 80.89 & 63.37 &  & $w=9$ & 86.58 & 48.13 \\
 & $w=12$ & 80.62 & 64.03 &  & $w=12$ & 86.15 & 48.93 \\
 & $w=21$ & 78.92 & 65.90 &  & $w=21$ & 85.00 & 51.03 \\
\cmidrule(r){1-4}\cmidrule(l){5-8}
SGF+ & $w=3$ & 92.09 & 65.83 & CF & $w=3$ & 80.58 & 65.93 \\
\reporthighlight
 & $\boldsymbol{w=6}$ & \textbf{91.67} & \textbf{62.60} &  & $\boldsymbol{w=6}$ & \textbf{80.63} & \textbf{64.90} \\
 & $w=9$ & 91.27 & 63.30 &  & & & \\
\bottomrule
\end{tabular}
\end{table}

\endgroup

\hypertarget{id-forcing-context-decomposition}{%
\subsection{Ablations and Trade-offs}\label{id-forcing-context-decomposition}\label{consistency-motion-tradeoffs}}

\paragraph{Comparison with self-only committing.}To test whether reading the preceding block is needed, we compare Commit Forcing with self-only committing, which reads no historical KV. Commit Forcing improves clip-to-clip Subject Consistency by 4.93 points on SF with a sink and by 2.51 points on RF-TF, where Quality also rises by 0.24 points. On LongLive, the gain is significant at all three durations, 1.00 to 1.18 points at 60 seconds and 1.30 and 1.31 points at 120 and 240 seconds. On the LongLive standard suite, Commit Forcing also scores significantly higher in 5 of the 7 Quality dimensions, including Subject and Background Consistency. At 120 and 240 seconds, LongLive shows lower Dynamic Degree and higher Motion Smoothness, consistent with the temporal smoothing in Section~\ref{committing-tradeoff-limits}. Quality falls by 0.17 and 0.21 points at those durations. Appendix~\ref{app:self-only-contrasts} shows no significant consistency gain on SGF, SGF+, or CF.

\paragraph{Committing-window effects.}
To test whether the window pattern in Section~\ref{committing-tradeoff-limits} holds on other generators, we vary their committing windows with denoising unchanged, as shown in Table~\ref{tab:committing-window-comparison}. On the first 100 prompts of the Rolling Forcing selection, shortening the 21-frame contexts of RF-TF and Rolling Sink to 6 frames improves clip-to-clip Subject Consistency by 2.27 and 1.85 points, while Dynamic Degree decreases on both, significantly only on RF-TF. Denser window scans show consistency plateaus at 6--12 and 6--9 frames, respectively. Across the 6 generators in Table~\ref{tab:committing-window-comparison} and the pooled SF-with-sink comparison, the 6-frame window of Commit Forcing is within 0.51 points of the best tested window, and the largest gap occurs on SF with a sink. Appendix~\ref{app:committing-tables} reports the full window results.

\paragraph{Comparison with complete ID-Forcing.}
Within the ID-Forcing framework, we replace its committing context with the current block and KV from the two preceding blocks on SF and LongLive. On SF, this adaptation also changes cache handover and omits tail recaching. At 60 seconds on the Rolling Forcing prompt selection, it improves clip-to-clip Subject Consistency by 2.27 to 2.53 points on SF and 2.05 points on LongLive. Quality rises by 0.20 to 0.23 points on SF and does not change significantly on LongLive. Table~\ref{tab:standard-idf-main} gives its standard-suite results. On LongLive, Commit Forcing exceeds complete ID-Forcing in clip-to-clip Subject Consistency by 4.38 to 4.83 points at 60 seconds and by 7.31 and 8.21 points at 120 and 240 seconds. At the two longer durations, its Dynamic Degree is 11.00 and 12.73 points lower, and Quality does not differ significantly, as shown in Appendices~\ref{app:long-idf-results} and~\ref{app:self-only-contrasts}.

\paragraph{Inference cost.}
\label{computational-cost}
\hypertarget{why-the-pass-matters}{}\label{why-the-pass-matters}

Within the base frameworks, Commit Forcing adds no DiT forward passes. On one H20 node, we time one 60-second video per configuration. Wall time is 2--7\% lower with Commit Forcing across the 6 main generators. Measured peak GPU memory increases by 5.6 GiB on Rolling Sink and is unchanged on the other 5 generators. Appendix~\ref{app:computational-cost} reports the measurements and ID-Forcing timings.


\hypertarget{limitations}{%
\section{Limitations}\label{limitations}}

Our main evaluation covers six Wan-based generators. Related methods such as Self-Forcing and Rolling Forcing use the same backbone~\citep{huang2025selfforcing,liu2025rollingforcing}. Commit Forcing applies to generators that write each block's KV to a persistent cache in a separate committing pass. Models that store historical latents and rebuild their KV for each new block fall outside this setting. WorldPlay~\citep{sun2025worldplay}, for example, jointly re-encodes selected past latents before denoising. Restricting access to older context in this re-encoding pass worsens block transitions in our tests, as reported in Appendix~\ref{app:transfer-boundaries}. For SF with a sink, the consistency plateau starts at two preceding blocks, as shown in Appendix~\ref{app:committing-tables}.

\hypertarget{conclusion}{%
\section{Conclusion}\label{conclusion}}

Historical context that helps denoise the current block can increase drift in later blocks when committing reads it. At a fixed frame count, reading older KV during committing lowers consistency. On LongLive, reading one preceding block during committing is enough to reach the consistency plateau. This motivates Commit Forcing, a training-free rule that keeps the base denoising policy and uses the preceding block's stored KV as the only historical input to committing. The rule improves VBench Quality and Total on all 6 generators at 60 seconds and yields consistency gains in the 240-second evaluations.

\clearpage
\phantomsection
\addcontentsline{toc}{section}{References}
\renewcommand{\bibfont}{\footnotesize\linespread{0.97}\selectfont}
\setlength{\bibsep}{0pt}
\bibliographystyle{plainnat}
\bibliography{references}

\clearpage
\appendix
\counterwithin{figure}{section}
\counterwithin{table}{section}
\section{Mechanistic Analysis Details}
\label{app:mechanism-details}

\subsection{Mechanism-study setup and diagnostics}

\paragraph{Diagnostic readouts.}
\phantomsection
\hypertarget{questions-tested}{}\label{questions-tested}
We use 48-second rollouts for the LongLive and SGF stage interventions, the Rolling Sink source intervention in Section~\ref{history-mechanisms}, and the SF switching intervention in Section~\ref{app:committing-exposure}. We report their diagnostic readouts as Consistency, Drift, and Motion.

\noindent\textbf{Consistency.} We compute Eq.~\ref{eq:diagnostic-consistency} from full-frame DINOv2 CLS features over the opening (0.5--2 seconds) and late (38--47.5 seconds) segments. We normalize each frame embedding before temporal averaging and normalize the resulting segment means.

\noindent\textbf{Motion.} We compute mean absolute RGB differences between adjacent decoded frames, multiplied by 255, using uncompressed values in $[0,1]$. We average these differences over 4--8 seconds and 38--47.5 seconds to obtain the early and late motion values \(m_i^{\mathrm E}\) and \(m_i^{\mathrm L}\).

\noindent\textbf{Drift.} We compute mean HSV saturation from the decoded MP4, subsampling every other frame and pixel in both spatial dimensions, and aggregate it into 1-second bins. We average bins 4--7 and 38--47 to obtain early and late saturation values \(s_i^{\mathrm E}\) and \(s_i^{\mathrm L}\). The per-video Drift and Motion scores $d_i$ and $r_i$ are defined in Eq.~\ref{eq:diagnostic-ratios}.
We average \(c_i\) and \(d_i\) across videos. For Motion, we report the median of \(r_i\), where 1 indicates unchanged motion relative to the early segment. We also report the proportion of videos with late-to-early motion ratios below 0.85. Motion decline under this diagnostic does not imply literal freezing, and the diagnostic is distinct from VBench Dynamic Degree.

Figure~\ref{fig:history-effect-summary} uses signed normalized-saturation differences from the 12-frame recent-only control on SF.

\paragraph{Diagnostic calibration.}
\label{app:diagnostic-calibration}
We calibrate the DINOv2 appearance-consistency diagnostic on the 40 SGF videos generated with the 9-frame recent-only committing context while denoising reads the sink. We edit only frames from 37.5 seconds onward and compare each edit with a decode-and-re-encode control. We vary saturation and brightness, freeze the tail, or crop and resize the view. For tail replacement, we use a tail segment from another video generated with the same prompt and a different seed, or from a video generated with a different prompt. Table~\ref{tab:diagnostic-calibration} reports DINOv2 and CLIP ViT-L/14 readouts.

Tail replacement causes larger changes than recoloring or freezing, and cropping also changes the diagnostic. We therefore interpret it as full-frame appearance consistency, including scene and view changes. The DINOv2 frozen-tail interval includes zero.

We repeat the mechanism comparisons with CLIP embeddings and adjust paired DINOv2 changes for Drift, brightness drift, and log relative motion, as reported in Table~\ref{tab:diagnostic-clip-replication}. The adjusted change is the regression intercept, with both seeds retained within each prompt during resampling. On SGF, the consistency loss from denoising exclusion remains under both checks, whereas the committing-only gain is not significant after adjustment. On LongLive, denoising exclusion lowers CLIP consistency, but its adjusted DINOv2 interval includes zero. Excluding the sink from both stages yields a negative adjusted change.

\begin{table}[!htbp]
\centering
\begin{minipage}{0.80\linewidth}
\centering
\footnotesize
\setlength{\tabcolsep}{3pt}
\renewcommand{\arraystretch}{1.1}
\caption{Controlled tail edits assess the appearance-consistency diagnostic.}
\label{tab:diagnostic-calibration}
\begin{tabularx}{\linewidth}{@{}Ycc@{}}
\reporttoprule
\textbf{Tail edit} & $\boldsymbol{\Delta}$\textbf{ DINOv2} & $\boldsymbol{\Delta}$\textbf{ CLIP} \\
\midrule
Saturation $\times1.04$, brightness $\times1.045$ & \shortstack{\textminus{}0.0003\\{}[\textminus{}0.0014,+0.0010]} & \shortstack{\textminus{}0.0002\\{}[\textminus{}0.0011,+0.0006]} \\
Saturation $\times1.20$, brightness $\times1.25$ & \shortstack{\textminus{}0.0008\\{}[\textminus{}0.0048,+0.0047]} & \shortstack{\textminus{}0.0044\sigmark\\{}[\textminus{}0.0071,\textminus{}0.0016]} \\
Frozen tail & \shortstack{\textminus{}0.0039\\{}[\textminus{}0.0155,+0.0080]} & \shortstack{\textminus{}0.0039\sigmark\\{}[\textminus{}0.0075,\textminus{}0.0005]} \\
Central 80\% crop, resized & \shortstack{\textminus{}0.0241\sigmark\\{}[\textminus{}0.0350,\textminus{}0.0135]} & \shortstack{\textminus{}0.0125\sigmark\\{}[\textminus{}0.0177,\textminus{}0.0075]} \\
Tail from another seed of the same prompt & \shortstack{\textminus{}0.3225\sigmark\\{}[\textminus{}0.3825,\textminus{}0.2653]} & \shortstack{\textminus{}0.1339\sigmark\\{}[\textminus{}0.1591,\textminus{}0.1117]} \\
Tail from another prompt & \shortstack{\textminus{}0.1623\sigmark\\{}[\textminus{}0.1944,\textminus{}0.1340]} & \shortstack{\textminus{}0.0646\sigmark\\{}[\textminus{}0.0766,\textminus{}0.0536]} \\
\bottomrule
\end{tabularx}
\smallskip
\parbox{\linewidth}{\footnotesize\raggedright Paired changes from decode-and-re-encode-only videos. The DINOv2 and CLIP reference means are 0.1612 and 0.0667. Both embeddings use the own-minus-other-seed diagnostic. Brackets show 95\% prompt-bootstrap intervals, and $^{*}$marks intervals excluding 0.}
\end{minipage}
\end{table}

\begin{table}[!htbp]
\centering
\footnotesize
\setlength{\tabcolsep}{4pt}
\renewcommand{\arraystretch}{1.12}
\caption{DINOv2 and CLIP readouts for the stage and source controls.}
\label{tab:diagnostic-clip-replication}
\begin{tabular*}{\linewidth}{@{\extracolsep{\fill}}>{\raggedright\arraybackslash}p{.36\linewidth}ccc@{}}
\reporttoprule
\textbf{Configuration} & \shortstack{\textbf{DINOv2}\\$\boldsymbol{\Delta}$ [95\% CI]} & \shortstack{\textbf{CLIP}\\$\boldsymbol{\Delta}$ [95\% CI]} & \shortstack{\textbf{Adjusted DINOv2}\\$\boldsymbol{\Delta}$ [95\% CI]} \\
\midrule
SGF: sink excluded from committing & \shortstack{+0.0194\sigmark\\{}[+0.0073,+0.0310]} & \shortstack{+0.0156\sigmark\\{}[+0.0090,+0.0223]} & \shortstack{+0.0132\\{}[\textminus{}0.0058,+0.0346]} \\
\addlinespace[2pt]
SGF: sink excluded from denoising & \shortstack{\textminus{}0.1373\sigmark\\{}[\textminus{}0.1686,\textminus{}0.1070]} & \shortstack{\textminus{}0.0505\sigmark\\{}[\textminus{}0.0615,\textminus{}0.0398]} & \shortstack{\textminus{}0.1318\sigmark\\{}[\textminus{}0.2004,\textminus{}0.0632]} \\
\addlinespace[2pt]
SGF: sink excluded from both stages & \shortstack{\textminus{}0.1327\sigmark\\{}[\textminus{}0.1620,\textminus{}0.1043]} & \shortstack{\textminus{}0.0521\sigmark\\{}[\textminus{}0.0636,\textminus{}0.0409]} & \shortstack{\textminus{}0.1280\sigmark\\{}[\textminus{}0.1856,\textminus{}0.0838]} \\
\addlinespace[2pt]
SGF: committing sink retained, denoising sink excluded & \shortstack{\textminus{}0.0046\\{}[\textminus{}0.0226,+0.0118]} & \shortstack{+0.0016\\{}[\textminus{}0.0028,+0.0065]} & \shortstack{\textminus{}0.0045\\{}[\textminus{}0.0428,+0.0268]} \\
\addlinespace[2pt]
LongLive: sink excluded from committing & \shortstack{\textminus{}0.0081\\{}[\textminus{}0.0185,+0.0009]} & \shortstack{\textminus{}0.0029\\{}[\textminus{}0.0083,+0.0022]} & \shortstack{\textminus{}0.0140\\{}[\textminus{}0.0342,+0.0022]} \\
\addlinespace[2pt]
LongLive: sink excluded from denoising & \shortstack{\textminus{}0.0853\sigmark\\{}[\textminus{}0.1212,\textminus{}0.0536]} & \shortstack{\textminus{}0.0272\sigmark\\{}[\textminus{}0.0382,\textminus{}0.0172]} & \shortstack{\textminus{}0.0427\\{}[\textminus{}0.0922,+0.0136]} \\
\addlinespace[2pt]
LongLive: sink excluded from both stages & \shortstack{\textminus{}0.0676\sigmark\\{}[\textminus{}0.0898,\textminus{}0.0458]} & \shortstack{\textminus{}0.0234\sigmark\\{}[\textminus{}0.0324,\textminus{}0.0150]} & \shortstack{\textminus{}0.0691\sigmark\\{}[\textminus{}0.1032,\textminus{}0.0380]} \\
\addlinespace[2pt]
LongLive: committing sink retained, denoising sink excluded & \shortstack{\textminus{}0.0177\sigmark\\{}[\textminus{}0.0358,\textminus{}0.0010]} & \shortstack{\textminus{}0.0038\\{}[\textminus{}0.0109,+0.0027]} & \shortstack{+0.0013\\{}[\textminus{}0.0160,+0.0285]} \\
\addlinespace[2pt]
Rolling Sink: base minus 21-frame recent-only control & \shortstack{\textminus{}0.0106\sigmark\\{}[\textminus{}0.0234,\textminus{}0.00002]} & \shortstack{\textminus{}0.0034\sigmark\\{}[\textminus{}0.0073,\textminus{}0.0001]} & \shortstack{\textminus{}0.0237\sigmark\\{}[\textminus{}0.0460,\textminus{}0.0042]} \\
\addlinespace[2pt]
\bottomrule
\end{tabular*}
\par\smallskip
\noindent\parbox{\linewidth}{\footnotesize\raggedright Changes use the original diagnostic scales with paired 95\% prompt-bootstrap intervals. Stage exclusions compare with reading the sink in both stages. Rows retaining the committing sink but excluding it from denoising compare with sink exclusion from both stages. Adjusted DINOv2 controls for Drift, brightness drift, and log relative motion. The Rolling Sink contrast reverses the paired estimate in Table~\ref{tab:diagnostic-similarity-components}. $^{*}$Interval excludes 0.}
\end{table}

\paragraph{Stage and exposure configurations.}
\label{app:sf-switching-config}
We use released SF DMD EMA weights with 20 prompts and generation seeds 0, 1, and 2. Each condition generates 192 latent frames in 64 blocks of 3 frames, corresponding to a nominal 48-second video. All conditions retain the same 160-frame physical cache without a sink. We use the released 4-step sampler with shift 5. The context of denoising step 0 contains the newest 12 trajectory frames, and that of steps 1--3 the newest 160, in every DiT layer 0--29. These windows include the current 3-frame block and use the available frames during warm-up. Historical attention reads all selected frames at their original RoPE positions. Cache rollover moves KV without renumbering or re-rotation.

Each committing pass runs the DiT at $t=0$ on the current clean block and reads historical KV to cover either the newest 12 or the newest 160 trajectory frames in total. The 4 schedules in Figure~\ref{fig:history-effect-summary} use recent context throughout, remote context throughout, remote then recent, or recent then remote. For the switching conditions, the first context is used for blocks 0--31 and the second for blocks 32--63. Block 32 completes denoising before its committing context changes. The newly constructed KV first affects generation of block 33. The nominal 24-second switch location is the latent-time boundary $32\times3/4$.

For the LongLive factorial study, we use 20 prompts and seeds 0 and 1 at 48 seconds. We cross whether denoising and committing read the sink at a fixed attention-context size of 12 latent frames. A stage that reads the sink uses 3 sink, 6 prior recent, and 3 current frames, and a stage that does not uses 9 prior recent and 3 current frames. Warm-up uses the available trajectory frames. Initial frames leave the recent-only read set once they fall outside its window. The current block is noisy during denoising and clean during committing. Sink KV remains stored in both conditions. For Rolling Sink, we use the same prompts and seeds, replacing the base committing context with the current block and KV from recent trajectory frames at a 21-frame committing-context size. We interpret this comparison as a rollout-policy effect because we observe repeat-run variation.

We use the released chunkwise SGF EMA weights on the same 20 prompts and generation seeds 0 and 1, generating 192 latent frames in 64 blocks. All 4 conditions use a 12-frame KV cache, including a 3-frame sink, with a 21-frame training-position range and top-aligned temporal positions. Each denoising and committing context contains either 3 sink, 3 prior recent, and 3 current frames or 6 prior recent and 3 current frames, 9 frames in total. We preserve the positions of selected historical KV and match initial noise and intermediate re-noising across the 4 conditions. The reference is the condition that reads the sink in both stages of this 9-frame experiment, rather than the 12-frame read set used in the base SGF benchmark comparison. We use the same diagnostic readouts and 4,000 prompt-cluster bootstrap draws as in the LongLive stage study.

\subsection{Intervals for the mechanism effects}
\label{app:mechanism-intervals}

Table~\ref{tab:longlive-stage-intervals} gives the paired 95\% intervals for the LongLive and SGF stage interventions in Table~\ref{tab:longlive-stage-effects}. The LongLive base means are 0.0969514 for Consistency and 0.0513247 for Drift. The SGF reference means are approximately 0.140 and 0.034, respectively. The Rolling Sink base Consistency mean is 0.1434. These diagnostic scales differ from VBench scores.

\begin{table}[!htbp]
\centering
\begin{minipage}{0.65\linewidth}
\centering
\footnotesize
\setlength{\tabcolsep}{3pt}
\renewcommand{\arraystretch}{1.1}
\caption{The sink is excluded from denoising, committing, or both stages at a fixed context size.}
\label{tab:longlive-stage-intervals}
\begin{tabularx}{\linewidth}{@{}cc>{\centering\arraybackslash}X>{\centering\arraybackslash}X@{}}
\reporttoprule
$D$ & $C$ & $\boldsymbol{\Delta}$ Consistency & $\boldsymbol{\Delta}$ Drift \\
\midrule
\multicolumn{4}{l}{\textit{LongLive: 12-frame reads}} \\
$\checkmark$ & $\checkmark$ & --- & --- \\
$\checkmark$ & $\times$ & \shortstack{\textminus{}0.0081\\{}[\textminus{}0.0180,+0.0006]} & \shortstack{\textminus{}0.0187\\{}[\textminus{}0.0287,\textminus{}0.0091]} \\
$\times$ & $\checkmark$ & \shortstack{\textminus{}0.0853\\{}[\textminus{}0.1215,\textminus{}0.0528]} & \shortstack{+0.2052\\{}[+0.1264,+0.2985]} \\
$\times$ & $\times$ & \shortstack{\textminus{}0.0676\\{}[\textminus{}0.0910,\textminus{}0.0459]} & \shortstack{+0.0822\\{}[+0.0460,+0.1198]} \\
\addlinespace[2pt]
\multicolumn{4}{l}{\textit{SGF: 9-frame reads}} \\
$\checkmark$ & $\checkmark$ & --- & --- \\
$\checkmark$ & $\times$ & \shortstack{+0.0194\\{}[+0.0075,+0.0304]} & \shortstack{\textminus{}0.0241\\{}[\textminus{}0.0428,\textminus{}0.0109]} \\
$\times$ & $\checkmark$ & \shortstack{\textminus{}0.1373\\{}[\textminus{}0.1680,\textminus{}0.1082]} & \shortstack{+0.2190\\{}[+0.1537,+0.2799]} \\
$\times$ & $\times$ & \shortstack{\textminus{}0.1327\\{}[\textminus{}0.1609,\textminus{}0.1060]} & \shortstack{+0.1264\\{}[+0.0719,+0.1918]} \\
\addlinespace[2pt]
\bottomrule
\end{tabularx}\par
\par\smallskip\noindent\parbox{\linewidth}{\footnotesize\raggedright $D$ and $C$ indicate whether denoising and committing read the sink. Changes are paired within each generator relative to reading the sink in both stages. Brackets give 95\% intervals on the original diagnostic scales.}\par
\end{minipage}
\end{table}

\paragraph{Reductions in drift and declining motion.}
\label{app:committing-drift-share}
Table~\ref{tab:committing-drift-share} reports reductions for the 9-frame SGF stage intervention and the LongLive source-and-position control in Section~\ref{app:content-position-controls}, where sink KV keeps its initial positions and recent context its trajectory positions. Both replace sink with recent trajectory context only during committing and still read the sink during denoising.

For Drift, brightness drift, and the fraction of videos with a late/early RGB-motion ratio below 0.85, we report $100(1-\bar q_{\mathrm R}/\bar q_{\mathrm S})$, where R and S denote recent-context and sink-context committing. Appearance retention is the ratio $a$ of mean late to mean early full-frame DINOv2 opening similarity. Its recovered loss is $100(a_{\mathrm R}-a_{\mathrm S})/(1-a_{\mathrm S})$, distinct from the own-minus-other-seed Consistency diagnostic. Intervals use 4,000 prompt-cluster bootstrap draws, retaining both seeds.

\begin{table}[!htbp]
\centering
\begin{minipage}{0.65\linewidth}
\centering
\footnotesize
\setlength{\tabcolsep}{3pt}
\renewcommand{\arraystretch}{1.1}
\caption{Drift, motion decline, and appearance retention with sink or recent KV during committing.}
\label{tab:committing-drift-share}
\noindent\textit{Drift and motion decline}\par\smallskip
\begin{tabular*}{\linewidth}{@{\extracolsep{\fill}}lccc@{}}
\reporttoprule
\textbf{Readout} & \textbf{Sink} & \textbf{Recent} & \textbf{Reduction} \\
\midrule
\multicolumn{4}{l}{\textit{SGF}} \\
Drift & 0.034 & 0.010 & \shortstack{71\%\sigmark\\{}[49\%,85\%]} \\
Brightness drift & 0.045 & 0.018 & \shortstack{59\%\sigmark\\{}[37\%,75\%]} \\
Declining-motion fraction & 0.25 & 0.05 & \shortstack{80\%\sigmark\\{}[50\%,100\%]} \\
\multicolumn{4}{l}{\textit{LongLive}} \\
Drift & 0.051 & 0.033 & \shortstack{37\%\sigmark\\{}[20\%,51\%]} \\
Brightness drift & 0.035 & 0.032 & \shortstack{9\%\\{}[-16\%,34\%]} \\
Declining-motion fraction & 0.28 & 0.05 & \shortstack{82\%\sigmark\\{}[57\%,100\%]} \\
\bottomrule
\end{tabular*}\par
\medskip
\noindent\textit{Appearance retention}\par\smallskip
\begin{tabular*}{\linewidth}{@{\extracolsep{\fill}}lccc@{}}
\reporttoprule
\textbf{Readout} & \textbf{Sink} & \textbf{Recent} & \textbf{Recovered loss} \\
\midrule
\multicolumn{4}{l}{\textit{SGF}} \\
Appearance retention & 0.804 & 0.905 & \shortstack{52\%\sigmark\\{}[20\%,101\%]} \\
\multicolumn{4}{l}{\textit{LongLive}} \\
Appearance retention & 0.817 & 0.770 & \shortstack{-25\%\\{}[-114\%,8\%]} \\
\bottomrule
\end{tabular*}\par
\medskip
\par\smallskip\noindent\parbox{\linewidth}{\footnotesize\raggedright Sink and Recent denote condition means on the original diagnostic scales. SGF uses the stage-intervention comparison. LongLive uses the source-and-position comparison in Section~\ref{app:content-position-controls}. Brackets show paired 95\% intervals. $^{*}$ marks an interval excluding 0.}\par
\end{minipage}
\end{table}

\paragraph{Decomposing appearance similarity.}
Table~\ref{tab:diagnostic-similarity-components} reports the own- and other-seed terms of Eq.~\ref{eq:diagnostic-consistency}. These terms measure full-frame appearance, including scene content and shared visual drift, rather than subject identity alone. Sink exclusion during LongLive committing and the 21-frame recent-only control on Rolling Sink increase both terms, compatible with reducing visual changes shared across generation seeds. Removing the sink from denoising or adding remote context during committing lowers both terms.

\begin{table}[!htbp]
\centering
\footnotesize
\setlength{\tabcolsep}{4pt}
\renewcommand{\arraystretch}{1.12}
\caption{Components of the DINOv2 Consistency diagnostic}
\label{tab:diagnostic-similarity-components}
\begin{tabular*}{\linewidth}{@{\extracolsep{\fill}}>{\raggedright\arraybackslash}p{.36\linewidth}ccc@{}}
\reporttoprule
\textbf{Configuration} & \shortstack{\textbf{Consistency}\\$\boldsymbol{\Delta}$ [95\% CI]} & \shortstack{\textbf{Own}\\$\boldsymbol{\Delta}$ [95\% CI]} & \shortstack{\textbf{Other}\\$\boldsymbol{\Delta}$ [95\% CI]} \\
\midrule
LongLive: no sink when committing & \shortstack{\textminus{}0.0081\\{}[\textminus{}0.0180, +0.0006]} & \shortstack{+0.0112\sigmark\\{}[+0.0025, +0.0219]} & \shortstack{+0.0193\sigmark\\{}[+0.0073, +0.0322]} \\
\addlinespace[2pt]
LongLive: no sink when denoising & \shortstack{\textminus{}0.0853\sigmark\\{}[\textminus{}0.1215, \textminus{}0.0528]} & \shortstack{\textminus{}0.3093\sigmark\\{}[\textminus{}0.3650, \textminus{}0.2546]} & \shortstack{\textminus{}0.2240\sigmark\\{}[\textminus{}0.2808, \textminus{}0.1711]} \\
\addlinespace[2pt]
LongLive: no sink in either stage & \shortstack{\textminus{}0.0676\sigmark\\{}[\textminus{}0.0910, \textminus{}0.0459]} & \shortstack{\textminus{}0.1575\sigmark\\{}[\textminus{}0.1956, \textminus{}0.1223]} & \shortstack{\textminus{}0.0898\sigmark\\{}[\textminus{}0.1218, \textminus{}0.0623]} \\
\addlinespace[2pt]
Rolling Sink: 21-frame recent-only control & \shortstack{+0.0106\sigmark\\{}[+0.00002, +0.0234]} & \shortstack{+0.0236\sigmark\\{}[+0.0085, +0.0409]} & \shortstack{+0.0130\\{}[\textminus{}0.0003, +0.0313]} \\
\addlinespace[2pt]
SF: committing with remote context throughout & \shortstack{\textminus{}0.0237\sigmark\\{}[\textminus{}0.0392, \textminus{}0.0102]} & \shortstack{\textminus{}0.0805\sigmark\\{}[\textminus{}0.1117, \textminus{}0.0547]} & \shortstack{\textminus{}0.0568\sigmark\\{}[\textminus{}0.0802, \textminus{}0.0374]} \\
\addlinespace[2pt]
SF: committing with remote context before 24 s & \shortstack{\textminus{}0.0105\\{}[\textminus{}0.0246, +0.0023]} & \shortstack{\textminus{}0.0257\sigmark\\{}[\textminus{}0.0461, \textminus{}0.0044]} & \shortstack{\textminus{}0.0152\sigmark\\{}[\textminus{}0.0269, \textminus{}0.0041]} \\
\addlinespace[2pt]
SF: committing with remote context after 24 s & \shortstack{\textminus{}0.0102\sigmark\\{}[\textminus{}0.0146, \textminus{}0.0058]} & \shortstack{\textminus{}0.0338\sigmark\\{}[\textminus{}0.0475, \textminus{}0.0214]} & \shortstack{\textminus{}0.0236\sigmark\\{}[\textminus{}0.0347, \textminus{}0.0138]} \\
\addlinespace[2pt]
\bottomrule
\end{tabular*}
\par\smallskip
\noindent\parbox{\linewidth}{\footnotesize\raggedright Consistency = Own \textminus{} Other. Own is similarity to the video's own opening. Other is similarity to openings from other seeds of the same prompt and configuration. Changes are relative to the base configuration for LongLive and Rolling Sink and to the 12-frame recent-only control for SF. Values use the original cosine-similarity scale with paired 95\% intervals. $^{*}$Interval excludes 0.}
\end{table}

\paragraph{Rolling Sink source readouts.}
\phantomsection
\label{app:rolling-sink-source}
The Consistency change in Table~\ref{tab:diagnostic-similarity-components} corresponds to an approximately 7.4\% relative increase over the base mean of 0.1434. The log relative-motion change is $+0.168$ [$+0.132,+0.208$]. Applying $100(\exp(\Delta)-1)$ gives an 18.3\% increase in the geometric-mean tail/early motion ratio, with an interval of 14.1--23.1\%.

The tail/early brightness-ratio change is $+0.0265$ [$+0.0142,+0.0413$], with mean ratios of $0.985$ and $1.011$. These values are on the original ratio scale. The saturation-ratio change is $-0.013$ [$-0.031,+0.002$]. The intervals for Drift and absolute brightness drift both span zero. The brightness level change therefore differs from a reduction in absolute drift magnitude.

\begin{figure}[!htbp]
\centering
\begin{minipage}{.65\linewidth}
\centering
\includegraphics[width=\linewidth]{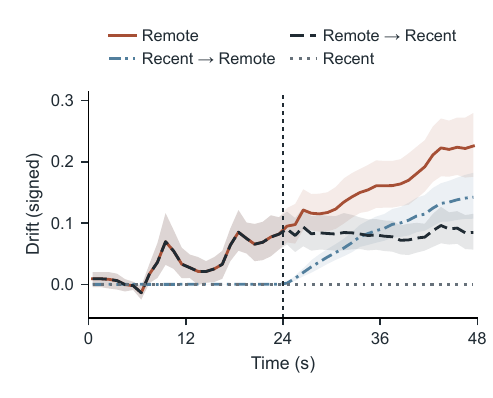}
\caption{Signed saturation differences on SF relative to the 12-frame recent-only control. Committing switches between 12- and 160-frame trajectory contexts after 24 seconds, with denoising fixed. Bands show pointwise 95\% intervals.}
\label{fig:history-effect-summary}
\end{minipage}
\end{figure}

\paragraph{Committing exposure and residual drift.}
\phantomsection
\label{app:committing-exposure}
The SF schedules in Section~\ref{app:sf-switching-config} hold denoising fixed and vary committing exposure, as shown in Figure~\ref{fig:history-effect-summary}. Switching from remote to recent context nearly stops further drift, while a positive normalized saturation offset remains in the 38--48-second window. Starting to read remote context at the same point produces a growing difference from the 12-frame recent-only control. Section~\ref{app:committing-propagation} reports the response to sink exposure on SGF.

We fit OLS slopes to paired saturation differences over 24--48 seconds, retaining all three seeds per prompt in 4,000 bootstrap draws. The slopes per 10 seconds are $+0.0593$ [$+0.0453,+0.0757$] for remote context throughout, $+0.0001$ [$-0.0046,+0.0056$] after switching to recent context, and $+0.0612$ [$+0.0463,+0.0781$] after switching to remote context. The switch-to-recent tail offset over 38--48 seconds is $+0.083$ [$+0.060,+0.109$]. Figure~\ref{fig:history-effect-summary} plots unsmoothed per-second saturation normalized by each video's 4--8-second mean.

\subsection{Committing pathways}
\label{app:committing-pathways}

\paragraph{Two SGF caches.}
We separate the cache read during denoising, $D$, from the chain cache read by later committing passes, $C$. Using released chunkwise SGF EMA weights, both caches retain the native 12-frame geometry (3 sink, 6 recent, and 3 current frames) and roll together. During denoising, $D$ is read with its base context. Each clean block undergoes two committing passes, one writing its entry in $C$ and the other in $D$. When a pass reads historical KV, it reads from $C$. Their read policies vary independently between the native 12-frame context and a context comprising the current block and KV from the two preceding blocks (9 frames in total). Table~\ref{tab:sgf-cache-pathways} reports the cache combinations. With $D$ constructed using the 9-frame context, changing $C$ from the native 12-frame context to 9 frames gives no significant additional gain in Quality or clip-to-clip Subject Consistency. Changing $C$ alone with native $D$ lowers Quality and consistency. With recent $D$, self-only construction of $C$ also lowers consistency relative to its 9-frame control.

\begin{table}[!htbp]
\centering
\begin{minipage}{0.80\linewidth}
\centering
\footnotesize
\setlength{\tabcolsep}{2pt}
\renewcommand{\arraystretch}{1.22}
\caption{The denoising-facing and chain caches use independently varied committing contexts on SGF.}
\label{tab:sgf-cache-pathways}
\begin{tabular*}{\linewidth}{@{\extracolsep{\fill}}cccccc@{}}
\reporttoprule
$D$ & $C$ & $\boldsymbol{\Delta}$ Q & \textbf{95\% interval} & $\boldsymbol{\Delta}$ DD & $\boldsymbol{\Delta}$ Subject$_{\mathrm{c2c}}$ \\
\midrule
$R_9$ & $R_9$ & +0.59\sigmark & [+0.16,+1.04] & +10.37\sigmark & +3.98\sigmark \\
$R_9$ & $N_{12}$ & +0.52\sigmark & [+0.11,+0.95] & +8.73\sigmark & +3.89\sigmark \\
$N_{12}$ & $R_9$ & \textminus{}0.53\sigmark & [\textminus{}0.75,\textminus{}0.31] & \textminus{}3.47\sigmark & \textminus{}1.69\sigmark \\
$R_9$ & $S_3$ & +0.54\sigmark & [+0.18,+0.93] & +12.47\sigmark & +0.99 \\
\bottomrule
\end{tabular*}\par
\par\smallskip\noindent\parbox{\linewidth}{\footnotesize\raggedright Base SGF constructs both $D$ and $C$ with $N_{12}$. Policies specify the current latent frames and historical KV read from $C$. $R_9$ uses 3 current latent frames and KV from 6 recent frames, $N_{12}$ uses 3 current latent frames and KV from 3 sink and 6 recent frames, and $S_3$ uses only the current block. All rows use Rolling Forcing prompts 0--99 at 60 seconds with generation seed 0. Q denotes Quality, DD Dynamic Degree, and Subject$_{\mathrm{c2c}}$ clip-to-clip Subject Consistency. Scores are $\times100$. Q brackets give 95\% intervals from 4,000 paired prompt-bootstrap draws. $^{*}$ marks an interval excluding 0.}\par
\end{minipage}
\end{table}

\paragraph{Extra historical context during denoising.}
\label{app:extra-old-history}
We test whether additional KV from older historical context can reproduce the consistency benefit of Commit Forcing on released LongLive weights. The denoising context contains the base 3 sink frames and the newest 9 trajectory frames, including the 3 current frames. Starting at block $b=4$, we append self-only KV from the distinct trajectory blocks in $\{1,\lfloor(b-2)/2\rfloor,b-3\}$, retaining their original temporal positions. Condition A adds these tokens at denoising steps 3--4 in layers 20--29, and B uses all 4 steps in layers 15--29. Blocks and layers are indexed from 0. Each source block supplies 4,680 tokens. A compressed condition averages each source block's KV into 20 tokens over a $5\times4$ spatial grid and its three frames, removing RoPE before pooling and restoring it at grid centers and the source block's middle-frame position while preserving token multiplicity. These conditions use self-only committing. An add-on condition applies A with 6-frame Commit Forcing and obtains source KV through separate self-only passes. Table~\ref{tab:extra-old-history} reports no significant clip-to-clip Subject Consistency gain in any condition. In A, extra KV receives mean attention share 0.187, compared with 0.106 for the sink, 0.216 for recent context, and 0.491 for the current block. The attention probe covers prompts 3 and 41 at blocks 5, 15, $\ldots$, 75. We average the attention shares over A's steps and layers. KV from older historical context is attended, but does not reproduce the 1.00-point consistency gain of 6-frame Commit Forcing over self-only committing.

\begin{table}[!htbp]
\centering
\begin{minipage}{0.80\linewidth}
\centering
\footnotesize
\setlength{\tabcolsep}{5pt}
\renewcommand{\arraystretch}{1.12}
\caption{Extra historical KV is added during LongLive denoising with self-only committing or Commit Forcing.}
\label{tab:extra-old-history}
\begin{tabular*}{\linewidth}{@{\extracolsep{\fill}}lcccc@{}}
\reporttoprule
\textbf{Metric} & \shortstack{$A-S_3$\\$n=200$} & \shortstack{$B-S_3$\\$n=100$} & \shortstack{$A_{20}-S_3$\\$n=100$} & \shortstack{$(A+R_6)-R_6$\\$n=100$} \\
\midrule
Quality & \shortstack{\textminus{}0.03\\{}[\textminus{}0.10,+0.05]} & \shortstack{+0.05\\{}[\textminus{}0.13,+0.22]} & \shortstack{+0.11\sigmark\\{}[+0.01,+0.21]} & \shortstack{\textminus{}0.12\sigmark\\{}[\textminus{}0.23,\textminus{}0.01]} \\
Dynamic Degree & \shortstack{+0.78\\{}[\textminus{}0.13,+1.67]} & \shortstack{+2.03\\{}[\textminus{}0.27,+4.13]} & \shortstack{+1.63\sigmark\\{}[+0.50,+2.93]} & \shortstack{\textminus{}0.43\\{}[\textminus{}1.90,+0.97]} \\
Subject$_{\mathrm{c2c}}$ & \shortstack{+0.03\\{}[\textminus{}0.14,+0.20]} & \shortstack{+0.09\\{}[\textminus{}0.31,+0.48]} & \shortstack{\textminus{}0.23\\{}[\textminus{}0.52,+0.04]} & \shortstack{+0.07\\{}[\textminus{}0.20,+0.33]} \\
Imaging Quality & \shortstack{\textminus{}0.39\sigmark\\{}[\textminus{}0.57,\textminus{}0.23]} & \shortstack{\textminus{}0.55\sigmark\\{}[\textminus{}0.83,\textminus{}0.27]} & \shortstack{+0.15\sigmark\\{}[+0.00,+0.29]} & \shortstack{\textminus{}0.50\sigmark\\{}[\textminus{}0.70,\textminus{}0.32]} \\
Aesthetic Quality & \shortstack{\textminus{}0.24\sigmark\\{}[\textminus{}0.34,\textminus{}0.13]} & \shortstack{\textminus{}0.38\sigmark\\{}[\textminus{}0.59,\textminus{}0.19]} & \shortstack{\textminus{}0.07\\{}[\textminus{}0.20,+0.06]} & \shortstack{\textminus{}0.21\sigmark\\{}[\textminus{}0.36,\textminus{}0.06]} \\
\bottomrule
\end{tabular*}
\par\smallskip
\noindent\parbox{\linewidth}{\footnotesize\raggedright A and B use self-only committing ($S_3$), $A_{20}$ compresses each source block's KV to 20 tokens, and $A+R_6$ adds A to 6-frame committing. A uses all 200 Rolling Forcing prompts. The other configurations use prompts 0--99. All videos last 60 seconds and use generation seed 0. Q denotes Quality, DD Dynamic Degree, Subject$_{\mathrm{c2c}}$ clip-to-clip Subject Consistency, IQ Imaging Quality, and AQ Aesthetic Quality. Scores are $\times100$, with 95\% intervals from 4,000 paired prompt-bootstrap draws. $^{*}$ marks an interval excluding 0 before rounding.}
\end{minipage}
\end{table}

\paragraph{Spatial averaging of the committing correction.}
\label{app:committing-variants}
On LongLive, we spatially average the committing attention correction within each frame, 16 content regions, a $4\times4$ spatial grid, or 64 content regions. Denoising follows the base policy. Table~\ref{tab:committing-averaging-controls} reports their changes from Commit Forcing.

\begin{figure}[!htbp]
\centering
\begin{minipage}{.65\linewidth}
\centering
\includegraphics[width=\linewidth]{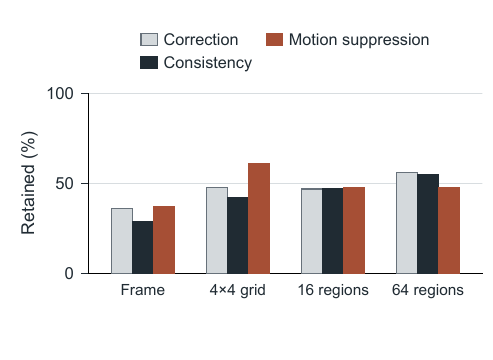}
\caption{Retained correction magnitude, consistency gain, and motion suppression for each spatial-averaging control. Unaveraged Commit Forcing defines the 100\% reference for each quantity.}
\label{fig:committing-correction-dose}
\end{minipage}
\end{figure}

For the analysis in Figure~\ref{fig:committing-correction-dose}, let $O_C$ be the committing self-attention output using the current block alone and $O_{C+H}$ the output when it also reads the preceding block's stored KV, both before the output projection. The correction is $D=O_{C+H}-O_C$.

We measure $\|\mathrm{pool}(D)\|_F/\max(\|D\|_F,10^{-12})$ in single precision over current tokens, heads, and channels. Ratios are averaged over 30 layers and 79 post-startup committing passes within each video, then over 100 prompts. The reported mean ratios are 0.36, 0.47, 0.48, and 0.56 for frame averaging, 16 content regions, the $4\times4$ grid, and 64 content regions, respectively. For each metric, the retained effect is $1-\Delta_{\mathrm{control}}/\Delta_{\mathrm{self}}$, using its paired difference from Commit Forcing and the reported 200-prompt self-only difference in Table~\ref{tab:longlive-window}. The consistency and motion-suppression fractions are $(0.29,0.37)$, $(0.47,0.48)$, $(0.42,0.61)$, and $(0.55,0.48)$, respectively. The retained effects differ across controls, each of which changes both correction magnitude and spatial structure.

\begin{table}[!htbp]
\centering
\begin{minipage}{0.65\linewidth}
\centering
\footnotesize
\setlength{\tabcolsep}{3pt}
\renewcommand{\arraystretch}{1.1}
\caption{Spatial averaging weakens the preceding block's attention correction during LongLive committing.}
\label{tab:committing-averaging-controls}
\begin{tabular*}{\linewidth}{@{\extracolsep{\fill}}lcc@{}}
\reporttoprule
\textbf{Averaging} & $\boldsymbol{\Delta}$ DD & $\boldsymbol{\Delta}$ Subject$_{\mathrm{c2c}}$ \\
\midrule
Within frame & +2.30 & \textminus{}0.71\sigmark \\
$4\times4$ grid & +1.43 & \textminus{}0.58\sigmark \\
16 regions & +1.90 & \textminus{}0.53\sigmark \\
64 regions & +1.87 & \textminus{}0.45\sigmark \\
\bottomrule
\end{tabular*}\par
\medskip
\begin{tabular*}{\linewidth}{@{\extracolsep{\fill}}lcc@{}}
\reporttoprule
\textbf{Averaging} & $\boldsymbol{\Delta}$ Quality & \textbf{95\% interval} \\
\midrule
Within frame & +0.02 & [\textminus{}0.15,\,+0.19] \\
$4\times4$ grid & \textminus{}0.04 & [\textminus{}0.19,\,+0.10] \\
16 regions & \textminus{}0.01 & [\textminus{}0.16,\,+0.14] \\
64 regions & \textminus{}0.00 & [\textminus{}0.15,\,+0.15] \\
\bottomrule
\end{tabular*}\par
\par\smallskip\noindent\parbox{\linewidth}{\footnotesize\raggedright Changes are relative to Commit Forcing on the first 100 Rolling Forcing prompts at 60 seconds with generation seed 0. Scores are $\times100$. Brackets give Quality intervals from 4,000 paired prompt-bootstrap draws, and $^{*}$ marks a paired 95\% interval excluding 0.}\par
\end{minipage}
\end{table}

\subsection{Historical source and temporal position}
\label{app:content-position-controls}

We compare 4 LongLive committing configurations on 20 prompts with generation seeds 0 and 1 at 48 seconds, giving 40 videos per configuration. We retain a 15-frame physical KV cache and the base denoising context of 3 sink frames plus the newest 9 trajectory frames. During committing, the current block and KV from the newest trajectory frames form a 9-frame context. We add KV from 3 frames taken either from the sink or from the oldest part of the newest 12 trajectory frames, assigning it to either the initial positions 0--2 or the corresponding recent positions. Committing-context sizes count latent frames and include the current 3-frame block.

When changing positions, we re-rotate the temporal RoPE component of K on the read copy, without editing the stored entries or V. We retain the same sampler and per-video random seed across configurations. During the first 4 blocks, we use the base context before all 15 cache frames are available. We report paired changes from recent KV at recent temporal positions in Table~\ref{tab:committing-content-positions}, with 4,000 bootstrap draws over prompts, retaining both generation seeds of each prompt.

At recent positions, replacing recent KV with sink KV increases drift. Moving recent KV to the initial positions gives a smaller change whose confidence interval includes 0. With sink KV, drift increases by $0.0504$ [$+0.0291,+0.0718$] at recent rather than initial positions. Historical source and temporal position therefore both affect subsequent drift. We vary stored KV provenance, with content, trajectory age, and KV construction history coupled within each source.

\begin{table}[!htbp]
\centering
\footnotesize
\setlength{\tabcolsep}{4pt}
\renewcommand{\arraystretch}{1.10}
\caption{Committing with different KV sources and positions on LongLive.}
\label{tab:committing-content-positions}
\begin{tabular*}{\linewidth}{@{\extracolsep{\fill}}lccccc@{}}
\reporttoprule
 & & \multicolumn{4}{c}{\textbf{Paired change [95\% interval]}} \\
\cmidrule(lr){3-6}
\textbf{Source} & \textbf{Position} & \textbf{Consistency} & \textbf{Drift} & \textbf{Brightness drift} & \textbf{Log motion ratio} \\
\midrule
Recent KV & Recent & --- & --- & --- & --- \\
Recent KV & Initial & \shortstack{+0.0041\\{}[\textminus{}0.0049,+0.0125]} & \shortstack{+0.0072\\{}[\textminus{}0.0027,+0.0218]} & \shortstack{+0.0006\\{}[\textminus{}0.0054,+0.0077]} & \shortstack{+0.0113\\{}[\textminus{}0.0501,+0.0724]} \\
Sink KV & Initial & \shortstack{+0.0084\\{}[\textminus{}0.0009,+0.0192]} & \shortstack{+0.0188\sigmark\\{}[+0.0091,+0.0288]} & \shortstack{+0.0034\\{}[\textminus{}0.0052,+0.0133]} & \shortstack{\textminus{}0.1179\sigmark\\{}[\textminus{}0.1766,\textminus{}0.0611]} \\
Sink KV & Recent & \shortstack{\textminus{}0.0164\sigmark\\{}[\textminus{}0.0329,\textminus{}0.0018]} & \shortstack{+0.0692\sigmark\\{}[+0.0481,+0.0923]} & \shortstack{+0.0178\sigmark\\{}[+0.0074,+0.0285]} & \shortstack{+0.0233\\{}[\textminus{}0.0528,+0.0984]} \\
\bottomrule
\end{tabular*}
\par\smallskip
\parbox{\linewidth}{\footnotesize\raggedright Paired changes from recent content at recent positions, with 95\% intervals on the original diagnostic scales. Drift and brightness drift are absolute deviations of the tail/early saturation and brightness ratios from 1. Motion is the log tail/early motion ratio. Absolute Consistency and Drift means appear in Table~\ref{tab:committing-source-position}. Reference brightness drift is 0.032. $^{*}$Interval excludes 0.}
\end{table}

\paragraph{Matched context size on SGF.}
On SGF, we replace sink KV with recent trajectory KV at a fixed 9-frame committing-context size, with denoising unchanged. Table~\ref{tab:sgf-committing-controls} reports this contrast as B$-$D, together with the self-only contrasts in Section~\ref{app:self-only-contrasts}. Paired intervals use 4,000 prompt-bootstrap draws. The descriptive Dynamic Degree adjustment fits an OLS regression of paired Subject$_{\mathrm{c2c}}$ changes on Dynamic Degree changes and reports the intercept from 2,000 prompt-bootstrap draws, retaining both seeds of each prompt. This comparison changes historical content, age, and position together.

\phantomsection
\label{app:sgf-committing-controls}
For the SGF committing controls in Table~\ref{tab:sgf-committing-controls}, we evaluate Rolling Forcing's 200 prompts at 60 seconds with generation seeds 0 and 1, holding released EMA weights, 12-frame cache geometry, top-aligned positions, denoising steps, sampler, and DiT forward-pass count fixed. The committing contexts are 3 sink, 6 recent, and 3 current frames for base SGF, 6 recent and 3 current frames for the 9-frame recent-context control, 3 current frames for self-only committing, and 3 sink, 3 recent, and 3 current frames for the equal-count sink control. We select historical KV after RoPE and retain its original positions. The base and 9-frame recent-context outputs are reused; the two controls use the same model and attention computation. Initial noise is matched across all 4 contexts, but intermediate re-noising is not controlled.

\begin{table}[!htbp]
\centering
\begin{minipage}{0.65\linewidth}
\centering
\footnotesize
\setlength{\tabcolsep}{3pt}
\renewcommand{\arraystretch}{1.1}
\caption{SGF committing controls pooled across generation seeds}
\label{tab:sgf-committing-controls}
\begin{tabularx}{\linewidth}{@{}>{\raggedright\arraybackslash}X>{\centering\arraybackslash}X>{\centering\arraybackslash}X@{}}
\reporttoprule
\textbf{Metric} & $\boldsymbol{\Delta}$ & \textbf{95\% interval} \\
\midrule
\multicolumn{3}{@{}p{\linewidth}@{}}{\textit{B\textminus{}A}} \\
Quality & +0.95\sigmark & [+0.66, +1.25] \\
Subject$_{\mathrm{c2c}}$ & +4.60\sigmark & [+3.86, +5.34] \\
Dynamic Degree & +13.35\sigmark & [+9.54, +17.26] \\
Motion Smoothness & \textminus{}0.20\sigmark & [\textminus{}0.27, \textminus{}0.13] \\
Temporal Flickering & \textminus{}0.43\sigmark & [\textminus{}0.57, \textminus{}0.31] \\
Background$_{\mathrm{c2c}}$ & +3.05\sigmark & [+2.59, +3.54] \\
Subject (fused) & +0.26\sigmark & [+0.19, +0.32] \\
Aesthetic Quality & +0.83\sigmark & [+0.38, +1.29] \\
Imaging Quality & \textminus{}0.00 & [\textminus{}0.42, +0.45] \\
Adjusted Subject$_{\mathrm{c2c}}$ & +4.53\sigmark & [+3.84, +5.30] \\
\addlinespace[2pt]
\multicolumn{3}{@{}p{\linewidth}@{}}{\textit{B\textminus{}C}} \\
Quality & \textminus{}0.10 & [\textminus{}0.30, +0.10] \\
Subject$_{\mathrm{c2c}}$ & \textminus{}0.07 & [\textminus{}0.51, +0.39] \\
Dynamic Degree & \textminus{}2.85\sigmark & [\textminus{}5.48, \textminus{}0.23] \\
Motion Smoothness & +0.11\sigmark & [+0.07, +0.16] \\
Temporal Flickering & +0.16\sigmark & [+0.08, +0.26] \\
Background$_{\mathrm{c2c}}$ & \textminus{}0.18 & [\textminus{}0.53, +0.15] \\
Subject (fused) & +0.04 & [\textminus{}0.01, +0.09] \\
Aesthetic Quality & \textminus{}0.21 & [\textminus{}0.61, +0.18] \\
Imaging Quality & +0.13 & [\textminus{}0.20, +0.47] \\
Adjusted Subject$_{\mathrm{c2c}}$ & \textminus{}0.17 & [\textminus{}0.65, +0.27] \\
\addlinespace[2pt]
\multicolumn{3}{@{}p{\linewidth}@{}}{\textit{B\textminus{}D}} \\
Quality & +1.25\sigmark & [+0.92, +1.57] \\
Subject$_{\mathrm{c2c}}$ & +4.54\sigmark & [+3.84, +5.23] \\
Dynamic Degree & +18.12\sigmark & [+13.76, +22.48] \\
Motion Smoothness & \textminus{}0.26\sigmark & [\textminus{}0.33, \textminus{}0.19] \\
Temporal Flickering & \textminus{}0.58\sigmark & [\textminus{}0.71, \textminus{}0.44] \\
Background$_{\mathrm{c2c}}$ & +3.23\sigmark & [+2.77, +3.68] \\
Subject (fused) & +0.20\sigmark & [+0.13, +0.26] \\
Aesthetic Quality & +0.89\sigmark & [+0.39, +1.40] \\
Imaging Quality & +0.27 & [\textminus{}0.17, +0.73] \\
Adjusted Subject$_{\mathrm{c2c}}$ & +4.48\sigmark & [+3.77, +5.24] \\
\addlinespace[2pt]
\bottomrule
\end{tabularx}\par
\par\smallskip\noindent\parbox{\linewidth}{\footnotesize\raggedright A denotes base SGF, B the 9-frame wider-window reference, C self-only committing, and D 9-frame committing with the sink retained. Values are signed changes with paired 95\% intervals. Pooled estimates average the two generation seeds within each prompt. Subject denotes fused Subject Consistency. Subject$_{\mathrm{c2c}}$ and Background$_{\mathrm{c2c}}$ denote clip-to-clip Subject and Background Consistency, and DD denotes Dynamic Degree. MS denotes Motion Smoothness, and TF denotes Temporal Flickering. Adjusted Subject$_{\mathrm{c2c}}$ controls for paired Dynamic Degree differences. $^{*}$Interval excludes 0.}\par
\end{minipage}
\end{table}

\subsection{Historical recency at fixed committing positions}
\label{app:committing-content-age}

We vary historical recency while keeping the committing-context size and assigned temporal positions fixed. We evaluate SGF and LongLive on Rolling Forcing prompts 0--99 at 60 seconds with generation seed 0. After the initial block, each committing context contains the three current latent frames and KV from three historical frames. We select this historical KV at target lags $d\in\{1,2,4,8,16\}$ blocks, or from block 0 in the sink condition. We assign all historical keys to the three temporal positions immediately preceding the current block. We retain the model weights and each generator's base denoising policy, including reading the sink during denoising.

For block $b\ge2$, we select historical block $h_b=\max(1,b-d)$ in the nonsink conditions. The context of block 0 contains only its current three frames, and that of block 1 adds the KV entry of block 0 in every condition. The actual source lag increases during startup until the target lag $d$ is available. In the sink condition, we use $h_b=0$ for all $b\ge1$, so its temporal lag grows with the generated trajectory. We treat the sink as a separate source and exclude it from the target-lag regression.

Each target-lag condition is a separate free-generation rollout, and historical KV comes from that condition's own earlier blocks. Initial and sampler noise are matched across conditions, and each committing context produces its own latent trajectory. This differs from the propagation probe in Section~\ref{app:committing-propagation}, which supplies the same clean latent blocks to both caches.

For each historical block, we use the KV from its final $t=0$ committing pass. SGF and LongLive use active KV caches of 12 and 15 frames, respectively. We preserve committed KV for up to 51 additional frames after eviction, allowing each condition to read its target lag from either the active cache or this additional storage.

SGF stores K before RoPE. In the controls of Section~\ref{app:sgf-committing-controls}, we select historical KV after applying RoPE and retain its original positions. The recency diagnostic instead selects historical KV from the cache or auxiliary store before applying RoPE at the assigned positions immediately before the current block. LongLive stores K after RoPE, so we re-rotate its temporal component to the assigned positions while leaving V unchanged. At $d=1$, the rotation offset is 0.

Tables~\ref{tab:committing-content-age-core} and~\ref{tab:committing-content-age-trends} report the 12 conditions and paired lag trends. Mean clip-to-clip Subject and Background Consistency decrease across the five tested lags in both generators. Relative to $d=1$, Subject changes are not significant at $d=2$, while the intervals are entirely negative at $d=4,8,16$.

\begin{table}[!htbp]
\centering
\footnotesize
\setlength{\tabcolsep}{3pt}
\renewcommand{\arraystretch}{1.12}
\caption{Historical-context lag affects consistency, motion, and Quality at fixed committing positions.}
\label{tab:committing-content-age-core}
\label{tab:committing-content-age-components}
\begin{tabular*}{\linewidth}{@{\extracolsep{\fill}}lccccccccccc@{}}
\reporttoprule
\textbf{Context} & \textbf{Q} & \textbf{DD} & \textbf{Subject}$_{\mathrm{c2c}}$ & \textbf{Bg.}$_{\mathrm{c2c}}$ & \textbf{SC} & \textbf{BC} & \shortstack{\textbf{Subject}\\\textbf{in-clip}} & \textbf{IQ} & \textbf{AQ} & \textbf{MS} & \textbf{TF} \\
\midrule
\multicolumn{12}{@{}l}{\textit{SGF}} \\
$d=1$ & 84.11 & 66.27 & 90.61 & 92.85 & 98.26 & 96.84 & 97.66 & 70.58 & 62.34 & 98.42 & 96.87 \\
$d=2$ & 84.21 & 68.37 & 90.53 & 92.74 & 98.22 & 96.80 & 97.57 & 70.61 & 62.36 & 98.38 & 96.80 \\
$d=4$ & 84.19 & 68.60 & 90.30 & 92.55 & 98.19 & 96.79 & 97.55 & 70.58 & 62.24 & 98.37 & 96.78 \\
$d=8$ & 84.24 & 69.67 & 90.02 & 92.42 & 98.17 & 96.76 & 97.53 & 70.61 & 62.22 & 98.36 & 96.75 \\
$d=16$ & 84.34 & 71.30 & 89.92 & 92.34 & 98.16 & 96.73 & 97.50 & 70.65 & 62.22 & 98.34 & 96.71 \\
Sink & 83.14 & 72.53 & 78.62 & 83.52 & 97.15 & 96.00 & 96.61 & 69.71 & 59.78 & 97.99 & 96.09 \\
\addlinespace[3pt]
\multicolumn{12}{@{}l}{\textit{LongLive}} \\
$d=1$ & 82.82 & 54.00 & 85.50 & 88.65 & 97.60 & 96.54 & 96.89 & 69.60 & 60.25 & 98.59 & 97.38 \\
$d=2$ & 82.78 & 55.13 & 85.16 & 88.37 & 97.50 & 96.43 & 96.71 & 69.34 & 60.18 & 98.56 & 97.33 \\
$d=4$ & 82.74 & 56.00 & 84.35 & 87.67 & 97.40 & 96.32 & 96.59 & 69.33 & 60.04 & 98.54 & 97.28 \\
$d=8$ & 82.75 & 57.93 & 83.68 & 86.91 & 97.31 & 96.24 & 96.47 & 69.12 & 59.90 & 98.50 & 97.20 \\
$d=16$ & 82.63 & 57.57 & 83.03 & 86.26 & 97.24 & 96.18 & 96.41 & 68.91 & 59.69 & 98.49 & 97.20 \\
Sink & 81.85 & 56.47 & 78.28 & 80.25 & 96.79 & 95.97 & 96.00 & 65.12 & 59.19 & 98.61 & 97.26 \\
\bottomrule
\end{tabular*}
\par\smallskip
\noindent\parbox{\linewidth}{\footnotesize\raggedright Each context contains three current latent frames and KV from three historical frames after startup. $d$ is the target temporal lag of the historical content, in blocks. In the sink condition, KV from block 0 is read at the same assigned positions. Scores use the $\times100$ scale. Subject$_{\mathrm{c2c}}$ and Bg.$_{\mathrm{c2c}}$ denote clip-to-clip Subject and Background Consistency, and DD denotes Dynamic Degree. SC and BC denote fused Subject and Background Consistency. IQ and AQ denote Imaging and Aesthetic Quality. MS and TF denote Motion Smoothness and Temporal Flickering. The in-clip column reports Subject Consistency within each clip.}
\end{table}

\paragraph{Sink KV.}
With sink KV at the same assigned positions and context size, clip-to-clip Subject Consistency is 11.30 points lower on SGF and 4.75 points lower on LongLive than with $d=16$, as shown in Table~\ref{tab:committing-content-age-contrasts}. This contrast changes the source block, its encoded content and its trajectory age together. Sink KV yields lower consistency than trajectory KV at every tested lag.

\begin{table}[!htbp]
\centering
\begin{minipage}{0.65\linewidth}
\centering
\footnotesize
\setlength{\tabcolsep}{3pt}
\renewcommand{\arraystretch}{1.1}
\caption{Lag slopes and paired source contrasts quantify the historical-context effects on SGF and LongLive.}
\label{tab:committing-content-age-trends}
\label{tab:committing-content-age-contrasts}
\noindent\textit{Change per doubling of target lag}\par\smallskip
\begin{tabularx}{\linewidth}{@{}>{\raggedright\arraybackslash}X>{\centering\arraybackslash}X>{\centering\arraybackslash}X@{}}
\reporttoprule
\textbf{Metric} & \textbf{SGF} & \textbf{LongLive} \\
\midrule
Subject$_{\mathrm{c2c}}$ & \shortstack{\textminus{}0.190\sigmark\\{}[\textminus{}0.273,\textminus{}0.117]} & \shortstack{\textminus{}0.641\sigmark\\{}[\textminus{}0.784,\textminus{}0.508]} \\
Bg.$_{\mathrm{c2c}}$ & \shortstack{\textminus{}0.134\sigmark\\{}[\textminus{}0.181,\textminus{}0.090]} & \shortstack{\textminus{}0.625\sigmark\\{}[\textminus{}0.712,\textminus{}0.536]} \\
Quality & \shortstack{+0.048\sigmark\\{}[+0.012,+0.085]} & \shortstack{\textminus{}0.041\\{}[\textminus{}0.084,+0.005]} \\
DD & \shortstack{+1.137\sigmark\\{}[+0.693,+1.647]} & \shortstack{+0.993\sigmark\\{}[+0.447,+1.593]} \\
Subject in-clip & \shortstack{\textminus{}0.034\sigmark\\{}[\textminus{}0.046,\textminus{}0.023]} & \shortstack{\textminus{}0.122\sigmark\\{}[\textminus{}0.148,\textminus{}0.096]} \\
MS & \shortstack{\textminus{}0.018\sigmark\\{}[\textminus{}0.025,\textminus{}0.012]} & \shortstack{\textminus{}0.028\sigmark\\{}[\textminus{}0.039,\textminus{}0.018]} \\
TF & \shortstack{\textminus{}0.039\sigmark\\{}[\textminus{}0.054,\textminus{}0.025]} & \shortstack{\textminus{}0.050\sigmark\\{}[\textminus{}0.068,\textminus{}0.032]} \\
\bottomrule
\end{tabularx}\par
\par\smallskip\noindent\parbox{\linewidth}{\footnotesize\raggedright Changes per doubling of $d$, on the $\times100$ scale. Entries are mean per-prompt least-squares slopes against $\log_2 d$ for $d\in\{1,2,4,8,16\}$, with 95\% intervals from 4,000 paired prompt-bootstrap draws. The sink is excluded from these fits. $^{*}$Interval excludes 0.}\par
\medskip
\noindent\textit{Paired changes between historical contexts}\par\smallskip
\begin{tabularx}{\linewidth}{@{}>{\raggedright\arraybackslash}X>{\centering\arraybackslash}X>{\centering\arraybackslash}X@{}}
\reporttoprule
\textbf{Metric} & \textbf{SGF} & \textbf{LongLive} \\
\midrule
\multicolumn{3}{@{}p{\linewidth}@{}}{\textit{$d=2$ minus $d=1$}} \\
Subject$_{\mathrm{c2c}}$ & \shortstack{\textminus{}0.08\\{}[\textminus{}0.30, +0.14]} & \shortstack{\textminus{}0.33\\{}[\textminus{}0.71, +0.06]} \\
\multicolumn{3}{@{}p{\linewidth}@{}}{\textit{$d=4$ minus $d=1$}} \\
Subject$_{\mathrm{c2c}}$ & \shortstack{\textminus{}0.31\sigmark\\{}[\textminus{}0.57, \textminus{}0.05]} & \shortstack{\textminus{}1.14\sigmark\\{}[\textminus{}1.67, \textminus{}0.67]} \\
\multicolumn{3}{@{}p{\linewidth}@{}}{\textit{$d=8$ minus $d=1$}} \\
Subject$_{\mathrm{c2c}}$ & \shortstack{\textminus{}0.59\sigmark\\{}[\textminus{}0.89, \textminus{}0.32]} & \shortstack{\textminus{}1.82\sigmark\\{}[\textminus{}2.35, \textminus{}1.32]} \\
\multicolumn{3}{@{}p{\linewidth}@{}}{\textit{$d=16$ minus $d=1$}} \\
Subject$_{\mathrm{c2c}}$ & \shortstack{\textminus{}0.69\sigmark\\{}[\textminus{}1.06, \textminus{}0.38]} & \shortstack{\textminus{}2.46\sigmark\\{}[\textminus{}3.09, \textminus{}1.88]} \\
DD & \shortstack{+5.03\sigmark\\{}[+2.83, +7.43]} & \shortstack{+3.57\sigmark\\{}[+1.53, +5.80]} \\
\multicolumn{3}{@{}p{\linewidth}@{}}{\textit{Sink minus $d=1$}} \\
Subject$_{\mathrm{c2c}}$ & \shortstack{\textminus{}11.99\sigmark\\{}[\textminus{}13.59, \textminus{}10.44]} & \shortstack{\textminus{}7.22\sigmark\\{}[\textminus{}8.16, \textminus{}6.28]} \\
Quality & \shortstack{\textminus{}0.97\sigmark\\{}[\textminus{}1.55, \textminus{}0.45]} & \shortstack{\textminus{}0.97\sigmark\\{}[\textminus{}1.30, \textminus{}0.63]} \\
\multicolumn{3}{@{}p{\linewidth}@{}}{\textit{Sink minus $d=16$}} \\
Subject$_{\mathrm{c2c}}$ & \shortstack{\textminus{}11.30\sigmark\\{}[\textminus{}12.86, \textminus{}9.76]} & \shortstack{\textminus{}4.75\sigmark\\{}[\textminus{}5.57, \textminus{}3.97]} \\
\multicolumn{3}{@{}p{\linewidth}@{}}{\textit{Base minus $d=1$}} \\
Subject$_{\mathrm{c2c}}$ & \textemdash & \shortstack{\textminus{}1.22\sigmark\\{}[\textminus{}1.83, \textminus{}0.63]} \\
\bottomrule
\end{tabularx}\par
\par\smallskip\noindent\parbox{\linewidth}{\footnotesize\raggedright Prompt-paired changes and their 95\% intervals on the $\times100$ scale, using 4,000 prompt-bootstrap draws. The final row is a reference comparison with base LongLive, which uses its original 12-frame committing context and sink positions. This reference is separate from the fixed 6-frame recency controls. $^{*}$Interval excludes 0.}\par

\end{minipage}
\end{table}

\paragraph{Historical recency with a fixed trajectory.}
\label{app:committing-recency-fixed-trajectory}
The fixed-trajectory probe uses a separate 20-prompt diagnostic list, comprising indices 0--20 except 5 in Self-Forcing's expanded MovieGen prompts, with zero-based indexing. We generate a 48-second reference SGF video with a 6-frame committing context ($d=1$) for each prompt and generation seeds 0 and 1, then fix its clean latent trajectory. For each historical condition, we reconstruct a separate cache by committing the same latent blocks in order, using the target-lag selection and assigned positions above. After blocks $m\in\{20,28,36,44,52\}$, we copy the cache and generate block $m+1$ with matched initial noise and intermediate re-noising across conditions. Denoising reads the base 12-frame context, including sink KV. The generated probes do not enter subsequent reconstruction.

We measure opening similarity as the mean cosine similarity of three generated-frame DINOv2 embeddings to the normalized mean opening embedding from the reference video. We sample the generated block at decoded frames 0, 5, and 11, and the reference opening every 0.25 seconds from 0.5 seconds up to, but excluding, 2 seconds. This is full-frame appearance similarity, distinct from the own-minus-other-seed Consistency diagnostic. We average over probe positions and generation seeds within each prompt and use 4,000 paired prompt-bootstrap draws. Slopes are fitted within prompts against $\log_2 d$ before averaging.

Table~\ref{tab:committing-recency-fixed-trajectory} shows a small negative opening-similarity slope with the trajectory fixed, although the means are not strictly monotonic. Sink context lowers opening similarity further.

\begin{table}[!htbp]
\centering
\begin{minipage}{0.65\linewidth}
\centering
\footnotesize
\setlength{\tabcolsep}{5pt}
\renewcommand{\arraystretch}{1.10}
\caption{Opening similarity with a fixed SGF trajectory.}
\label{tab:committing-recency-fixed-trajectory}
\begin{tabular*}{\linewidth}{@{\extracolsep{\fill}}lc@{}}
\reporttoprule
\textbf{Context} & \textbf{Opening similarity} \\
\midrule
$d=1$ & 0.9347 \\
$d=2$ & 0.9351 \\
$d=4$ & 0.9344 \\
$d=8$ & 0.9339 \\
$d=16$ & 0.9338 \\
Sink & 0.9295 \\
\bottomrule
\end{tabular*}
\par\medskip
\begin{tabular*}{\linewidth}{@{\extracolsep{\fill}}lc@{}}
\reporttoprule
\textbf{Contrast} & \textbf{$\boldsymbol{\Delta}$ [95\% CI]} \\
\midrule
$d=2$ minus $d=1$ & \shortstack{+0.0004\\{}[\textminus{}0.0001, +0.0009]} \\
$d=4$ minus $d=1$ & \shortstack{\textminus{}0.0003\\{}[\textminus{}0.0011, +0.0003]} \\
$d=8$ minus $d=1$ & \shortstack{\textminus{}0.0008\\{}[\textminus{}0.0019, +0.0001]} \\
$d=16$ minus $d=1$ & \shortstack{\textminus{}0.0010\sigmark\\$[-0.0020, -3.5\times10^{-5}]$} \\
Sink minus $d=1$ & \shortstack{\textminus{}0.0053\sigmark\\{}[\textminus{}0.0072, \textminus{}0.0034]} \\
Sink minus $d=16$ & \shortstack{\textminus{}0.0043\sigmark\\{}[\textminus{}0.0063, \textminus{}0.0025]} \\
Slope per doubling of $d$ & \shortstack{\textminus{}0.0003\sigmark\\{}[\textminus{}0.0006, \textminus{}0.0001]} \\
\bottomrule
\end{tabular*}
\par\smallskip
\parbox{\linewidth}{\footnotesize\raggedright Each prompt contributes both generation seeds and all 5 probe positions. Values are on the original cosine-similarity scale. Paired intervals use 4,000 prompt-bootstrap draws. $^{*}$Interval excludes 0 before rounding.}
\end{minipage}
\end{table}

\subsection{Propagation and duration of committing effects}
\label{app:committing-propagation}

The SGF interventions use released EMA weights, a 12-frame physical cache storing KV from 3 sink frames, top-aligned positions within the 21-frame training range, a 4-step sampler, and committing at $t=0$. Each nominal 48-second video contains 64 blocks of three latent frames. The denoising context contains KV from 3 sink and 3 prior recent frames alongside 3 current latent frames throughout, and the committing context contains KV from 6 prior recent frames alongside 3 current latent frames. Sink interventions replace KV from 3 recent frames with sink KV at a fixed 9-frame context size.

\paragraph{A single committing intervention with fixed content.}
We fix the clean latents from videos generated with the 9-frame recent-only committing context for 20 prompts and generation seed 0, then reconstruct two separate KV caches using committing passes alone. The control always uses this 9-frame context. In the intervention, we use sink context only during the committing pass for block $b^*=20$ or 40. The self-only committing control uses three current frames for regular committing passes and adds KV from three sink frames at $b^*$. The same fixed latent block is committed in both caches in every subsequent pass. At lag $k$, we compare K and V for later KV entries that remain cached after committing block $b^*+k$. For isolated generation probes, we copy each cache and generate the next block with paired initial noise and re-noising, using the preceding fixed block as VAE context. These generated probes do not enter subsequent cache reconstruction.

Let $\mathcal J_k$ contain the cached KV tokens from blocks strictly after $b^*$ at lag $k\ge1$. For $T\in\{K,V\}$, let $T_\ell$ and $\widetilde T_\ell$ denote the control and intervention tensors in layer $\ell$. We compute
\[
\delta_k^T=\frac{1}{30}\sum_{\ell=1}^{30}
\frac{\|\widetilde T_\ell[\mathcal J_k]-T_\ell[\mathcal J_k]\|_2}
{\max\!\left(\|T_\ell[\mathcal J_k]\|_2,10^{-12}\right)}.
\]
The norm is taken over all selected tokens, heads, and channels. K and V are normalized separately within each layer before averaging over the 30 layers. We then average these values over the 20 prompts and two intervention positions $b^*\in\{20,40\}$.

\begin{figure}[!htbp]
\centering
\begin{minipage}{.65\linewidth}
\centering
\includegraphics[width=\linewidth]{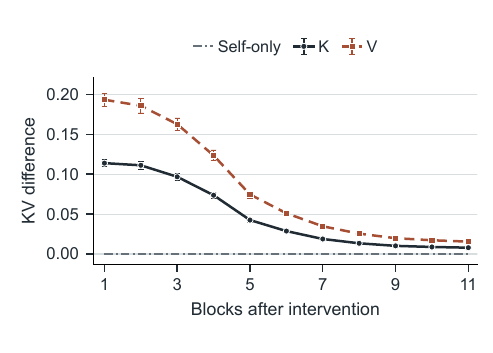}
\caption{A single sink read changes subsequent K and V after the intervened entry leaves the cache. Clean latents are fixed, and regular committing reads two preceding blocks. Error bars show prompt-bootstrap intervals.}
\label{fig:committing-impulse}
\end{minipage}
\end{figure}

Figure~\ref{fig:committing-impulse} shows nonzero KV differences after the directly affected entries leave the cache, decaying over subsequent blocks. With self-only committing on the same fixed latents, later-block KV and sampled probe outputs are identical to their controls. Table~\ref{tab:committing-impulse} reports the output probes. With the 9-frame context, small absolute output differences remain at probe lags $k=5,11$. The confidence intervals for signed saturation and brightness differences include zero.

\paragraph{Consecutive committing passes with sink context during free generation.}
We also intervene during free SGF generation on 20 prompts with generation seeds 0 and 1, pairing initial noise and intermediate re-noising. Committing reads sink context in blocks 16--31 and recent context afterward. Table~\ref{tab:committing-exposure-sgf} reports window means and isolated fixed-content probes. The probes reconstruct caches on control latents, use fresh paired noise, and decode with the fixed preceding block. Free generation keeps its rollout noise sequence and full-trajectory decoding. In blocks 36--43 and later windows, signed saturation and brightness intervals include zero, although absolute trajectory differences remain.

The SF curves in Figure~\ref{fig:history-effect-summary} use 160-frame trajectory context for 24 seconds and normalized saturation. These SGF probes use 3 sink frames for 12 seconds and raw color differences, so their tail readouts have different scales.

All intervals in Table~\ref{tab:committing-impulse} use 4,000 prompt-cluster bootstrap draws. For the fixed-content single intervention, we retain both intervention blocks within each prompt cluster. For free generation, we retain both generation seeds. Saturation and brightness differences use their original 0--1 scales.

\begin{table}[!htbp]
\centering
\footnotesize
\setlength{\tabcolsep}{4pt}
\renewcommand{\arraystretch}{1.10}
\caption{SGF output probes after one sink read or sink reads in blocks 16--31. Denoising is unchanged.}
\label{tab:committing-impulse}
\label{tab:committing-exposure-sgf}

\par\medskip
\begin{tabular*}{\linewidth}{@{\extracolsep{\fill}}cccccc@{}}
\reporttoprule
\textbf{Lag / block} & \textbf{Latent rel. L2} & \textbf{Later K rel. L2} & \textbf{Pixel MAE} & \textbf{$\boldsymbol{\Delta}$ saturation} & \textbf{$\boldsymbol{\Delta}$ brightness} \\
\midrule
\multicolumn{6}{@{}l}{\textit{One sink read: next-block probes at cache lag $k$}} \\
0 & 0.2894 & --- & 13.436 & \shortstack{\textminus{}0.00256\\{}[\textminus{}0.00458,\textminus{}0.00059]} & \shortstack{\textminus{}0.00104\\{}[\textminus{}0.00257,+0.00031]} \\
2 & 0.1171 & --- & 5.225 & \shortstack{+0.00081\\{}[+0.00043,+0.00119]} & \shortstack{\textminus{}0.00158\\{}[\textminus{}0.00198,\textminus{}0.00119]} \\
5 & 0.0306 & --- & 1.449 & \shortstack{\textminus{}0.00006\\{}[\textminus{}0.00020,+0.00007]} & \shortstack{+0.00005\\{}[\textminus{}0.00005,+0.00015]} \\
11 & 0.0172 & --- & 0.848 & \shortstack{+0.00002\\{}[\textminus{}0.00004,+0.00009]} & \shortstack{+0.00000\\{}[\textminus{}0.00004,+0.00005]} \\
\addlinespace[3pt]
\multicolumn{6}{@{}l}{\textit{Sink reads in blocks 16--31: free-generation block ranges}} \\
16--31 & 0.548 & --- & 28.52 & \shortstack{+0.00607\\{}[+0.00036,+0.01245]} & \shortstack{\textminus{}0.00848\\{}[\textminus{}0.01364,\textminus{}0.00364]} \\
32--35 & 0.574 & --- & 29.50 & \shortstack{+0.00663\\{}[+0.00068,+0.01366]} & \shortstack{\textminus{}0.00496\\{}[\textminus{}0.00895,\textminus{}0.00119]} \\
36--43 & 0.444 & --- & 21.69 & \shortstack{\textminus{}0.00001\\{}[\textminus{}0.00444,+0.00426]} & \shortstack{+0.00089\\{}[\textminus{}0.00107,+0.00325]} \\
44--51 & 0.408 & --- & 20.04 & \shortstack{\textminus{}0.00029\\{}[\textminus{}0.00427,+0.00346]} & \shortstack{+0.00006\\{}[\textminus{}0.00216,+0.00271]} \\
52--63 & 0.404 & --- & 19.61 & \shortstack{\textminus{}0.00127\\{}[\textminus{}0.00342,+0.00062]} & \shortstack{\textminus{}0.00042\\{}[\textminus{}0.00347,+0.00283]} \\
\addlinespace[3pt]
\multicolumn{6}{@{}l}{\textit{Sink reads in blocks 16--31: fixed-content probe blocks}} \\
24 & --- & --- & 13.40 & \shortstack{+0.00069\\{}[\textminus{}0.00094,+0.00257]} & \shortstack{\textminus{}0.00180\\{}[\textminus{}0.00335,\textminus{}0.00028]} \\
32 & --- & --- & 14.19 & \shortstack{+0.00022\\{}[\textminus{}0.00128,+0.00190]} & \shortstack{\textminus{}0.00266\\{}[\textminus{}0.00382,\textminus{}0.00158]} \\
34 & --- & 0.1485 & 5.77 & \shortstack{+0.00039\\{}[\textminus{}0.00023,+0.00106]} & \shortstack{\textminus{}0.00157\\{}[\textminus{}0.00220,\textminus{}0.00099]} \\
37 & --- & 0.0530 & 1.76 & \shortstack{\textminus{}0.00014\\{}[\textminus{}0.00032,+0.00003]} & \shortstack{\textminus{}0.00011\\{}[\textminus{}0.00028,+0.00004]} \\
43 & --- & 0.0081 & 0.90 & \shortstack{+0.00004\\{}[\textminus{}0.00003,+0.00010]} & \shortstack{\textminus{}0.00003\\{}[\textminus{}0.00009,+0.00003]} \\
\addlinespace[3pt]
\bottomrule
\end{tabular*}
\par\smallskip
\parbox{\linewidth}{\footnotesize\raggedright Cache lag $k$ is measured after committing $b^*+k$; the next-block probe generates $b^*+k+1$. The intervened KV is evicted at cache lag 3, or during the probe at lag 2. Next-block probes use the 9-frame recent-only committing context. Self-only probe differences are exactly 0 at sampled lags 2, 5, and 11. For sink reads in blocks 16--31, committing uses recent trajectory context afterward. Free-generation rows report window means; fixed-content probes reconstruct caches on control latents with the same committing schedule. Later K is measured in the reconstructed cache preceding the probe. All changes are intervention minus control. Brackets show 95\% confidence intervals, and Pixel MAE uses the 0--255 scale.}
\end{table}

\subsection{Committing as a recurrence}
\label{app:committing-recurrence}

\paragraph{Dependence along the chain.}
We fix the generated clean blocks \(x_0,\dots,x_b\), the prompt, and the temporal-position assignments. Let \(\mathrm{KV}^{(\ell)}_b\) denote the keys and values of block \(b\) in layer \(\ell\in\{0,\dots,L-1\}\), with \(L=30\) for the SGF implementation below. KV is projected from the input to each layer, before that layer's attention. Under Commit Forcing, historical attention during committing in layer \(\ell\) attends only to \(\mathrm{KV}^{(\ell)}_{b-1}\).

\textbf{Proposition.} With these inputs fixed, Commit Forcing satisfies
\[
\frac{\partial\mathrm{KV}_b}{\partial\mathrm{KV}_j}
=J_bJ_{b-1}\cdots J_{j+1},\qquad j<b,
\]
where \(J_k=\partial\mathcal{F}^{\mathrm c}_{\theta}(x_k;\mathrm{KV}_{k-1})/\partial\mathrm{KV}_{k-1}\) for the chain in Eq.~\ref{eq:kv-chain}. Moreover, \(\mathrm{KV}^{(\ell)}_b\) depends only on \(x_{\max(0,b-\ell)},\dots,x_b\).

\textit{Proof.} Setting \(R_b=\{b-1\}\) in Eq.~\ref{eq:history-committing} makes \(\mathrm{KV}_{b-1}\) the only historical input to the committing pass for block \(b\), so the derivative follows by the chain rule. For the layerwise bound, \(\mathrm{KV}^{(0)}_b\) depends only on \(x_b\). Suppose the bound holds for every layer \(\ell'<\ell\) and every block. The input to layer \(\ell\) of block \(b\) depends on \(x_b\) and on \(\mathrm{KV}^{(\ell')}_{b-1}\) for \(\ell'<\ell\). By the induction hypothesis, those entries depend only on blocks from \(\max(0,b-\ell)\) through \(b-1\). The bound follows. \(\square\)

Self-only committing has no historical input, so its derivatives with respect to earlier entries vanish under the same fixed-block condition. With KV from \(h_0\) preceding blocks, the corresponding bound reaches back to \(\max(0,b-h_0\ell)\). At startup, only available entries are read. Under base committing, the persistent sink KV is also read in every later committing pass, introducing dependence on block 0 from layer 1 onward. In free generation, older historical context can also affect \(x_b\) through denoising.

\paragraph{Fixed-latent measurement on SGF.}
We use the released SGF EMA weights, a 12-frame physical cache with slots for 3 sink, 6 recent, and 3 current frames, top-aligned positions, and committing at \(t=0\). We use the 20 diagnostic prompts of Section~\ref{app:committing-recency-fixed-trajectory}. With generation seed 0, we generate 60-second latent trajectories of 80 blocks using base denoising and Commit Forcing, then fix the clean latents. We rebuild each cache using committing passes alone, without denoising or VAE decoding. The 4 committing policies are Commit Forcing (6 frames), a 9-frame context with KV from two preceding blocks, base committing over all 12 frames including the sink, and self-only committing (3 frames).

For each policy, we compare the control reconstruction with three interventions:
\begin{itemize}
\item \textit{Content change.} Block \(b^*\in\{20,40\}\) is committed from the latent of block \((b^*+40)\bmod80\) in the same video. All other block inputs are unchanged. We test Commit Forcing, base committing, and self-only committing.
\item \textit{Sink change.} Block 0, which also supplies the persistent sink, is committed from block 40's latent under the same 3 policies.
\item \textit{Single sink read.} The sink is read in one committing pass at \(b^*\in\{20,40\}\), after which the selected policy resumes. The context of this pass contains 3 sink and 3 current frames under Commit Forcing instead of the usual 6 frames, and 3 sink, 3 recent, and 3 current frames under the 9-frame control. These 2 contrasts keep the context size fixed. For self-only committing, the context of the intervened pass grows from 3 current frames to 3 sink and 3 current frames.
\end{itemize}
For the content change and the single sink read, the paired caches are copied immediately before \(b^*\), and we measure lags 1--32. For the sink change, we measure blocks 1--79. The directly modified entry is excluded from these readouts.

For \(T\in\{K,V\}\), let \(T^{A}_{m,\ell}\) and \(T^{B}_{m,\ell}\) be the control and intervention tensors over block \(m\)'s own slots in layer \(\ell\). We compute the dimensionless relative difference
\[
\delta^T_{m,\ell}
=\frac{\|T^{B}_{m,\ell}-T^{A}_{m,\ell}\|_2}
{\max\!\left(\|T^{A}_{m,\ell}\|_2,10^{-12}\right)}.
\]
We compare the tensors in single precision, with the norm covering all selected tokens, heads, and channels. K and V are normalized separately. Layer-mean KV differences average the K and V values across all 30 layers. For the single sink read, we report the layer-mean V difference. For the content change and the single sink read, we average the two intervention positions within each prompt before averaging prompts. Intervals use 4,000 prompt-bootstrap draws, keeping both positions within each sampled prompt.

\begin{figure}[!htbp]
\centering
\includegraphics[width=\linewidth]{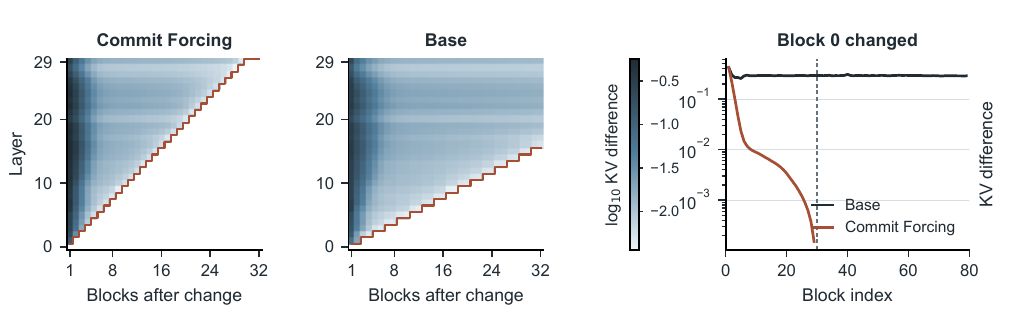}
\caption{Fixed-latent SGF probes after changing a block's content (left and middle) or block 0 (right). White cells are exactly zero. Lines on the heatmaps mark the layerwise dependence bounds.}
\label{fig:committing-recurrence}
\end{figure}

In Figure~\ref{fig:committing-recurrence}, every layer outside the predicted dependence bound matches the control. For entries that change, the first changed layer is the earliest permitted by the bound. After a content change, Commit Forcing first changes KV for block \(b^*+d\) in layer \(d\), giving no difference from lag 30 onward. Base committing first changes it in layer \(\lceil d/2\rceil\) and still differs by 0.0052 at lag 32.

After a sink change, the altered sink is read again in every block under base committing, and the mean difference over blocks 1--79 is 0.29. Under Commit Forcing, the difference is 0.42 at block 1, 0.023 at block 5, 0.0088 at block 10, and exactly 0 from block 30.

Figure~\ref{fig:committing-window-mechanisms} shows the single-sink-read response. The next entry's V differs by 0.28 with Commit Forcing, followed by 0.13, 0.055, and 0.028 over the next three blocks. The difference is exactly 0 from lag 29 onward. The cutoff differs from the content change because changing a committing read leaves the intervened block's layer-0 KV unchanged. With KV from two preceding blocks, the difference remains 0.007 at lag 32. Self-only committing gives exactly 0 for every subsequent entry in all three interventions.

\section{Experiment Details}
\label{app:implementation-details}
\label{app:result-details}

\subsection{Generators and baselines}
\label{app:model-configurations}
\hypertarget{instantiations}{}\label{instantiations}

We generate benchmark videos at 16 fps and 832 $\times$ 480 resolution. The main comparison includes 6 generators: SGF, SGF+, CF, LongLive, Rolling Sink, and RF-TF. We use released weights, pair outputs by prompt, and retain each base generator's cache and sampler configuration.

For all 6 generators in the standard-suite main comparison, the committing context contains the current 3-frame block and the stored KV of the immediately preceding 3-frame block, for 6 latent frames in total. On CF, the historical KV comes from the three committed frames in the fast FIFO. The denoising policy is unchanged. The comparisons on Rolling Forcing's 200-prompt selection~\citep{liu2025rollingforcing} use the same 6-frame context, including SF with a sink. Wider-window references use 9 frames on SGF and SGF+, 21 on RF-TF, Rolling Sink, and SF with a sink, and 12 on LongLive. The ID-Forcing adaptations use 9 frames.

The supporting SF comparisons add a sink to the base SF configuration. On CF, we exclude sink KV, slow memory, and stale slots from committing, with separate sink-only and slow-memory-only exclusions as ablations. Stale slots hold non-promoted historical KV outside the fast FIFO while slow memory is still filling.

For newly generated context controls, we match initial noise across conditions for each prompt. Reused SGF, RF-TF, and CF outputs are paired by prompt and generation seed. Intermediate re-noising cannot be verified for reused SGF outputs, so the standard-suite comparison retains prompt-level pairing.

The standard suite uses Wan2.1-augmented generation prompts for all 946 official VBench entries, corresponding to 944 distinct original prompt strings. Each arm generates one video per entry with generation seed 0 and 957 RGB frames. We preserve the original metadata, including dimension assignments for repeated prompts, and evaluate all 16 dimensions on their designated subsets.

The standard-suite SGF+ comparison uses the released chunkwise configuration with a 12-frame cache, 3 sink frames, a 12-frame training-position range, and contiguous temporal positions. We retain SGF+'s separate committing parameters. The supporting comparisons on Rolling Forcing's prompts use 6- and 9-frame committing contexts, a 21-frame training-position range, and top-aligned positions while retaining a 12-frame cache.

In the standard-suite LongLive comparison, we retain the base 12-frame physical cache, including the 3-frame sink KV. The denoising context contains the sink and the newest 9 trajectory frames, whereas the committing context of Commit Forcing contains only the newest 6, comprising 3 prior recent and 3 current frames. The 12-frame committing-window ablation uses a 15-frame KV cache to store the sink alongside KV from 12 trajectory frames. In the comparisons using Rolling Forcing's prompts and the long-video comparisons, both committing policies use a 15-frame physical cache.

We evaluate complete ID-Forcing and its 9-frame adaptation with the same sink KV storage, conditioning window, and position re-rotation. On SF, we retain the base KV for blocks 5 and 6 at handover and omit ID-Forcing's per-block tail recaching, changing both committing and cache handover. In both adaptations, the committing context contains the 3-frame current block and KV from 6 recent historical frames. Forward-pass counts for these adaptations and cache capacities for cost measurements appear in Section~\ref{app:computational-cost}.

For the separate 120- and 240-second comparisons on ID-Forcing's 128-prompt MovieGen list, we use generation seed 0 with our per-prompt noise initialization. We evaluate the 6-frame LongLive committing context at both durations and include the 12-frame context as an ablation. The 240-second LongLive comparisons contain 128 complete videos per arm, with 960 latent frames per video.

\paragraph{RF implementation details.}
\label{app:rf-configurations}
As listed in Table~\ref{tab:rf-configurations}, we distinguish the SF-based RF-TF used in the main evaluation, RF denoising on SGF weights, and an alternative SF-based implementation using the SGF cache geometry. We use the released EMA weights for the DMD version of SF and the chunkwise version of SGF. All three use blocks of three latent frames and the same Temporal Response Bias. For a non-sink historical block at distance $d$ from the current block and denoising-step index $s$, the implemented bias is
\begin{equation}
\begin{gathered}
B(d,s)=-4\left[\min\!\left(1,\frac{\max(d-1,0)}{4}\right)\right]^{0.1\,2^{s+1}},\\
s\in\{0,1,2,3\}.
\end{gathered}
\label{eq:rf-implemented-bias}
\end{equation}
The bias applies to non-sink context with $d>1$ in all 30 DiT layers and all 4 denoising passes. The sink, preceding block, and current block receive zero bias. The raw schedule $[1000,750,500,250]$ uses a shift of 5, giving approximately $[1000,937.5,833.33,625]$. The bias is off during committing at $t=0$.

\begin{table}[!htbp]
\centering
\begin{minipage}{0.80\linewidth}
\caption{RF implementations and historical-context layouts. Context counts include the current 3-frame block. The main RF-TF comparison uses the 6-frame rule. Wider contexts are reference controls.}
\label{tab:rf-configurations}
\centering
\footnotesize
\setlength{\tabcolsep}{5pt}
\renewcommand{\arraystretch}{1.12}
\begin{tabular*}{\linewidth}{@{\extracolsep{\fill}}lccc@{}}
\reporttoprule
\textbf{Implementation} & \textbf{Weights} & \textbf{Stored frames} & \textbf{Recent} \\
\midrule
RF-TF main & SF EMA & 24 & 0+6 \\
RF-TF control & SF EMA & 24 & 0+21 \\
RF on SGF & SGF EMA & 21 & 0+18 \\
Alternative RF on SF & SF EMA & 21 & 0+18 \\
\bottomrule
\end{tabular*}
\par\smallskip
\parbox{\linewidth}{\footnotesize\raggedright Stored frames gives physical-cache capacity; Recent gives the recent committing context. Denoising and base committing both use 3+18 frames in all implementations. RF-TF uses absolute RoPE for trajectory frames, with the sink phase shifted to precede the recent window. RF on SGF and the alternative RF on SF use the SGF positional layout, with a 21-frame position range, sink positions 0--2, and steady-state trajectory positions 3--20. Temporal Response Bias is off during committing in all implementations.}
\end{minipage}
\end{table}

\paragraph{Baseline sources.}
\label{app:published-quality-sources}
Table~\ref{tab:published-long-comparison} lists the sources and evaluation protocols for the reproduced baselines.
\begin{table}[!htbp]
\caption{Sources and protocols for reproduced baselines.}
\label{tab:published-long-comparison}
\centering
\footnotesize
\setlength{\tabcolsep}{6pt}
\renewcommand{\arraystretch}{1.12}
\begin{tabularx}{\linewidth}{@{}>{\raggedright\arraybackslash}p{.13\linewidth}>{\raggedright\arraybackslash}p{.30\linewidth}>{\raggedright\arraybackslash}X@{}}
\reporttoprule
\textbf{Baseline} & \textbf{Source} & \textbf{Evaluation details} \\
\midrule
SF with sink & Self-Forcing weights~\citep{huang2025selfforcing} and RECAP-Forcing's plain-sink reference~\citep{xu2026recapforcing}. & The published reference uses 128 prompts, 5 seeds and a 6-frame cache. Our configuration on Rolling Forcing's prompts uses a 24-frame physical cache and a 6-frame committing context, with the 21-frame context retained as a window reference. \\
\addlinespace[5pt]
RF-TF & RF's training-free configuration and Table~3~\citep{cao2026recencyforcing}. & The reported evaluation uses 200 MovieGen prompts at 60 seconds. Exact consistency-fusion and Temporal Flickering filtering settings are unspecified. \\
\addlinespace[5pt]
LongLive & Released LongLive weights~\citep{yang2026longlive} and RF's Table~3 evaluation~\citep{cao2026recencyforcing}. & We evaluate the released weights under our prompt protocols, with cache implementations specified in Section~\ref{app:model-configurations}. \\
\addlinespace[5pt]
Rolling Sink & Released method~\citep{li2026rollingsink} and RECAP-Forcing's evaluation~\citep{xu2026recapforcing}. & The original paper reports components and average rank. RECAP-Forcing's reference uses 128 prompts and 5 seeds, rather than Rolling Forcing's prompt set. \\
\addlinespace[5pt]
CF & Released Context Forcing weights and Tables~2--3~\citep{chen2026contextforcing}. & Published 60-second results use official extended VBench prompts, rather than Rolling Forcing's prompt list. \\
\addlinespace[5pt]
SGF / SGF+ & SGF Table~2 and SGF+ Table~1~\citep{zhuang2026selfgradientforcing,su2026sgfplus}. & The papers report 7 Quality components without a composite Quality. Exact prompt lists and consistency-fusion settings are not established by the released evaluation materials. \\
\bottomrule
\end{tabularx}
\end{table}

\subsection{Evaluation protocols and scoring}
\label{app:evaluation-protocols}
\label{app:benchmark-scoring}

We display benchmark scores on a 0--100 scale and compute all differences and confidence intervals before rounding. Unless stated otherwise, paired comparisons use prompt-bootstrap 95\% intervals. An asterisk on a paired difference indicates that its interval excludes 0.
We use SC and BC for fused Subject and Background Consistency, TF for Temporal Flickering, MS for Motion Smoothness, DD for Dynamic Degree, AQ for Aesthetic Quality, and IQ for Imaging Quality. Subject$_{\mathrm{c2c}}$ and Background$_{\mathrm{c2c}}$ denote their clip-to-clip components. Semantic abbreviations are OC (object class), MO (multiple objects), HA (human action), Col (color), SR (spatial relationship), Sc (scene), AS (appearance style), TS (temporal style), and OCon (overall consistency). Window sizes count the current latent block unless stated otherwise.

We follow VBench-Long's standard evaluation protocol, using dimension-specific clip lengths and static filtering for Temporal Flickering. Scores use the official clip and video aggregation, denominators based on valid outputs, Subject/Background fusion, and normalization. Total combines Quality and Semantic in a 4:1 ratio.

For paired 95\% intervals, we use 4,000 joint bootstrap draws over the 944 distinct original prompt strings. Each draw resamples prompt clusters across dimensions and uses the official aggregation weights of the valid outputs in each arm. Subject Consistency and Background Consistency use their designated subsets of 72 and 86 prompts. Temporal Flickering filtering and invalid Color outputs can give different numbers of valid videos across the paired arms.

For custom prompts, we reconstruct Quality from Subject Consistency, Background Consistency, Temporal Flickering, Motion Smoothness, Dynamic Degree, Aesthetic Quality, and Imaging Quality. Subject and Background Consistency combine per-video within-clip scores and mapped clip-to-clip scores with equal weights. We also report the clip-to-clip Subject Consistency component before mapping as Subject$_{\mathrm{c2c}}$.

We use VBench's published min-max normalization and weights. We weight Dynamic Degree by 0.5 and each other Quality dimension by 1. For custom-input reconstruction, we divide unnormalized MUSIQ Imaging Quality scores by 100 before using them in the Quality formula. Standard-suite Imaging Quality aggregates are already normalized to 0--1.

We evaluate custom-prompt Temporal Flickering without static filtering and report this 7-dimension Quality score separately from standard-suite Total.

For custom prompts, we compute intervals using 4,000 paired prompt-bootstrap draws. All intervals condition on the evaluated weights and generation seeds and do not estimate variation across training runs.

We use Self-Forcing's released expanded prompts~\citep{huang2025selfforcing} for the comparisons on Rolling Forcing's 200-prompt selection. This set differs from the first 200 extended MovieGen prompts.

\subsection{Benchmark results}
\label{app:benchmark-results}
\suppressfloats[t]
\begingroup

\begin{table}[!htbp]
\centering
\begin{minipage}{0.65\linewidth}
\caption{Commit Forcing minus base changes on the 60-second standard suite. Intervals use 4,000 paired bootstrap draws over 944 prompt clusters.}
\label{tab:standard-main-intervals}
\centering
\footnotesize
\setlength{\tabcolsep}{2pt}
\renewcommand{\arraystretch}{1.05}
\begin{tabular*}{\linewidth}{@{\extracolsep{\fill}}lcc@{}}
\reporttoprule
\textbf{Generator} & \textbf{$\boldsymbol{\Delta}$Quality} & \textbf{$\boldsymbol{\Delta}$Total} \\
\midrule
CF & \shortstack{+0.46\\{}[+0.09,\,+0.85]} & \shortstack{+0.44\\{}[+0.10,\,+0.81]} \\
Rolling Sink & \shortstack{+0.41\\{}[+0.15,\,+0.69]} & \shortstack{+0.36\\{}[+0.10,\,+0.63]} \\
RF-TF & \shortstack{+0.65\\{}[+0.27,\,+1.01]} & \shortstack{+1.10\\{}[+0.74,\,+1.45]} \\
LongLive & \shortstack{+0.29\\{}[+0.05,\,+0.52]} & \shortstack{+0.31\\{}[+0.10,\,+0.52]} \\
SGF+ & \shortstack{+0.32\\{}[\textminus{}0.15,\,+0.81]} & \shortstack{+0.34\\{}[\textminus{}0.07,\,+0.76]} \\
SGF & \shortstack{+1.27\\{}[+0.76,\,+1.88]} & \shortstack{+0.98\\{}[+0.45,\,+1.53]} \\
\bottomrule
\end{tabular*}
\end{minipage}
\end{table}

\paragraph{Standard-suite components.}
\label{app:standard-components}
Tables~\ref{tab:standard-main-intervals}--\ref{tab:standard-semantic-aggregate} report aggregate scores and intervals, component scores, and the weighted Quality decomposition.

For the 7 contributions in Table~\ref{tab:quality-decomposition}, we multiply each unrounded component change by its official normalization factor and weight, then divide by the total Quality weight of 6.5. Dynamic Degree dominates the SGF Quality gain, while Aesthetic and Imaging Quality contribute most on RF-TF.

\begin{table}[!htbp]
\centering
\begin{minipage}{0.80\linewidth}
\caption{Weighted component contributions to the standard-suite Quality change.}
\label{tab:quality-decomposition}
\centering
\footnotesize
\setlength{\tabcolsep}{2pt}
\renewcommand{\arraystretch}{1.05}
\begin{tabular*}{\linewidth}{@{\extracolsep{\fill}}lcccccc@{}}
\reporttoprule
\textbf{Metric} & \textbf{CF} & \textbf{RS} & \textbf{RF-TF} & \textbf{LL} & \textbf{SGF+} & \textbf{SGF} \\
\midrule
$\boldsymbol{\Delta}$Quality & +0.46 & +0.41 & +0.65 & +0.29 & +0.32 & +1.27 \\
SC & +0.04 & +0.03 & +0.11 & +0.01 & +0.05 & +0.05 \\
BC & +0.06 & +0.06 & +0.16 & +0.03 & +0.13 & +0.10 \\
TF & \textminus{}0.04 & \textminus{}0.02 & +0.04 & 0.00 & +0.04 & +0.02 \\
MS & \textminus{}0.06 & \textminus{}0.06 & \textminus{}0.03 & \textminus{}0.07 & \textminus{}0.05 & \textminus{}0.15 \\
DD & +0.26 & +0.27 & \textminus{}0.22 & +0.21 & +0.12 & +1.01 \\
AQ & +0.06 & +0.08 & +0.37 & +0.06 & +0.04 & +0.18 \\
IQ & +0.13 & +0.06 & +0.23 & +0.07 & \textminus{}0.02 & +0.08 \\
\bottomrule
\end{tabular*}
\par\smallskip
\parbox{\linewidth}{\footnotesize RS and LL denote Rolling Sink and LongLive. Unrounded contributions sum to the aggregate Quality change. Other abbreviations and normalization follow Section~\ref{app:benchmark-scoring}.}
\end{minipage}
\end{table}

\begin{table}[!htbp]
\caption{Quality component scores for Base and Commit Forcing on the 60-second standard suite (0--100 scale). Abbreviations are defined in Section~\ref{app:benchmark-scoring}.}
\label{tab:standard-quality-components}
\centering
\footnotesize
\setlength{\tabcolsep}{3pt}
\renewcommand{\arraystretch}{1.12}
\begin{tabular*}{\linewidth}{@{\extracolsep{\fill}}l*{12}{c}@{}}
\reporttoprule
 & \multicolumn{2}{c}{\textbf{CF}} & \multicolumn{2}{c}{\textbf{Rolling Sink}} & \multicolumn{2}{c}{\textbf{RF-TF}} & \multicolumn{2}{c}{\textbf{LongLive}} & \multicolumn{2}{c}{\textbf{SGF+}} & \multicolumn{2}{c}{\textbf{SGF}} \\
\cmidrule(lr){2-3}\cmidrule(lr){4-5}\cmidrule(lr){6-7}\cmidrule(lr){8-9}\cmidrule(lr){10-11}\cmidrule(lr){12-13}
\textbf{Metric} & \textbf{Base} & \textbf{Commit} & \textbf{Base} & \textbf{Commit} & \textbf{Base} & \textbf{Commit} & \textbf{Base} & \textbf{Commit} & \textbf{Base} & \textbf{Commit} & \textbf{Base} & \textbf{Commit} \\
\midrule
SC & 97.75 & 97.96 & 98.09 & 98.28 & 96.73 & 97.36 & 97.96 & 98.00 & 98.51 & 98.81 & 98.26 & 98.51 \\
BC & 96.63 & 96.92 & 97.07 & 97.34 & 95.92 & 96.70 & 97.11 & 97.23 & 96.93 & 97.53 & 96.69 & 97.17 \\
TF & 99.30 & 99.21 & 99.11 & 99.06 & 98.88 & 98.96 & 99.29 & 99.28 & 99.23 & 99.32 & 98.95 & 98.99 \\
MS & 98.55 & 98.43 & 98.77 & 98.66 & 98.37 & 98.31 & 98.69 & 98.55 & 98.75 & 98.67 & 98.63 & 98.35 \\
DD & 58.89 & 62.31 & 53.94 & 57.41 & 70.28 & 67.36 & 52.13 & 54.81 & 64.86 & 66.48 & 67.45 & 80.56 \\
AQ & 65.41 & 65.82 & 65.53 & 66.02 & 62.65 & 65.03 & 66.30 & 66.71 & 67.86 & 68.10 & 67.06 & 68.20 \\
IQ & 69.47 & 70.32 & 70.88 & 71.28 & 70.34 & 71.82 & 70.03 & 70.46 & 72.17 & 72.04 & 71.82 & 72.32 \\
\bottomrule
\end{tabular*}
\end{table}

\begin{table}[!htbp]
\caption{Semantic component scores for Base and Commit Forcing on the 60-second standard suite (0--100 scale). Abbreviations are defined in Section~\ref{app:benchmark-scoring}.}
\label{tab:standard-semantic-components}
\centering
\footnotesize
\setlength{\tabcolsep}{3pt}
\renewcommand{\arraystretch}{1.12}
\begin{tabular*}{\linewidth}{@{\extracolsep{\fill}}l*{12}{c}@{}}
\reporttoprule
 & \multicolumn{2}{c}{\textbf{CF}} & \multicolumn{2}{c}{\textbf{Rolling Sink}} & \multicolumn{2}{c}{\textbf{RF-TF}} & \multicolumn{2}{c}{\textbf{LongLive}} & \multicolumn{2}{c}{\textbf{SGF+}} & \multicolumn{2}{c}{\textbf{SGF}} \\
\cmidrule(lr){2-3}\cmidrule(lr){4-5}\cmidrule(lr){6-7}\cmidrule(lr){8-9}\cmidrule(lr){10-11}\cmidrule(lr){12-13}
\textbf{Metric} & \textbf{Base} & \textbf{Commit} & \textbf{Base} & \textbf{Commit} & \textbf{Base} & \textbf{Commit} & \textbf{Base} & \textbf{Commit} & \textbf{Base} & \textbf{Commit} & \textbf{Base} & \textbf{Commit} \\
\midrule
OC & 88.63 & 87.23 & 89.97 & 89.85 & 88.05 & 90.49 & 89.18 & 88.52 & 87.61 & 87.89 & 86.00 & 90.82 \\
MO & 70.79 & 73.34 & 74.51 & 74.31 & 63.58 & 70.46 & 78.40 & 79.91 & 77.62 & 79.00 & 71.24 & 71.72 \\
HA & 91.33 & 93.67 & 91.17 & 90.67 & 92.83 & 94.67 & 94.83 & 95.50 & 91.83 & 93.33 & 92.33 & 91.50 \\
Col & 84.31 & 85.09 & 91.58 & 92.98 & 87.96 & 89.78 & 92.17 & 92.04 & 89.74 & 89.55 & 89.60 & 87.71 \\
SR & 69.19 & 67.40 & 72.81 & 73.46 & 73.15 & 75.76 & 75.80 & 76.30 & 79.78 & 77.27 & 76.29 & 74.20 \\
Sc & 40.20 & 39.47 & 42.81 & 43.63 & 35.47 & 43.01 & 44.77 & 45.71 & 47.11 & 49.55 & 48.09 & 44.73 \\
AS & 20.88 & 20.97 & 20.77 & 20.74 & 20.66 & 20.60 & 20.69 & 20.73 & 20.65 & 20.72 & 20.61 & 20.76 \\
TS & 20.83 & 21.10 & 22.85 & 22.71 & 21.96 & 22.36 & 22.49 & 22.65 & 22.44 & 22.54 & 22.05 & 22.35 \\
OCon & 23.60 & 23.89 & 24.98 & 24.92 & 24.59 & 24.77 & 25.08 & 25.09 & 24.82 & 24.84 & 24.84 & 24.97 \\
\bottomrule
\end{tabular*}
\end{table}

\begin{table}[!htbp]
\centering
\begin{minipage}{0.65\linewidth}
\caption{Aggregate Semantic scores on the 60-second standard suite (0--100 scale). Brackets give paired 95\% intervals over 944 prompt clusters.}
\label{tab:standard-semantic-aggregate}
\centering
\footnotesize
\setlength{\tabcolsep}{2pt}
\renewcommand{\arraystretch}{1.05}
\begin{tabular*}{\linewidth}{@{\extracolsep{\fill}}lccc@{}}
\reporttoprule
\textbf{Generator} & \textbf{Base} & \textbf{Commit Forcing} & \textbf{$\boldsymbol{\Delta}$Semantic} \\
\midrule
CF & 72.03 & 72.41 & \shortstack{+0.38\\{}[\textminus{}0.49,\,+1.25]} \\
Rolling Sink & 75.13 & 75.31 & \shortstack{+0.17\\{}[\textminus{}0.53,\,+0.88]} \\
RF-TF & 72.10 & 75.00 & \shortstack{+2.90\\{}[+1.97,\,+3.82]} \\
LongLive & 76.43 & 76.84 & \shortstack{+0.41\\{}[\textminus{}0.09,\,+0.91]} \\
SGF+ & 76.22 & 76.66 & \shortstack{+0.45\\{}[\textminus{}0.20,\,+1.13]} \\
SGF & 74.99 & 74.78 & \shortstack{\textminus{}0.21\\{}[\textminus{}1.62,\,+1.18]} \\
\bottomrule
\end{tabular*}
\end{minipage}
\end{table}

For the controls in Table~\ref{tab:standard-consistency-details}, we resample paired outputs within each designated prompt subset.

\begin{table}[!htbp]
\centering
\begin{minipage}{0.65\linewidth}
\caption{Clip-to-clip consistency changes in the standard-suite controls with 9-frame SGF, 21-frame RF-TF, and 6-frame CF committing.}
\label{tab:standard-consistency-details}
\centering
\footnotesize
\setlength{\tabcolsep}{3pt}
\renewcommand{\arraystretch}{1.04}
\begin{tabular*}{\linewidth}{@{\extracolsep{\fill}}lcc@{}}
\reporttoprule
\textbf{Generator} & $\boldsymbol{\Delta}$\textbf{ Subject$_{\mathrm{c2c}}$} & $\boldsymbol{\Delta}$\textbf{ Background$_{\mathrm{c2c}}$} \\
\midrule
SGF & +2.71 [+1.28,+4.08] & +4.82 [+4.05,+5.64] \\
RF-TF & +4.44 [+3.14,+5.67] & +4.37 [+3.78,+4.97] \\
CF & +2.35 [+1.22,+3.49] & +3.35 [+2.56,+4.18] \\
\bottomrule
\end{tabular*}
\end{minipage}
\end{table}

\paragraph{Rolling Forcing prompt selection.}
Table~\ref{tab:rollf-main} gives the paired results on Rolling Forcing's prompts for each generation seed.

\begin{table}[!htbp]
\caption{Base committing and 6-frame Commit Forcing on Rolling Forcing's 200 prompts at 60 seconds, for each generation seed. The CF sink-only row is a component control.}
\label{tab:rollf-main}
\label{tab:generation-seed-replication}
\centering
\footnotesize
\setlength{\tabcolsep}{2pt}
\renewcommand{\arraystretch}{1.12}
\begin{tabular*}{\linewidth}{@{\extracolsep{\fill}}lc*{7}{c}@{}}
\reporttoprule
& & \multicolumn{3}{c}{\textbf{Quality}} & \multicolumn{3}{c}{\textbf{Subject$_{\mathrm{c2c}}$}} & \textbf{DD} \\
\cmidrule(lr){3-5}\cmidrule(lr){6-8}\cmidrule(l){9-9}
\textbf{Generator} & \textbf{Seed} & \textbf{Base} & \textbf{Modified} & $\boldsymbol{\Delta}$ [95\% CI] & \textbf{Base} & \textbf{Modified} & $\boldsymbol{\Delta}$ [95\% CI] & $\boldsymbol{\Delta}$ [95\% CI] \\
\midrule
\multicolumn{9}{@{}l}{\textit{6-frame Commit Forcing}} \\
SGF & 0 & 83.58 & 84.51 & +0.93\sigmark\enspace [+0.55, +1.30] & 86.54 & 91.30 & +4.76\sigmark\enspace [+3.79, +5.73] & +11.33\sigmark\enspace [+6.53, +15.98] \\
 & 1 & 83.85 & 84.42 & +0.57\sigmark\enspace [+0.24, +0.91] & 86.60 & 91.52 & +4.91\sigmark\enspace [+3.92, +5.90] & +8.55\sigmark\enspace [+3.98, +13.07] \\
\addlinespace[2pt]
RF-TF & 0 & 82.01 & 82.43 & +0.42\sigmark\enspace [+0.18, +0.66] & 75.02 & 81.38 & +6.36\sigmark\enspace [+5.46, +7.24] & \textminus{}3.92\sigmark\enspace [\textminus{}6.75, \textminus{}1.17] \\
 & 1 & 82.00 & 82.69 & +0.69\sigmark\enspace [+0.46, +0.92] & 74.42 & 80.99 & +6.57\sigmark\enspace [+5.71, +7.46] & \textminus{}0.05\enspace [\textminus{}2.35, +2.23] \\
\addlinespace[2pt]
LongLive & 0 & 82.23 & 82.73 & +0.49\sigmark\enspace [+0.33, +0.66] & 86.01 & 87.21 & +1.20\sigmark\enspace [+0.84, +1.58] & +6.42\sigmark\enspace [+4.27, +8.58] \\
 & 1 & 82.19 & 82.66 & +0.46\sigmark\enspace [+0.30, +0.64] & 86.02 & 87.09 & +1.06\sigmark\enspace [+0.72, +1.43] & +6.62\sigmark\enspace [+4.53, +8.73] \\
\addlinespace[2pt]
CF & 0 & 82.78 & 82.91 & +0.13\enspace [\textminus{}0.10, +0.35] & 78.72 & 81.95 & +3.23\sigmark\enspace [+2.56, +3.94] & \textminus{}1.05\enspace [\textminus{}4.05, +1.92] \\
 & 1 & 82.70 & 83.12 & +0.43\sigmark\enspace [+0.22, +0.64] & 78.63 & 82.16 & +3.53\sigmark\enspace [+2.83, +4.26] & +1.33\enspace [\textminus{}1.45, +4.13] \\
\addlinespace[2pt]
Rolling Sink & 0 & 82.61 & 82.68 & +0.07\enspace [\textminus{}0.13, +0.27] & 86.00 & 88.45 & +2.45\sigmark\enspace [+1.81, +3.17] & +2.02\enspace [\textminus{}0.83, +4.73] \\
 & 1 & 82.40 & 82.66 & +0.26\sigmark\enspace [+0.07, +0.45] & 86.34 & 88.89 & +2.54\sigmark\enspace [+1.96, +3.18] & +4.05\sigmark\enspace [+1.83, +6.30] \\
\addlinespace[2pt]
SF with sink & 0 & 81.33 & 81.64 & +0.31\enspace [\textminus{}0.15, +0.84] & 75.18 & 79.62 & +4.44\sigmark\enspace [+3.60, +5.25] & \textminus{}0.93\enspace [\textminus{}3.92, +2.17] \\
 & 1 & 81.35 & 81.57 & +0.23\enspace [\textminus{}0.42, +0.80] & 74.71 & 78.87 & +4.16\sigmark\enspace [+3.33, +4.96] & +0.68\enspace [\textminus{}2.67, +4.07] \\
\midrule
\multicolumn{9}{@{}l}{\textit{CF sink-exclusion component control}} \\
CF & 0 & 82.78 & 83.11 & +0.32\sigmark\enspace [+0.11, +0.54] & 78.72 & 80.52 & +1.80\sigmark\enspace [+1.20, +2.42] & +6.70\sigmark\enspace [+4.08, +9.20] \\
 & 1 & 82.70 & 83.19 & +0.50\sigmark\enspace [+0.31, +0.69] & 78.63 & 80.24 & +1.61\sigmark\enspace [+1.02, +2.22] & +7.18\sigmark\enspace [+4.87, +9.52] \\
\bottomrule
\end{tabular*}
\par\smallskip
\parbox{\linewidth}{\footnotesize Each configuration and seed uses 200 paired prompts. Changes are Modified minus Base. Modified denotes Commit Forcing in the first group and sink exclusion in the component control. DD reports paired changes without absolute scores. Counts include the current block. Statistical definitions follow Section~\ref{app:benchmark-scoring}.}
\end{table}

\phantomsection
\label{app:rf-combination}

\paragraph{Consistency after Dynamic Degree adjustment.}
\label{app:sgf-six-frame-consistency}
For the regression-adjusted estimates in Table~\ref{tab:matched-motion}, we compute intervals using 2,000 prompt-bootstrap draws. The OLS estimates use paired videos in which motion is not held fixed.

\par
\endgroup

\subsection{Human evaluation}
\label{app:human-evaluation}

All participants gave informed consent to take part and to have their judgments used in this paper.
The evaluation did not undergo formal ethics review due to time constraints.

\paragraph{60-second comparisons.}
Each of the three comparisons contains 18 prompt-matched pairs of 60-second videos from Rolling Forcing's prompt set. We sample prompts without replacement before viewing the videos. SGF versus base SGF and LongLive versus base LongLive share one 18-prompt subset. The comparison with complete ID-Forcing uses another. Generation configurations follow Section~\ref{app:model-configurations}, with generation seed 0 for videos from the base generator configurations and the prompt index as the rollout seed for complete ID-Forcing. The SGF comparison uses the 9-frame reference. LongLive uses Commit Forcing.

For each video, participants view 5-second excerpts from the beginning, middle, and end. Paired excerpts play synchronously side by side, with method identities hidden and left-right positions randomized. Participants judge subject and scene consistency relative to each video's own opening, allowing normal changes in action, pose, and viewpoint, and give an overall preference based on prompt alignment, visual quality, and motion. Participants can select either video or a tie.\par

The 60-second results aggregate three response sets from 5 participants. Of these sets, two were completed individually and one jointly by the other three participants. Each video pair contributes three responses per criterion, with the joint response counted once. Table~\ref{tab:human-evaluation-counts} gives the pooled counts, and win rates are wins divided by wins plus losses, excluding ties. The 9-frame SGF reference receives win rates of 75.0\% for subject and scene consistency and 74.5\% for overall preference.

\paragraph{120-second LongLive comparison.}
We compare Commit Forcing with base LongLive using existing 120-second videos from ID-Forcing's 128-prompt MovieGen list and generation seed 0. A fixed random seed selects 30 prompt-matched pairs without replacement before packaging. Each participant sees a separately randomized order, with Commit Forcing on the left in 15 pairs and on the right in 15. Videos use the same three-excerpt presentation and questions described above. Five participants not involved in the project answer independently, giving 150 judgments per criterion. Two additional pairs show identical videos. The preregistered rule excludes a participant only if the consistency question receives a nontie response on both identical-video check pairs. All 5 participants are retained.

For the 120-second study, 4,000 bootstrap draws resample the 30 prompts while retaining all 5 responses for each sampled prompt. The subject-and-scene consistency win rate is 83.5\% [73.2,92.2], and overall preference is 82.9\% [72.6,92.1], with ties excluded from each denominator. Both 95\% intervals exclude 50\%. Table~\ref{tab:human-evaluation-counts} lists counts for both studies.

\begin{table}[!htbp]
\centering
\begin{minipage}{0.65\linewidth}
\caption{Human-evaluation response counts. Wins favor the indicated method or reference. SGF uses the 9-frame reference, and LongLive uses Commit Forcing. Win rates exclude ties.}
\label{tab:human-evaluation-counts}
\centering
\footnotesize
\setlength{\tabcolsep}{2pt}
\renewcommand{\arraystretch}{1.05}
\textit{SGF: 9-frame reference vs. Base, 60 s}\par\smallskip
\begin{tabular*}{\linewidth}{@{\extracolsep{\fill}}lccc@{}}
\reporttoprule
\textbf{Criterion} & \textbf{Wins} & \textbf{Losses} & \textbf{Ties} \\
\midrule
Subject and scene & 39 & 13 & 2 \\
Overall preference & 35 & 12 & 7 \\
\bottomrule
\end{tabular*}
\par\medskip
\textit{LongLive: Commit Forcing vs. Base, 60 s}\par\smallskip
\begin{tabular*}{\linewidth}{@{\extracolsep{\fill}}lccc@{}}
\reporttoprule
\textbf{Criterion} & \textbf{Wins} & \textbf{Losses} & \textbf{Ties} \\
\midrule
Subject and scene & 19 & 25 & 10 \\
Overall preference & 20 & 24 & 10 \\
\bottomrule
\end{tabular*}
\par\medskip
\textit{LongLive: Commit Forcing vs. Base, 120 s}\par\smallskip
\begin{tabular*}{\linewidth}{@{\extracolsep{\fill}}lccc@{}}
\reporttoprule
\textbf{Criterion} & \textbf{Wins} & \textbf{Losses} & \textbf{Ties} \\
\midrule
Subject and scene & 111 & 22 & 17 \\
Overall preference & 107 & 22 & 21 \\
\bottomrule
\end{tabular*}
\par\medskip
\textit{LongLive: Commit Forcing vs. ID-Forcing, 60 s}\par\smallskip
\begin{tabular*}{\linewidth}{@{\extracolsep{\fill}}lccc@{}}
\reporttoprule
\textbf{Criterion} & \textbf{Wins} & \textbf{Losses} & \textbf{Ties} \\
\midrule
Subject and scene & 47 & 5 & 2 \\
Overall preference & 45 & 6 & 3 \\
\bottomrule
\end{tabular*}
\end{minipage}
\end{table}

\subsection{Long-video results}
\label{app:long-video-components}

The LongLive and SF comparisons at 120 and 240 seconds use ID-Forcing's 128-prompt MovieGen list with generation seed 0. RF-TF videos retain the 21-frame recent-only control in Section~\ref{app:model-configurations}. Without an RF-only arm under this protocol, these results do not estimate the marginal benefit of the committing-context change at either duration.

\paragraph{LongLive and ID-Forcing at long durations.}
\phantomsection
\label{app:long-idf-results}
Commit Forcing gives higher clip-to-clip consistency and lower Dynamic Degree than complete ID-Forcing, with no significant Quality difference in Table~\ref{tab:longlive-six-frame-horizons}. At 240 seconds, Imaging Quality decreases by 0.45 points relative to the 12-frame context.

\begin{table}[!htbp]
\caption{LongLive 6-frame Commit Forcing at 120 and 240 seconds on the 128-prompt MovieGen list, with generation seed 0.}
\label{tab:longlive-six-frame-horizons}
\centering
\footnotesize
\setlength{\tabcolsep}{3pt}
\renewcommand{\arraystretch}{1.1}
\begin{tabular*}{\linewidth}{@{\extracolsep{\fill}}l*{6}{c}@{}}
\reporttoprule
& \multicolumn{3}{c}{\textbf{120 s}} & \multicolumn{3}{c}{\textbf{240 s}} \\
\cmidrule(lr){2-4}\cmidrule(l){5-7}
\textbf{Metric} & \textbf{vs. Base} & \textbf{vs. 12 frames} & \textbf{vs. ID-Forcing} & \textbf{vs. Base} & \textbf{vs. 12 frames} & \textbf{vs. ID-Forcing} \\
\midrule
Quality & \shortstack{+0.33\sigmark\\{}[+0.16, +0.50]} & \shortstack{+0.12\sigmark\\{}[+0.03, +0.21]} & \shortstack{\textminus{}0.22\\{}[\textminus{}0.57, +0.12]} & \shortstack{+0.28\sigmark\\{}[+0.08,+0.49]} & \shortstack{+0.00\\{}[\textminus{}0.10,+0.11]} & \shortstack{\textminus{}0.34\\{}[\textminus{}0.71,+0.02]} \\
Subject$_{\mathrm{c2c}}$ & \shortstack{+1.87\sigmark\\{}[+1.39, +2.36]} & \shortstack{+0.78\sigmark\\{}[+0.43, +1.14]} & \shortstack{+7.31\sigmark\\{}[+5.88, +8.85]} & \shortstack{+2.55\sigmark\\{}[+1.98,+3.15]} & \shortstack{+0.60\sigmark\\{}[+0.32,+0.87]} & \shortstack{+8.21\sigmark\\{}[+6.63,+9.91]} \\
Background$_{\mathrm{c2c}}$ & \shortstack{+1.89\sigmark\\{}[+1.57, +2.22]} & \shortstack{+0.70\sigmark\\{}[+0.47, +0.92]} & \shortstack{+5.49\sigmark\\{}[+4.52, +6.53]} & \shortstack{+2.59\sigmark\\{}[+2.21,+2.98]} & \shortstack{+0.62\sigmark\\{}[+0.42,+0.83]} & \shortstack{+6.51\sigmark\\{}[+5.49,+7.64]} \\
DD & \shortstack{+3.41\sigmark\\{}[+1.18, +5.70]} & \shortstack{\textminus{}0.05\\{}[\textminus{}1.35, +1.19]} & \shortstack{\textminus{}11.00\sigmark\\{}[\textminus{}15.21, \textminus{}6.91]} & \shortstack{+3.82\sigmark\\{}[+1.19,+6.52]} & \shortstack{\textminus{}0.59\\{}[\textminus{}1.99,+0.85]} & \shortstack{\textminus{}12.73\sigmark\\{}[\textminus{}17.32,\textminus{}8.36]} \\
MS & \shortstack{\textminus{}0.13\sigmark\\{}[\textminus{}0.18, \textminus{}0.08]} & \shortstack{+0.02\sigmark\\{}[+0.00, +0.04]} & \shortstack{+0.19\sigmark\\{}[+0.08, +0.32]} & \shortstack{\textminus{}0.16\sigmark\\{}[\textminus{}0.22,\textminus{}0.10]} & \shortstack{+0.02\sigmark\\{}[+0.002,+0.04]} & \shortstack{+0.14\sigmark\\{}[+0.02,+0.27]} \\
TF & \shortstack{\textminus{}0.18\sigmark\\{}[\textminus{}0.28, \textminus{}0.09]} & \shortstack{+0.06\sigmark\\{}[+0.02, +0.10]} & \shortstack{+0.63\sigmark\\{}[+0.40, +0.88]} & \shortstack{\textminus{}0.22\sigmark\\{}[\textminus{}0.33,\textminus{}0.12]} & \shortstack{+0.07\sigmark\\{}[+0.03,+0.11]} & \shortstack{+0.58\sigmark\\{}[+0.34,+0.85]} \\
\bottomrule
\end{tabular*}
\par\smallskip
\parbox{\linewidth}{\footnotesize Changes are 6-frame Commit Forcing minus the indicated comparison. Brackets show paired 95\% confidence intervals. The base, 6-frame, and 12-frame arms share the 15-frame physical cache and base denoising. Complete ID-Forcing uses its full cache-construction framework.}
\end{table}

\paragraph{Self-only committing at long durations.}
\label{app:longlive-self-only}
As shown in Table~\ref{tab:longlive-self-only}, Commit Forcing lowers Dynamic Degree and Quality relative to self-only committing but increases clip-to-clip consistency. Motion Smoothness also increases by +0.15 [+0.11,+0.20] and Temporal Flickering by +0.33 [+0.24,+0.41] at 120 seconds, and by +0.15 [+0.11,+0.20] and +0.34 [+0.24,+0.44] at 240 seconds.

\begin{table}[!htbp]
\caption{LongLive self-only committing at 120 and 240 seconds on the 128-prompt MovieGen list, with generation seed 0. Absolute rows give scores; other rows give paired changes with 95\% intervals.}
\label{tab:longlive-self-only}
\centering
\footnotesize
\setlength{\tabcolsep}{4pt}
\renewcommand{\arraystretch}{1.10}
\begin{tabular*}{\linewidth}{@{\extracolsep{\fill}}clcccc@{}}
\reporttoprule
\textbf{Duration} & \textbf{Comparison} & \textbf{Quality} & \textbf{DD} & \textbf{Subject$_{\mathrm{c2c}}$} & \textbf{Background$_{\mathrm{c2c}}$} \\
\midrule
120 s & Self-only (absolute) & 82.22 & 52.68 & 84.73 & 88.05 \\
120 s & Self-only \textminus{} base & \shortstack{+0.50\sigmark\\{}[+0.27, +0.72]} & \shortstack{+9.73\sigmark\\{}[+6.91, +12.79]} & \shortstack{+0.56\\{}[\textminus{}0.11, +1.23]} & \shortstack{+0.88\sigmark\\{}[+0.49, +1.26]} \\
120 s & 6 frames \textminus{} self-only & \shortstack{\textminus{}0.17\sigmark\\{}[\textminus{}0.31, \textminus{}0.03]} & \shortstack{\textminus{}6.32\sigmark\\{}[\textminus{}8.40, \textminus{}4.32]} & \shortstack{+1.30\sigmark\\{}[+0.89, +1.75]} & \shortstack{+1.01\sigmark\\{}[+0.75, +1.28]} \\
120 s & Self-only \textminus{} complete ID & \shortstack{\textminus{}0.05\\{}[\textminus{}0.37, +0.27]} & \shortstack{\textminus{}4.69\sigmark\\{}[\textminus{}8.37, \textminus{}1.11]} & \shortstack{+6.01\sigmark\\{}[+4.60, +7.42]} & \shortstack{+4.48\sigmark\\{}[+3.59, +5.40]} \\
\addlinespace[4pt]
240 s & Self-only (absolute) & 82.05 & 51.45 & 84.65 & 88.00 \\
240 s & Self-only \textminus{} base & \shortstack{+0.49\sigmark\\{}[+0.24, +0.74]} & \shortstack{+9.89\sigmark\\{}[+6.61, +13.35]} & \shortstack{+1.24\sigmark\\{}[+0.57, +1.91]} & \shortstack{+1.47\sigmark\\{}[+0.98, +1.95]} \\
240 s & 6 frames \textminus{} self-only & \shortstack{\textminus{}0.21\sigmark\\{}[\textminus{}0.34, \textminus{}0.08]} & \shortstack{\textminus{}6.07\sigmark\\{}[\textminus{}8.09, \textminus{}4.21]} & \shortstack{+1.31\sigmark\\{}[+0.99, +1.67]} & \shortstack{+1.12\sigmark\\{}[+0.82, +1.43]} \\
240 s & Self-only \textminus{} complete ID & \shortstack{\textminus{}0.13\\{}[\textminus{}0.46, +0.20]} & \shortstack{\textminus{}6.66\sigmark\\{}[\textminus{}10.74, \textminus{}2.87]} & \shortstack{+6.89\sigmark\\{}[+5.38, +8.37]} & \shortstack{+5.39\sigmark\\{}[+4.47, +6.35]} \\
\bottomrule
\end{tabular*}
\par\smallskip
\parbox{\linewidth}{\footnotesize\raggedright The base, self-only, and 6-frame arms share the 15-frame physical cache and base denoising. Complete ID-Forcing uses its full cache-construction framework.}
\end{table}

We evaluate color and motion drift with ID-Forcing's metrics, where larger values are better~\citep{shin2026idforcing}. Color-drift scores appear in Table~\ref{tab:id-drift-comparison}. On SF at 120 seconds, motion-drift scores are 94.53 with complete ID-Forcing and 88.28 with the 9-frame adaptation within its framework. On LongLive, complete ID-Forcing gives motion-drift scores of 96.88 and 98.44 at 120 and 240 seconds, compared with 93.75 and 90.62 for 6-frame Commit Forcing in the base LongLive configuration.

\begin{table}[!htbp]
\centering
\begin{minipage}{0.80\linewidth}
\centering\footnotesize
\setlength{\tabcolsep}{2pt}
\renewcommand{\arraystretch}{1.10}
\caption{Long-video comparisons on ID-Forcing's 128-prompt MovieGen list. Higher color-drift scores indicate better results under ID-Forcing's metric.}
\label{tab:id-drift-comparison}
\par\medskip
\textit{LongLive}\par\smallskip
\begin{tabular*}{\linewidth}{@{\extracolsep{\fill}}clccc@{}}
\reporttoprule
\textbf{Duration} & \textbf{Configuration} & \textbf{Quality} & \textbf{Subject$_{\mathrm{c2c}}$} & \textbf{Color drift} \\
\midrule
120 s & Base & 81.72 & 84.16 & 60.15 \\
120 s & ID-Forcing & 82.27 & 78.72 & 70.35 \\
120 s & Commit Forcing (6) & 82.05 & 86.03 & 74.45 \\
120 s & 12-frame ablation & 81.93 & 85.25 & 74.71 \\
240 s & Base & 81.56 & 83.41 & 55.92 \\
240 s & ID-Forcing & 82.18 & 77.75 & 70.13 \\
240 s & Commit Forcing (6) & 81.84 & 85.96 & 68.95 \\
240 s & 12-frame ablation & 81.84 & 85.36 & 68.66 \\
\bottomrule
\end{tabular*}
\par\medskip
\textit{SF}\par\smallskip
\begin{tabular*}{\linewidth}{@{\extracolsep{\fill}}clccc@{}}
\reporttoprule
\textbf{Duration} & \textbf{Configuration} & \textbf{Quality} & \textbf{Subject$_{\mathrm{c2c}}$} & \textbf{Color drift} \\
\midrule
120 s & Base & 77.63 & 64.02 & 23.96 \\
120 s & RF-TF (21 frames) & 81.78 & 76.27 & 65.33 \\
120 s & ID-Forcing & 81.62 & 76.47 & 68.29 \\
120 s & IDF adaptation (9) & 81.68 & 79.51 & 73.63 \\
240 s & RF-TF (21 frames) & 81.99 & 76.13 & 65.89 \\
240 s & ID-Forcing & 81.32 & 75.07 & 67.96 \\
240 s & IDF adaptation (9) & 81.65 & 78.65 & 73.36 \\
\bottomrule
\end{tabular*}
\par\smallskip
\parbox{\linewidth}{\footnotesize RF-TF uses its 21-frame recent-only reference; IDF adaptation (9) denotes the 9-frame adaptation within ID-Forcing. The LongLive main method uses 6 frames.}
\end{minipage}
\end{table}

\paragraph{SGF and CF at 240 seconds.}
The 240-second SGF and CF comparisons use the first 32 extended MovieGen prompts, generation seed 0, and 6-frame committing contexts. Each video comprises 120 two-second clips, with results in Table~\ref{tab:committing-sources-240}.

\begin{table}[!htbp]
\centering
\begin{minipage}{0.65\linewidth}
\caption{Committing-source controls at 240 seconds on the first 32 extended MovieGen prompts, with generation seed 0. Panel (a) gives absolute scores; panel (b) gives paired gains of the 6-frame rule over base committing.}
\label{tab:committing-sources-240}
\centering
\footnotesize
\setlength{\tabcolsep}{2pt}
\renewcommand{\arraystretch}{1.05}
\textit{(a) CF absolute scores}\par\smallskip
\begin{tabular*}{\linewidth}{@{\extracolsep{\fill}}lcccc@{}}
\reporttoprule
\textbf{Metric} & \textbf{Base} & \textbf{No slow} & \textbf{No sink} & \textbf{Commit Forcing} \\
\midrule
Quality & 83.60 & 83.41 & 84.10 & 84.04 \\
SC & 98.03 & 98.23 & 97.97 & 98.21 \\
BC & 96.78 & 97.00 & 96.70 & 96.87 \\
DD & 58.46 & 48.83 & 65.52 & 59.64 \\
\bottomrule
\end{tabular*}
\par\medskip
\textit{(a) SGF absolute scores}\par\smallskip
\begin{tabular*}{\linewidth}{@{\extracolsep{\fill}}lcc@{}}
\reporttoprule
\textbf{Metric} & \textbf{Base} & \textbf{Commit Forcing} \\
\midrule
Quality & 83.97 & 84.87 \\
SC & 98.26 & 98.59 \\
BC & 96.75 & 97.17 \\
DD & 49.11 & 60.62 \\
\bottomrule
\end{tabular*}
\par\medskip
\textit{(b) Paired changes from base}\par\smallskip
\begin{tabular*}{\linewidth}{@{\extracolsep{\fill}}lcc@{}}
\reporttoprule
\textbf{Metric} & \textbf{CF} & \textbf{SGF} \\
\midrule
Quality & \shortstack{+0.44\\{}[\textminus{}0.08,+0.97]} & \shortstack{+0.90\\{}[\textminus{}0.002,+1.82]} \\
SC & \shortstack{+0.18\\{}[+0.06,+0.31]} & \shortstack{+0.34\\{}[+0.14,+0.55]} \\
BC & \shortstack{+0.10\\{}[\textminus{}0.01,+0.20]} & \shortstack{+0.42\\{}[+0.22,+0.65]} \\
Subject$_{\mathrm{c2c}}$ & \shortstack{+2.23\\{}[+0.51,+3.89]} & \shortstack{+5.42\\{}[+3.02,+8.09]} \\
\bottomrule
\end{tabular*}
\par\smallskip
\parbox{\linewidth}{\footnotesize No slow and No sink exclude only the indicated historical source during committing. The main SGF and CF rows use the current block and the preceding block's stored KV. SC and BC are fused scores.}
\end{minipage}
\end{table}

For CF sink exclusion during the committing pass, we obtain Quality +0.35 [+0.16,+0.54] at 60 seconds on the first 200 extended MovieGen prompts and +0.50 [+0.06,+0.96] at 240 seconds on the first 32 prompts, using generation seed 0. At 60 seconds under this first-200-prompt protocol, we obtain fused Subject Consistency $-$0.02 [$-$0.08,+0.05] and Background +0.00 [$-$0.05,+0.05]. At 240 seconds, we obtain fused Subject Consistency $-$0.07 [$-$0.21,+0.07] and Background $-$0.07 [$-$0.19,+0.03]. For slow-memory exclusion alone on the first 200 prompts, we obtain 60-second Quality $-$0.15 [$-$0.29,$-$0.01].

Commit Forcing on SGF increases Dynamic Degree by 11.51 points, with a paired interval of [+0.60,+23.44].

\paragraph{Extended MovieGen prompts.}
\label{app:extended-moviegen-results}
These comparisons use the first 200 extended MovieGen prompts at 60 seconds, with RF-TF using SF weights.

\phantomsection
\hypertarget{official-vbench-long-metrics-at-60-seconds}{}\label{official-vbench-long-metrics-at-60-seconds}
Table~\ref{tab:components60} reports the first 200 extended MovieGen prompts at 60 seconds. On RF-TF, higher Quality accompanies higher Imaging Quality, while Dynamic Degree and Motion Smoothness differences are not significant.

\begin{table}[!htbp]
\centering
\begin{minipage}{0.65\linewidth}
\caption{Quality components on the first 200 extended MovieGen prompts at 60 seconds. Subject and Background Consistency use fused scores.}
\label{tab:components60}
\centering
\footnotesize
\setlength{\tabcolsep}{2pt}
\renewcommand{\arraystretch}{1.05}
\textit{RF-TF}\par\smallskip
\begin{tabular*}{\linewidth}{@{\extracolsep{\fill}}lccc@{}}
\reporttoprule
\textbf{Metric} & \textbf{Base} & \textbf{Commit Forcing} & \textbf{$\boldsymbol{\Delta}$ [95\% CI]} \\
\midrule
Quality & 82.29 & 82.86 & +0.57 [+0.35, +0.80] \\
SC & 96.62 & 97.17 & +0.55 [+0.44, +0.67] \\
BC & 95.87 & 96.40 & +0.53 [+0.46, +0.59] \\
TF & 96.44 & 96.49 & +0.05 [\textminus{}0.08, +0.17] \\
MS & 98.18 & 98.11 & \textminus{}0.06 [\textminus{}0.15, +0.01] \\
DD & 62.67 & 62.47 & \textminus{}0.20 [\textminus{}2.57, +2.27] \\
AQ & 58.35 & 60.37 & +2.02 [+1.58, +2.46] \\
IQ & 69.73 & 70.26 & +0.53 [+0.10, +0.98] \\
\bottomrule
\end{tabular*}
\par\medskip
\textit{SGF}\par\smallskip
\begin{tabular*}{\linewidth}{@{\extracolsep{\fill}}lccc@{}}
\reporttoprule
\textbf{Metric} & \textbf{Base} & \textbf{Commit Forcing} & \textbf{$\boldsymbol{\Delta}$ [95\% CI]} \\
\midrule
Quality & 83.98 & 84.70 & +0.72 [+0.33, +1.13] \\
SC & 98.11 & 98.35 & +0.24 [+0.15, +0.33] \\
BC & 96.79 & 97.03 & +0.24 [+0.16, +0.31] \\
TF & 97.44 & 97.01 & \textminus{}0.42 [\textminus{}0.61, \textminus{}0.25] \\
MS & 98.71 & 98.50 & \textminus{}0.21 [\textminus{}0.30, \textminus{}0.13] \\
DD & 56.43 & 65.65 & +9.22 [+4.50, +14.40] \\
AQ & 62.65 & 64.09 & +1.44 [+0.80, +2.08] \\
IQ & 72.04 & 71.93 & \textminus{}0.11 [\textminus{}0.64, +0.44] \\
\bottomrule
\end{tabular*}
\par\medskip
\textit{CF}\par\smallskip
\begin{tabular*}{\linewidth}{@{\extracolsep{\fill}}lccc@{}}
\reporttoprule
\textbf{Metric} & \textbf{Base} & \textbf{Commit Forcing} & \textbf{$\boldsymbol{\Delta}$ [95\% CI]} \\
\midrule
Quality & 82.67 & 82.92 & +0.25 [+0.05, +0.47] \\
SC & 97.43 & 97.64 & +0.21 [+0.13, +0.29] \\
BC & 96.56 & 96.73 & +0.18 [+0.13, +0.23] \\
TF & 97.03 & 96.91 & \textminus{}0.12 [\textminus{}0.26, +0.01] \\
MS & 98.34 & 98.18 & \textminus{}0.15 [\textminus{}0.26, \textminus{}0.06] \\
DD & 56.02 & 58.30 & +2.28 [+0.07, +4.62] \\
AQ & 60.61 & 60.99 & +0.37 [+0.04, +0.74] \\
IQ & 69.24 & 69.76 & +0.52 [+0.13, +0.90] \\
\bottomrule
\end{tabular*}
\par\smallskip
\parbox{\linewidth}{\footnotesize Quality is reconstructed. Metric abbreviations follow Section~\ref{app:benchmark-scoring}.}
\end{minipage}
\end{table}

\paragraph{Qualitative examples.}
\label{app:qualitative-selection}
Figure~\ref{fig:qualitative-grid} uses two manually selected 240-second cases, SGF prompt 19 (the Amalfi Coast church) and CF prompt 10 (a figure raking a Zen garden inside a glass sphere). Both compare base generation with 6-frame Commit Forcing at generation seed 0, using frames at 0, 120, and 238 seconds. The CF example has Quality $-$0.40 with unchanged Dynamic Degree. Its committing pass reads no sink, slow-memory, or stale-slot KV, with base denoising unchanged.

Figure~\ref{fig:history-phase} shows prompt 0 with generation seed 0 at 238 seconds using base generation and 6-frame Commit Forcing. Frames are extracted from the generated videos and resized for display without retouching.

\subsection{Gain over time}
\label{app:gain-over-time}
\begin{figure}[!htbp]
\centering
\begin{minipage}{.80\linewidth}
\centering
\includegraphics[width=\linewidth]{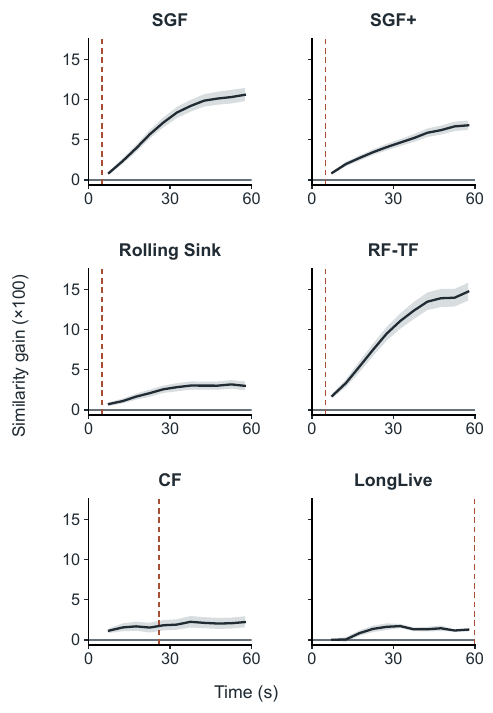}
\caption{Commit Forcing minus base in DINOv2 cosine similarity to the opening 5-second window on the 60-second VBench suite. Each curve averages paired changes over the 946 evaluated entries, with pointwise 95\% bootstrap intervals. Dashed lines mark training horizons.}
\label{fig:gain-over-time}
\end{minipage}
\end{figure}

\begin{figure}[!htbp]
\centering
\begin{minipage}{.80\linewidth}
\centering
\includegraphics[width=\linewidth]{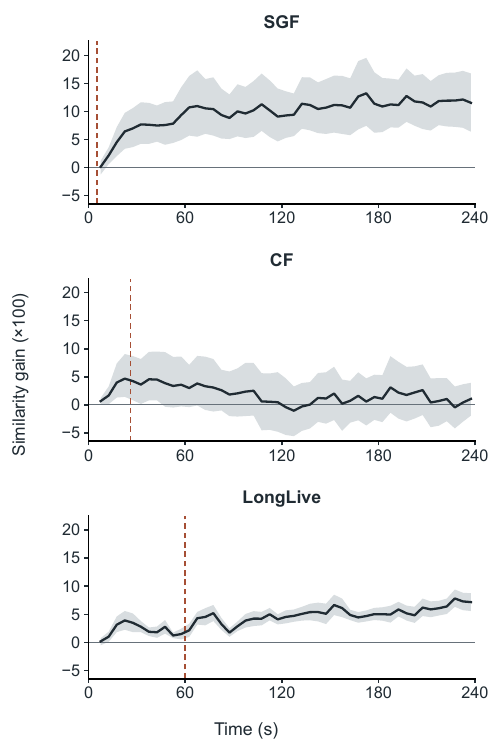}
\caption{Gain from Commit Forcing in similarity to the opening window of 240-second videos. We report the paired difference in DINOv2 feature cosine similarity from the base generator, scaled by 100. Bands give pointwise bootstrap intervals and dashed lines mark training horizons. SGF and CF each use 32 prompts, and LongLive uses 128.}
\label{fig:gain-over-time-240s}
\end{minipage}
\end{figure}

We split videos into 5-second windows and compute the paired change in cosine similarity between each window's mean DINOv2 feature and the opening window, scaled by 100. Frames are sampled at 2 fps. Figures~\ref{fig:gain-over-time} and~\ref{fig:gain-over-time-240s} use the 60-second standard suite and the 240-second videos of Section~\ref{context-forcing-at-240-seconds-and-memory-ablations}. Dashed lines mark the training horizons: about 5 seconds for SGF, SGF+, Rolling Sink, and RF-TF, 26 seconds for CF, and 60 seconds for LongLive. LongLive's gain is larger after its training horizon; the corresponding difference on CF at 240 seconds is not significant.

\subsection{Self-only comparisons}
\label{app:self-only-contrasts}
\begingroup

Table~\ref{tab:native-write-context} reports base, self-only, and wider-window committing controls.

\begin{table}[!htbp]
\centering
\begin{minipage}{0.65\linewidth}
\centering\footnotesize
\setlength{\tabcolsep}{2pt}
\renewcommand{\arraystretch}{1.10}
\caption{Historical committing-context controls on Rolling Forcing's 200 prompts at 60 seconds, with generation seed 0.}
\label{tab:native-write-context}
\par\medskip
\textit{SF + sink}\par\smallskip
\begin{tabular*}{\linewidth}{@{\extracolsep{\fill}}lcccc@{}}
\reporttoprule
\textbf{Committing context} & \textbf{Q} & \textbf{DD} & \textbf{SC} & \textbf{Subject$_{\mathrm{c2c}}$} \\
\midrule
Sink + recent & 81.33 & 57.52 & 96.72 & 75.18 \\
Self-only & 81.28 & 67.80 & 96.56 & 74.69 \\
Recent 21 & 81.63 & 59.58 & 96.87 & 78.56 \\
\bottomrule
\end{tabular*}
\par\medskip
\textit{LongLive}\par\smallskip
\begin{tabular*}{\linewidth}{@{\extracolsep{\fill}}lcccc@{}}
\reporttoprule
\textbf{Committing context} & \textbf{Q} & \textbf{DD} & \textbf{SC} & \textbf{Subject$_{\mathrm{c2c}}$} \\
\midrule
Sink + recent & 82.23 & 42.98 & 97.76 & 86.01 \\
Self-only & 82.73 & 53.03 & 97.69 & 86.21 \\
Recent 12 & 82.65 & 49.38 & 97.70 & 86.93 \\
\bottomrule
\end{tabular*}
\par\medskip
\textit{SGF}\par\smallskip
\begin{tabular*}{\linewidth}{@{\extracolsep{\fill}}lcccc@{}}
\reporttoprule
\textbf{Committing context} & \textbf{Q} & \textbf{DD} & \textbf{SC} & \textbf{Subject$_{\mathrm{c2c}}$} \\
\midrule
Sink + recent (12) & 83.58 & 54.92 & 98.03 & 86.54 \\
Sink + recent (9) & 83.35 & 50.63 & 98.11 & 86.87 \\
Self-only (3) & 84.65 & 71.37 & 98.23 & 91.09 \\
Recent 9 & 84.62 & 68.30 & 98.29 & 90.93 \\
\bottomrule
\end{tabular*}
\par\smallskip
\parbox{\linewidth}{\footnotesize Recent 21, 12, and 9 denote wider-window references, with frame counts including the current block. Q, DD, and SC denote Quality, Dynamic Degree, and fused Subject Consistency. These controls are separate from 6-frame Commit Forcing.}
\end{minipage}
\end{table}

Table~\ref{tab:sgf-committing-controls} also gives the SGF self-only contrast as B$-$C. Relative to self-only committing, reading recent KV trades lower Dynamic Degree for higher Motion Smoothness and Temporal Flickering scores, with no clear Quality or clip-to-clip Subject Consistency gain.

\paragraph{Self-only committing on the standard suite.}
As shown in Table~\ref{tab:longlive-self-only-standard}, self-only committing increases Dynamic Degree without a significant Quality or Total change.

\begin{table}[!htbp]
\centering
\begin{minipage}{0.65\linewidth}
\caption{LongLive self-only versus 6-frame committing on the standard suite.}
\label{tab:longlive-self-only-standard}
\centering
\footnotesize
\setlength{\tabcolsep}{2pt}
\renewcommand{\arraystretch}{1.05}
\begin{tabular*}{\linewidth}{@{\extracolsep{\fill}}lccc@{}}
\reporttoprule
\textbf{Metric} & \textbf{6 frames} & \textbf{Self-only} & \textbf{$\boldsymbol{\Delta}$ [95\% CI]} \\
\midrule
Quality & 84.99 & 85.17 & +0.18  [\textminus{}0.07, +0.45] \\
Semantic & 76.84 & 76.49 & \textminus{}0.35  [\textminus{}0.74, +0.03] \\
Total & 83.36 & 83.43 & +0.07  [\textminus{}0.14, +0.31] \\
SC & 98.00 & 97.84 & \textminus{}0.16\sigmark  [\textminus{}0.26, \textminus{}0.08] \\
BC & 97.23 & 97.05 & \textminus{}0.18\sigmark  [\textminus{}0.25, \textminus{}0.12] \\
TF & 99.28 & 99.21 & \textminus{}0.07\sigmark  [\textminus{}0.12, \textminus{}0.02] \\
MS & 98.55 & 98.46 & \textminus{}0.09\sigmark  [\textminus{}0.14, \textminus{}0.05] \\
DD & 54.81 & 60.14 & +5.32\sigmark  [+1.97, +8.95] \\
AQ & 66.71 & 66.32 & \textminus{}0.39\sigmark  [\textminus{}0.57, \textminus{}0.21] \\
IQ & 70.46 & 70.26 & \textminus{}0.19  [\textminus{}0.44, +0.06] \\
OC & 88.52 & 88.10 & \textminus{}0.42  [\textminus{}1.61, +0.55] \\
MO & 79.91 & 79.76 & \textminus{}0.15  [\textminus{}1.33, +1.23] \\
HA & 95.50 & 95.50 & +0.00  [\textminus{}0.79, +0.80] \\
Col & 92.04 & 91.79 & \textminus{}0.25  [\textminus{}1.54, +0.95] \\
SR & 76.30 & 74.20 & \textminus{}2.10\sigmark  [\textminus{}3.74, \textminus{}0.60] \\
Sc & 45.71 & 45.56 & \textminus{}0.15  [\textminus{}1.93, +1.58] \\
AS & 20.73 & 20.76 & +0.03  [\textminus{}0.04, +0.10] \\
TS & 22.65 & 22.64 & \textminus{}0.02  [\textminus{}0.11, +0.08] \\
OCon & 25.09 & 25.06 & \textminus{}0.04  [\textminus{}0.12, +0.05] \\
\bottomrule
\end{tabular*}
\par\smallskip
\parbox{\linewidth}{\footnotesize Self-only minus 6-frame committing on all 946 entries at 60 seconds, with generation seed 0 and a 12-frame physical cache in both conditions. Intervals use 4,000 paired draws over 944 prompt clusters. Abbreviations follow Section~\ref{app:benchmark-scoring}.}
\end{minipage}
\end{table}

\paragraph{Wider-window references.}
Relative to base SGF, the 9-frame reference at 240 seconds changes Motion Smoothness by $-$0.18 [$-$0.35,$-$0.04], Temporal Flickering by $-$0.42 [$-$0.79,$-$0.09], Imaging Quality by $-$0.40 [$-$1.56,+0.67], and Aesthetic Quality by +0.86 [$-$0.88,+2.69].

The 6- and 12-frame LongLive contexts both increase clip-to-clip consistency and reduce Dynamic Degree relative to self-only committing. The SF-with-sink controls use the wider windows in Table~\ref{tab:native-write-context}.

\paragraph{Complete ID-Forcing comparisons.}
Tables~\ref{tab:standard-idf-main} and~\ref{tab:standard-idf-quality-components} give aggregate and component scores for the 9-frame ID-Forcing adaptations. Neither adaptation changes Quality or Total significantly relative to complete ID-Forcing, while Semantic increases on both. Dynamic Degree decreases by 3.80 [1.02,6.92] on SF and 6.16 [2.05,10.59] on LongLive.

\begin{table}[!htbp]
\centering
\begin{minipage}{0.65\linewidth}
\caption{Complete ID-Forcing and its 9-frame adaptations on the 60-second standard suite.}
\label{tab:standard-idf-main}
\centering
\footnotesize
\setlength{\tabcolsep}{2pt}
\renewcommand{\arraystretch}{1.05}
\textit{IDF-SF}\par\smallskip
\begin{tabular*}{\linewidth}{@{\extracolsep{\fill}}lccc@{}}
\reporttoprule
\textbf{Score} & \textbf{Complete ID} & \textbf{9-frame adaptation} & \textbf{$\boldsymbol{\Delta}$ [95\% CI]} \\
\midrule
Quality & 85.26 & 85.21 & \shortstack{\textminus{}0.05\\{}[\textminus{}0.30,\,+0.19]} \\
Semantic & 73.76 & 74.90 & \shortstack{+1.14\\{}[+0.39,\,+1.91]} \\
Total & 82.96 & 83.15 & \shortstack{+0.19\\{}[\textminus{}0.07,\,+0.43]} \\
\bottomrule
\end{tabular*}
\par\medskip
\textit{IDF-LL}\par\smallskip
\begin{tabular*}{\linewidth}{@{\extracolsep{\fill}}lccc@{}}
\reporttoprule
\textbf{Score} & \textbf{Complete ID} & \textbf{9-frame adaptation} & \textbf{$\boldsymbol{\Delta}$ [95\% CI]} \\
\midrule
Quality & 84.91 & 84.64 & \shortstack{\textminus{}0.27\\{}[\textminus{}0.61,\,+0.05]} \\
Semantic & 74.93 & 75.98 & \shortstack{+1.05\\{}[+0.35,\,+1.75]} \\
Total & 82.92 & 82.91 & \shortstack{\textminus{}0.01\\{}[\textminus{}0.32,\,+0.29]} \\
\bottomrule
\end{tabular*}
\par\smallskip
\parbox{\linewidth}{\footnotesize The adaptations read the current 3-frame block and stored KV from two preceding blocks. On SF they also change cache handover and omit tail recaching. Intervals use 944 prompt clusters.}
\end{minipage}
\end{table}

\begin{table}[!htbp]
\centering
\begin{minipage}{0.80\linewidth}
\caption{Component scores for complete ID-Forcing and its 9-frame adaptations on the 60-second standard suite (0--100 scale).}
\label{tab:standard-idf-quality-components}
\label{tab:standard-idf-semantic-components}
\centering
\footnotesize
\setlength{\tabcolsep}{3pt}
\renewcommand{\arraystretch}{1.08}
\begin{tabular*}{\linewidth}{@{\extracolsep{\fill}}lcccc@{}}
\reporttoprule
& \multicolumn{2}{c}{\textbf{IDF-LL}} & \multicolumn{2}{c}{\textbf{IDF-SF}} \\
\cmidrule(lr){2-3}\cmidrule(lr){4-5}
\textbf{Metric} & \textbf{Complete ID} & \textbf{Adaptation} & \textbf{Complete ID} & \textbf{Adaptation} \\
\midrule
\multicolumn{5}{@{}l}{\textit{Quality components}} \\
SC & 97.80 & 97.90 & 97.32 & 97.65 \\
BC & 96.89 & 97.05 & 96.63 & 96.77 \\
TF & 99.21 & 99.29 & 98.91 & 99.00 \\
MS & 98.51 & 98.64 & 98.32 & 98.37 \\
DD & 56.85 & 50.69 & 68.01 & 64.21 \\
AQ & 66.08 & 66.26 & 64.23 & 64.70 \\
IQ & 70.58 & 70.72 & 71.48 & 71.65 \\
\addlinespace[3pt]
\multicolumn{5}{@{}l}{\textit{Semantic components}} \\
OC & 86.00 & 87.30 & 87.73 & 88.66 \\
MO & 75.55 & 76.99 & 67.96 & 73.12 \\
HA & 94.17 & 94.50 & 92.83 & 92.33 \\
Col & 91.45 & 92.37 & 87.30 & 89.66 \\
SR & 71.69 & 74.47 & 74.71 & 76.41 \\
Sc & 43.05 & 45.54 & 42.26 & 42.78 \\
AS & 20.81 & 20.67 & 20.55 & 20.40 \\
TS & 22.44 & 22.43 & 22.26 & 22.41 \\
OCon & 25.00 & 25.05 & 25.06 & 25.10 \\
\bottomrule
\end{tabular*}
\par\smallskip
\parbox{\linewidth}{\footnotesize\raggedright Abbreviations are defined in Section~\ref{app:benchmark-scoring}.}
\end{minipage}
\end{table}

\begin{table}[!htbp]
\centering
\begin{minipage}{0.65\linewidth}
\centering\footnotesize
\setlength{\tabcolsep}{2pt}
\renewcommand{\arraystretch}{1.10}
\caption{Complete ID-Forcing and its 9-frame adaptations on Rolling Forcing's 200 prompts at 60 seconds, with generation seed 0.}
\label{tab:idf-framework}
\par\medskip
\textit{SF}\par\smallskip
\begin{tabular*}{\linewidth}{@{\extracolsep{\fill}}lcccc@{}}
\reporttoprule
\textbf{Configuration} & \textbf{Q} & \textbf{DD} & \textbf{SC} & \textbf{Subject$_{\mathrm{c2c}}$} \\
\midrule
Base & 80.22 & 34.58 & 96.99 & 74.22 \\
Complete ID & 82.20 & 61.95 & 97.03 & 79.07 \\
9-frame adaptation & 82.44 & 60.78 & 97.22 & 81.60 \\
\bottomrule
\end{tabular*}
\par\medskip
\textit{LongLive}\par\smallskip
\begin{tabular*}{\linewidth}{@{\extracolsep{\fill}}lcccc@{}}
\reporttoprule
\textbf{Configuration} & \textbf{Q} & \textbf{DD} & \textbf{SC} & \textbf{Subject$_{\mathrm{c2c}}$} \\
\midrule
Base & 82.23 & 42.98 & 97.76 & 86.01 \\
Complete ID & 82.54 & 52.52 & 97.69 & 82.83 \\
9-frame adaptation & 82.45 & 47.05 & 97.84 & 84.88 \\
\bottomrule
\end{tabular*}
\par\smallskip
\parbox{\linewidth}{\footnotesize On SF the adaptation also changes cache handover and omits tail recaching. Metric abbreviations follow Section~\ref{app:benchmark-scoring}.}
\end{minipage}
\end{table}

On Rolling Forcing's prompts, complete ID-Forcing improves Quality over base SF by +1.98 [+1.48,+2.50] and over base LongLive by +0.31 [+0.06,+0.57]. The 9-frame SF adaptation adds Quality +0.23 [+0.07,+0.40] and clip-to-clip Subject Consistency +2.53. The seed-1 SF adaptation gives Quality +0.20 [+0.04,+0.37] and consistency +2.27. Relative to the 21-frame SF-with-sink reference, the seed-0 adaptation has Quality +0.80 [+0.29,+1.42].

The LongLive adaptation changes Quality by $-$0.09 [$-$0.25,+0.07], fused Subject Consistency by +0.15, clip-to-clip Subject Consistency by +2.05, and Dynamic Degree by $-$5.47 relative to complete ID-Forcing. Complete ID-Forcing scores 3.18 points below base LongLive in clip-to-clip Subject Consistency. Six-frame Commit Forcing in base LongLive exceeds complete ID-Forcing by 4.38 and 4.83 points for seeds 0 and 1, with both intervals excluding 0. The 12-frame LongLive reference has Quality +0.20 [+0.03,+0.39] relative to the 9-frame adaptation within ID-Forcing.
\par
\endgroup

\subsection{Committing-window results}
\label{app:committing-tables}

Table~\ref{tab:standard-window-ablation} compares 6-frame Commit Forcing with wider-window references on the standard suite. Quality differences are not significant in the four SGF, SGF+, RF-TF, and Rolling Sink contrasts, and RF-TF has a significant Total increase. LongLive's 6- and 12-frame contexts do not differ significantly in Quality or Total.

\begin{table}[!htbp]
\centering
\begin{minipage}{0.80\linewidth}
\caption{Six-frame Commit Forcing versus wider committing windows on the 60-second standard suite.}
\label{tab:standard-window-ablation}
\label{tab:longlive-standard-window}
\centering
\footnotesize
\setlength{\tabcolsep}{4pt}
\renewcommand{\arraystretch}{1.10}
\begin{tabular*}{\linewidth}{@{\extracolsep{\fill}}lccc@{}}
\reporttoprule
\textbf{Comparison} & \textbf{$\boldsymbol{\Delta}$ Quality [95\% CI]} & \textbf{$\boldsymbol{\Delta}$ Semantic [95\% CI]} & \textbf{$\boldsymbol{\Delta}$ Total [95\% CI]} \\
\midrule
SGF, 6 \textminus{} 9 frames & \shortstack{+0.11\\{}[\textminus{}0.25,\,+0.46]} & \shortstack{\textminus{}1.01\\{}[\textminus{}2.55,\,+0.50]} & \shortstack{\textminus{}0.11\\{}[\textminus{}0.53,\,+0.29]} \\
RF-TF, 6 \textminus{} 21 frames & \shortstack{+0.18\\{}[\textminus{}0.07,\,+0.41]} & \shortstack{+0.56\\{}[\textminus{}0.24,\,+1.36]} & \shortstack{+0.26\\{}[+0.00007,\,+0.51]} \\
SGF+, 6 \textminus{} 9 frames & \shortstack{+0.03\\{}[\textminus{}0.08,\,+0.14]} & \shortstack{+0.11\\{}[\textminus{}0.11,\,+0.33]} & \shortstack{+0.04\\{}[\textminus{}0.05,\,+0.14]} \\
Rolling Sink, 6 \textminus{} 21 frames & \shortstack{+0.09\\{}[\textminus{}0.05,\,+0.25]} & \shortstack{+0.04\\{}[\textminus{}0.42,\,+0.50]} & \shortstack{+0.08\\{}[\textminus{}0.06,\,+0.24]} \\
LongLive, 12 frames \textminus{} Base & \shortstack{+0.21\\{}[\textminus{}0.01,\,+0.44]} & \shortstack{+0.15\\{}[\textminus{}0.25,\,+0.56]} & \shortstack{+0.20\\{}[+0.01,\,+0.41]} \\
LongLive, 6 \textminus{} 12 frames & \shortstack{+0.08\\{}[\textminus{}0.06,\,+0.21]} & \shortstack{+0.25\\{}[\textminus{}0.14,\,+0.67]} & \shortstack{+0.11\\{}[\textminus{}0.02,\,+0.25]} \\
\bottomrule
\end{tabular*}
\par\smallskip
\parbox{\linewidth}{\footnotesize\raggedright Frame counts include the current 3-frame block. Denoising and physical-cache settings are held fixed within each comparison. The LongLive 6-frame versus Base gains appear in Table~\ref{tab:standard-main-intervals}.}
\end{minipage}
\end{table}

Table~\ref{tab:longlive-window} gives the 200-prompt window comparison on LongLive using Rolling Forcing's prompts.
\begin{table}[!htbp]
\caption{LongLive committing windows on Rolling Forcing's 200 prompts at 60 seconds for each generation seed. Changes are relative to self-only committing. Figure~\ref{fig:committing-window-mechanisms} uses seed 0, where $w=3$, 6, 9, and 12 read 0--3 preceding blocks.}
\label{tab:longlive-window}
\centering
\footnotesize
\setlength{\tabcolsep}{4pt}
\renewcommand{\arraystretch}{1.10}
\begin{tabular*}{\linewidth}{@{\extracolsep{\fill}}cccccccc@{}}
\reporttoprule
& & \multicolumn{2}{c}{\textbf{Quality}} & \multicolumn{2}{c}{\textbf{DD}} & \multicolumn{2}{c}{\textbf{Subject$_{\mathrm{c2c}}$}} \\
\cmidrule(lr){3-4}\cmidrule(lr){5-6}\cmidrule(lr){7-8}
\textbf{Seed} & $\boldsymbol{w}$ & \textbf{Mean} & \textbf{$\boldsymbol{\Delta}$ [95\% CI]} & \textbf{Mean} & \textbf{$\boldsymbol{\Delta}$ [95\% CI]} & \textbf{Mean} & \textbf{$\boldsymbol{\Delta}$ [95\% CI]} \\
\midrule
0 & 3 & 82.73 & --- & 53.03 & --- & 86.21 & --- \\
0 & 6 & 82.73 & \shortstack{+0.00\\{}[\textminus{}0.12,+0.12]} & 49.40 & \shortstack{\textminus{}3.63\\{}[\textminus{}5.23,\textminus{}2.03]} & 87.21 & \shortstack{+1.00\\{}[+0.68,+1.34]} \\
0 & 9 & 82.72 & \shortstack{\textminus{}0.01\\{}[\textminus{}0.16,+0.13]} & 49.47 & \shortstack{\textminus{}3.57\\{}[\textminus{}5.62,\textminus{}1.57]} & 87.23 & \shortstack{+1.02\\{}[+0.69,+1.35]} \\
0 & 12 & 82.65 & \shortstack{\textminus{}0.08\\{}[\textminus{}0.22,+0.06]} & 49.38 & \shortstack{\textminus{}3.65\\{}[\textminus{}5.53,\textminus{}1.88]} & 86.93 & \shortstack{+0.72\\{}[+0.34,+1.12]} \\
\addlinespace[3pt]
1 & 3 & 82.83 & --- & 54.18 & --- & 85.91 & --- \\
1 & 6 & 82.66 & \shortstack{\textminus{}0.17\\{}[\textminus{}0.30,\textminus{}0.04]} & 48.65 & \shortstack{\textminus{}5.53\\{}[\textminus{}7.37,\textminus{}3.80]} & 87.09 & \shortstack{+1.18\\{}[+0.86,+1.50]} \\
1 & 12 & 82.67 & \shortstack{\textminus{}0.15\\{}[\textminus{}0.30,\textminus{}0.01]} & 50.20 & \shortstack{\textminus{}3.98\\{}[\textminus{}6.00,\textminus{}2.08]} & 86.70 & \shortstack{+0.79\\{}[+0.38,+1.21]} \\
\bottomrule
\end{tabular*}
\par\smallskip
\parbox{\linewidth}{\footnotesize\raggedright The base reference reads 3 sink frames and the latest 9 trajectory frames, including the current 3. The 6-frame context is the main method, 12 frames is a reference, and 9 frames is evaluated only for seed 0. Denoising and sink storage are unchanged; tested windows do not read the sink directly.}
\end{table}

\paragraph{Committing windows across generators.}
Table~\ref{tab:committing-window-sweep} reports the 9 contrasts relative to each generator's reference committing window. None has a significant Quality difference. These are distinct from self-only versus 6-frame comparisons, where RF-TF has significantly lower Quality with self-only committing. Self-only committing generally lowers Motion Smoothness and Temporal Flickering scores.

\begin{table}[!htbp]
\caption{Window changes from each generator's reference committing window on the first 100 Rolling Forcing prompts at 60 seconds, with generation seed 0.}
\label{tab:committing-window-sweep}
\centering
\footnotesize
\setlength{\tabcolsep}{3pt}
\renewcommand{\arraystretch}{1.10}
\begin{tabular*}{\linewidth}{@{\extracolsep{\fill}}lcccccccc@{}}
\reporttoprule
\textbf{Generator} & \textbf{Ref. $w$} & \textbf{New $w$} & \textbf{$\boldsymbol{\Delta}$ Quality} & \textbf{$\boldsymbol{\Delta}$ DD} & \textbf{$\boldsymbol{\Delta}$ Subject$_{\mathrm{c2c}}$} & \textbf{$\boldsymbol{\Delta}$ Background$_{\mathrm{c2c}}$} & \textbf{$\boldsymbol{\Delta}$ MS} & \textbf{$\boldsymbol{\Delta}$ TF} \\
\midrule
SGF & 9 & 6 & \textminus{}0.13 & \textminus{}2.20 & +0.20 & +0.08 & \textminus{}0.02 & \textminus{}0.03 \\
SGF & 9 & 3 & \textminus{}0.03 & +1.87 & \textminus{}0.29 & \textminus{}0.18 & \textminus{}0.11\sigmark & \textminus{}0.14 \\
\addlinespace[2pt]
SGF+ & 9 & 6 & \textminus{}0.10 & \textminus{}0.70 & +0.40 & +0.13 & \textminus{}0.00 & +0.03 \\
SGF+ & 9 & 3 & \textminus{}0.02 & +2.53 & +0.82 & +0.02 & \textminus{}0.16\sigmark & \textminus{}0.22\sigmark \\
\addlinespace[2pt]
RF-TF & 21 & 6 & +0.16 & \textminus{}2.70\sigmark & +2.27\sigmark & +2.17\sigmark & \textminus{}0.00 & +0.05 \\
RF-TF & 21 & 3 & \textminus{}0.08 & +1.50 & \textminus{}0.24 & +0.55 & \textminus{}0.15\sigmark & \textminus{}0.29\sigmark \\
\addlinespace[2pt]
Rolling Sink & 21 & 6 & \textminus{}0.12 & \textminus{}2.57 & +1.85\sigmark & +1.67\sigmark & \textminus{}0.08 & \textminus{}0.02 \\
Rolling Sink & 21 & 3 & +0.03 & +3.43 & +1.26\sigmark & +1.64\sigmark & \textminus{}0.31\sigmark & \textminus{}0.36\sigmark \\
\addlinespace[2pt]
CF & 6 & 3 & +0.03 & +1.03 & \textminus{}0.06 & +0.13 & \textminus{}0.04 & \textminus{}0.02 \\
\bottomrule
\end{tabular*}
\par\smallskip
\parbox{\linewidth}{\footnotesize\raggedright Changes are new context minus the reference window. Counts include the current 3-frame block. CF uses its fast 6-frame reference. No context reads the sink directly during committing. The RF-TF and Rolling Sink contexts contain recent frames from the generated trajectory. Subject$_{\mathrm{c2c}}$ and Background$_{\mathrm{c2c}}$ denote clip-to-clip Subject and Background Consistency, and DD denotes Dynamic Degree. MS denotes Motion Smoothness, and TF denotes Temporal Flickering. $^{*}$A 95\% interval from 4,000 paired prompt-bootstrap draws excludes 0. Direct contrasts against 6 frames appear in Table~\ref{tab:committing-window-plateaus}.}
\end{table}

\paragraph{Denser window scans.}
On the first 100 Rolling Forcing prompts, RF-TF and Rolling Sink reach their highest mean clip-to-clip Subject Consistency at 6 frames. The 6--12-frame RF-TF windows and 6--9-frame Rolling Sink windows have no significant pairwise difference from 6 frames, while both 21-frame windows score lower, as shown in Table~\ref{tab:committing-window-plateaus}. These scans use shared generation noise and retain base denoising. On SGF, SGF+, and CF, the tested windows within the released recent-cache capacity show no significant consistency improvement over self-only committing.

\begin{table}[!htbp]
\centering
\begin{minipage}{0.80\linewidth}
\caption{Paired changes from the 6-frame committing window on the first 100 Rolling Forcing prompts at 60 seconds, with generation seed 0. Means appear in Table~\ref{tab:committing-window-comparison}.}
\label{tab:committing-window-plateaus}
\centering
\footnotesize
\setlength{\tabcolsep}{4pt}
\renewcommand{\arraystretch}{1.10}
\begin{tabular*}{\linewidth}{@{\extracolsep{\fill}}cccc@{}}
\reporttoprule
$\boldsymbol{w}$ & \textbf{$\boldsymbol{\Delta}$ Quality \enspace [95\% CI]} & \textbf{$\boldsymbol{\Delta}$ DD \enspace [95\% CI]} & \textbf{$\boldsymbol{\Delta}$ Subject$_{\mathrm{c2c}}$ \enspace [95\% CI]} \\
\midrule
\multicolumn{4}{@{}l}{\textit{RF-TF}} \\
3 & \textminus{}0.24\sigmark\enspace \enspace [\textminus{}0.44,\textminus{}0.03] & +4.20\sigmark\enspace \enspace [+1.93,+6.63] & \textminus{}2.51\sigmark\enspace \enspace [\textminus{}3.34,\textminus{}1.74] \\
\textbf{6} & --- & --- & --- \\
9 & \textminus{}0.08\enspace \enspace [\textminus{}0.19,+0.04] & +0.17\enspace \enspace [\textminus{}1.13,+1.53] & \textminus{}0.29\enspace \enspace [\textminus{}0.69,+0.10] \\
12 & \textminus{}0.04\enspace \enspace [\textminus{}0.18,+0.10] & +0.83\enspace \enspace [\textminus{}0.73,+2.60] & \textminus{}0.57\enspace \enspace [\textminus{}1.19,+0.03] \\
21 & \textminus{}0.16\enspace \enspace [\textminus{}0.33,+0.02] & +2.70\sigmark\enspace \enspace [+0.70,+5.03] & \textminus{}2.27\sigmark\enspace \enspace [\textminus{}2.98,\textminus{}1.58] \\
\addlinespace[3pt]
\multicolumn{4}{@{}l}{\textit{Rolling Sink}} \\
3 & +0.15\enspace \enspace [\textminus{}0.10,+0.38] & +6.00\sigmark\enspace \enspace [+2.97,+9.03] & \textminus{}0.59\sigmark\enspace \enspace [\textminus{}1.17,\textminus{}0.0025] \\
\textbf{6} & --- & --- & --- \\
9 & +0.03\enspace \enspace [\textminus{}0.13,+0.17] & \textminus{}0.33\enspace \enspace [\textminus{}2.33,+1.47] & \textminus{}0.28\enspace \enspace [\textminus{}0.65,+0.05] \\
12 & +0.03\enspace \enspace [\textminus{}0.17,+0.23] & +0.47\enspace \enspace [\textminus{}1.93,+3.07] & \textminus{}0.70\sigmark\enspace \enspace [\textminus{}1.28,\textminus{}0.22] \\
21 & +0.12\enspace \enspace [\textminus{}0.15,+0.40] & +2.57\enspace \enspace [\textminus{}0.77,+6.17] & \textminus{}1.85\sigmark\enspace \enspace [\textminus{}2.53,\textminus{}1.23] \\
\bottomrule
\end{tabular*}
\end{minipage}
\end{table}

\paragraph{SF with a sink.}
As shown in Table~\ref{tab:sf-sink-committing-windows}, this configuration has a 9--12-frame consistency plateau, and 21 frames scores lower than 9 in the pooled comparison. The 6-frame configuration still improves consistency over base committing with both seeds, as shown in Table~\ref{tab:rollf-main}.

\begin{table}[!htb]
\caption{Committing windows on SF with a sink on Rolling Forcing's 200 prompts at 60 seconds. Seed-specific changes use the 6-frame context of the same seed as the reference.}
\label{tab:sf-sink-committing-windows}
\label{tab:sf-sink-committing-pooled}
\centering
\footnotesize
\setlength{\tabcolsep}{4pt}
\renewcommand{\arraystretch}{1.10}
\textit{Seed-specific means and changes}\par\smallskip
\begin{tabular*}{\linewidth}{@{\extracolsep{\fill}}ccccccc@{}}
\reporttoprule
\textbf{Seed} & $\boldsymbol{w}$ & \textbf{Quality} & \textbf{DD} & \textbf{Subject$_{\mathrm{c2c}}$} & \textbf{$\boldsymbol{\Delta}$ Quality [95\% CI]} & \textbf{$\boldsymbol{\Delta}$ Subject$_{\mathrm{c2c}}$ [95\% CI]} \\
\midrule
0 & $w=3$ & 81.28 & 67.80 & 74.69 & \shortstack{\textminus{}0.36\\{}[\textminus{}1.00,+0.23]} & \shortstack{\textminus{}4.93\sigmark\\{}[\textminus{}5.82,\textminus{}4.07]} \\
0 & $\boldsymbol{w=6}$ & \textbf{81.64} & \textbf{56.58} & \textbf{79.62} & --- & --- \\
0 & $w=9$ & 81.54 & 56.68 & 80.09 & \shortstack{\textminus{}0.09\\{}[\textminus{}0.56,+0.20]} & \shortstack{+0.47\sigmark\\{}[+0.03,+0.90]} \\
0 & $w=12$ & 81.77 & 58.08 & 79.97 & \shortstack{+0.14\\{}[\textminus{}0.40,+0.57]} & \shortstack{+0.35\\{}[\textminus{}0.16,+0.86]} \\
0 & $w=21$ & 81.63 & 59.58 & 78.56 & \shortstack{\textminus{}0.00\\{}[\textminus{}0.57,+0.42]} & \shortstack{\textminus{}1.06\sigmark\\{}[\textminus{}1.77,\textminus{}0.31]} \\
\addlinespace[3pt]
1 & $\boldsymbol{w=6}$ & \textbf{81.57} & \textbf{58.90} & \textbf{78.87} & --- & --- \\
1 & $w=9$ & 81.80 & 60.58 & 79.42 & \shortstack{+0.22\sigmark\\{}[+0.07,+0.37]} & \shortstack{+0.55\sigmark\\{}[+0.04,+1.02]} \\
1 & $w=12$ & 81.69 & 61.02 & 79.29 & \shortstack{+0.11\\{}[\textminus{}0.16,+0.36]} & \shortstack{+0.42\\{}[\textminus{}0.12,+0.94]} \\
1 & $w=21$ & 81.54 & 58.87 & 78.11 & \shortstack{\textminus{}0.03\\{}[\textminus{}0.48,+0.36]} & \shortstack{\textminus{}0.76\sigmark\\{}[\textminus{}1.48,\textminus{}0.05]} \\
\bottomrule
\end{tabular*}
\par\medskip
\textit{Paired contrasts pooled across seeds 0 and 1}\par\smallskip
\begin{tabular*}{\linewidth}{@{\extracolsep{\fill}}lccc@{}}
\reporttoprule
\textbf{Contrast} & \textbf{$\boldsymbol{\Delta}$ Quality [95\% CI]} & \textbf{$\boldsymbol{\Delta}$ DD [95\% CI]} & \textbf{$\boldsymbol{\Delta}$ Subject$_{\mathrm{c2c}}$ [95\% CI]} \\
\midrule
9 minus 6 & \shortstack{+0.06\\{}[\textminus{}0.16,+0.23]} & \shortstack{+0.89\\{}[\textminus{}0.45,+2.29]} & \shortstack{+0.51\sigmark\\{}[+0.18,+0.84]} \\
12 minus 6 & \shortstack{+0.13\\{}[\textminus{}0.14,+0.36]} & \shortstack{+1.81\sigmark\\{}[+0.25,+3.37]} & \shortstack{+0.39\\{}[\textminus{}0.0005,+0.78]} \\
12 minus 9 & \shortstack{+0.06\\{}[\textminus{}0.11,+0.24]} & \shortstack{+0.92\\{}[\textminus{}0.47,+2.33]} & \shortstack{\textminus{}0.12\\{}[\textminus{}0.44,+0.21]} \\
21 minus 9 & \shortstack{\textminus{}0.08\\{}[\textminus{}0.36,+0.17]} & \shortstack{+0.59\\{}[\textminus{}1.38,+2.58]} & \shortstack{\textminus{}1.42\sigmark\\{}[\textminus{}1.86,\textminus{}0.96]} \\
21 minus 6 & \shortstack{\textminus{}0.02\\{}[\textminus{}0.31,+0.23]} & \shortstack{+1.48\\{}[\textminus{}0.59,+3.52]} & \shortstack{\textminus{}0.91\sigmark\\{}[\textminus{}1.47,\textminus{}0.33]} \\
\bottomrule
\end{tabular*}
\par\smallskip
\parbox{\linewidth}{\footnotesize\raggedright The model uses SF EMA weights, a 24-frame cache with a 3-frame sink, and no Temporal Response Bias. Tested committing windows exclude direct sink reads and include the current 3-frame block. For the pooled contrasts, differences are averaged across the two seeds within each prompt before bootstrapping.}
\end{table}

\subsection{Transfer across cache constructions}
\label{app:transfer-boundaries}

\paragraph{Released Rolling Forcing.}
We evaluate the released Rolling Forcing model~\citep{liu2025rollingforcing} on Rolling Forcing prompts 0--99 at 60 seconds with generation seed 0. Its 21-frame committing context contains the current 3-frame latent block and KV from the preceding trajectory frames, excluding the initial global block. We either replace KV from 3 recent frames with KV from the 3 global frames at the same context size or shorten the context to the current block and the preceding block's stored KV (6 frames). Changes apply after the cache is full. Denoising, sampler, cache geometry, positional encoding, and generation noise are retained. This model uses Rolling Forcing's released weights, whereas RF-TF uses Self-Forcing weights with a training-free denoising bias, as specified in Section~\ref{app:rf-configurations}.

Adding global context reduces Quality and clip-to-clip consistency. As shown in Table~\ref{tab:released-rolling-forcing-context}, the 6-frame context also lowers clip-to-clip Subject and Background Consistency in this released model, while its Quality interval includes 0.

\begin{table}[!htb]
\centering
\begin{minipage}{0.65\linewidth}
\caption{Released Rolling Forcing committing contexts relative to its 21-frame recent-only update.}
\label{tab:released-rolling-forcing-context}
\centering
\footnotesize
\setlength{\tabcolsep}{3pt}
\renewcommand{\arraystretch}{1.08}
\begin{tabular*}{\linewidth}{@{\extracolsep{\fill}}lcc@{}}
\reporttoprule
\textbf{Metric} & \shortstack{Global 3 +\\recent 18} & \textbf{Recent 6} \\
\midrule
Quality & \shortstack{\textminus{}0.18\sigmark\\{}[\textminus{}0.31,\textminus{}0.03]} & \shortstack{\textminus{}0.09\\{}[\textminus{}0.25,+0.07]} \\
DD & \textminus{}0.20 & \textminus{}0.67 \\
Subject$_{\mathrm{c2c}}$ & \textminus{}0.61\sigmark & \textminus{}0.65\sigmark \\
Background$_{\mathrm{c2c}}$ & \textminus{}0.57\sigmark & \textminus{}0.31\sigmark \\
\bottomrule
\end{tabular*}
\par\smallskip
\parbox{\linewidth}{\footnotesize Metric abbreviations follow Section~\ref{app:benchmark-scoring}.}
\end{minipage}
\end{table}

\paragraph{WorldPlay context re-encoding.}
WorldPlay~\citep{sun2025worldplay} uses a HunyuanVideo-1.5 action model with 4 denoising steps at 480p. Before each 4-latent-frame block, it jointly re-encodes the latest 12 latent frames and up to 8 older frames in a noncausal context pass. The older historical context contains the initial 4-frame block and a 4-frame clip retrieved by camera-view overlap. A persistent committing chain is absent. Our intervention restricts recent-frame queries in the context pass to text and recent-frame keys, while older-frame queries retain all keys. The constructed KV for all context frames is still read during denoising.

We use the 10 released test scenes with generation seeds 1 and 2. The three camera trajectories are two full yaw rotations over 40 seconds, a 20-second forward-and-back walk, and a 40-second forward walk. Arms are paired by scene, trajectory, and seed, with 4,000 bootstrap draws over scenes. Final-return similarity is DINOv2 similarity to the opening at the last return pose and uses the rotation and forward-and-back trajectories. Color drift sums the absolute deviations of the late-to-early saturation and brightness ratios from 1, using the last and first 5 seconds. The seam ratio measures frame changes at block boundaries, and the 1-second similarity compares DINOv2 embeddings of frames 1 second apart. A separate 40-frame context-budget test uses the same 10 scenes on the 40-second rotation trajectory with seed 1, comparing restricted and unrestricted context construction at that budget.

As shown in Table~\ref{tab:worldplay-context-transfer}, at the default 20-frame budget, the return-similarity and color-drift intervals include 0, while the seam ratio increases and 1-second similarity decreases significantly. At 40 frames, re-encoding recent frames without reading older historical context significantly lowers final-return similarity. In this joint context-re-encoding configuration, excluding older historical context during recent-KV construction does not reproduce the gains observed in the persistent-cache generators.

\smallskip
\noindent\makebox[\linewidth][c]{%
\begin{minipage}{0.65\linewidth}
\captionof{table}{WorldPlay restricted minus unrestricted context re-encoding.}
\label{tab:worldplay-context-transfer}
\centering
\footnotesize
\setlength{\tabcolsep}{3pt}
\renewcommand{\arraystretch}{1.08}
\par\medskip
\textit{(a) 20-frame context}\par\smallskip
\begin{tabular*}{\linewidth}{@{\extracolsep{\fill}}lc@{}}
\reporttoprule
\textbf{Metric} & $\boldsymbol{\Delta}$ [95\% CI] \\
\midrule
Final-return similarity $\uparrow$ & \shortstack{\textminus{}0.0317\\{}[\textminus{}0.1141,+0.0545]} \\
Color drift $\downarrow$ & \shortstack{\textminus{}0.0142\\{}[\textminus{}0.0676,+0.0353]} \\
Seam ratio $\downarrow$ & \shortstack{+0.2067\sigmark\\{}[+0.0338,+0.3842]} \\
1-second similarity $\uparrow$ & \shortstack{\textminus{}0.0089\sigmark\\{}[\textminus{}0.0167,\textminus{}0.0025]} \\
\bottomrule
\end{tabular*}
\par\medskip
\textit{(b) 40-frame context}\par\smallskip
\begin{tabular*}{\linewidth}{@{\extracolsep{\fill}}lc@{}}
\reporttoprule
\textbf{Metric} & $\boldsymbol{\Delta}$ [95\% CI] \\
\midrule
Final-return similarity $\uparrow$ & \shortstack{\textminus{}0.1769\sigmark\\{}[\textminus{}0.2926,\textminus{}0.0636]} \\
Color drift $\downarrow$ & \shortstack{\textminus{}0.0885\\{}[\textminus{}0.2350,+0.0796]} \\
Seam ratio $\downarrow$ & \shortstack{+0.1542\\{}[\textminus{}0.0019,+0.3111]} \\
1-second similarity $\uparrow$ & \shortstack{\textminus{}0.0114\\{}[\textminus{}0.0242,+0.0024]} \\
\bottomrule
\end{tabular*}
\par\smallskip
\parbox{\linewidth}{\footnotesize The 20-frame budget pools evaluated trajectories, with return similarity restricted to returning trajectories. The 40-frame budget uses the rotation trajectory and seed 1. Metrics retain their original scales, with scene-bootstrap intervals.}
\end{minipage}
}\par
\smallskip

\subsection{Computational cost}
\label{app:computational-cost}

\label{app:longlive-cost}
\begin{table}[!htbp]
\centering
\begin{minipage}{0.80\linewidth}
\caption{Measured costs of 60-second generation on one dedicated H20 node. All adaptations here use 6-frame committing.}
\label{tab:computational-cost}
\centering
\footnotesize
\setlength{\tabcolsep}{3pt}
\renewcommand{\arraystretch}{1.08}
\begin{tabular*}{\linewidth}{@{\extracolsep{\fill}}lcccc@{}}
\reporttoprule
& \multicolumn{2}{c}{\textbf{Wall time (s)}} & \multicolumn{2}{c}{\textbf{Peak memory (GiB)}} \\
\cmidrule(lr){2-3}\cmidrule(lr){4-5}
\textbf{Configuration} & \textbf{Base} & \textbf{Adapted} & \textbf{Base} & \textbf{Adapted} \\
\midrule
\multicolumn{5}{@{}l}{\textit{Base generator frameworks}} \\
SGF & 218.1 & 211.0 & 46.2 & 46.2 \\
SGF+ & 220.5 & 212.9 & 49.0 & 49.0 \\
RF-TF & 287.4 & 269.0 & 48.9 & 48.9 \\
CF & 158.4 & 146.9 & 26.2 & 26.2 \\
LongLive & 229.2 & 224.1 & 47.0 & 47.0 \\
Rolling Sink & 269.6 & 252.4 & 49.1 & 54.7 \\
\multicolumn{5}{@{}l}{\textit{ID-Forcing adaptations}} \\
IDF-SF & 225.9 & 212.4 & 29.1 & 29.9 \\
IDF-LL & 220.4 & 226.1 & 31.8 & 31.8 \\
\bottomrule
\end{tabular*}
\end{minipage}
\end{table}

Each base configuration and its 6-frame adaptation run on neighboring GPUs. Each condition generates two consecutive 60-second videos using standard-suite prompts 0 and 1. Table~\ref{tab:computational-cost} reports the second video in each condition; the first is warm-up. Peak GPU memory is sampled once per second and includes model weights, activations, and runtime memory. Wall time excludes model loading and includes VAE decoding.

The two ID-Forcing timing adaptations use 6-frame committing, separately from the 9-frame adaptations in the quality comparisons. On SF, complete ID-Forcing uses 155 clean-KV DiT forward passes at 60 seconds, including tail recaching. The adaptation uses 80 after changing cache handover and omitting tail recaching. Both LongLive conditions use 80 clean-KV DiT forward passes. The 6 main-generator comparisons preserve DiT forward-pass counts.

\end{document}